\documentclass[10pt,a4paper]{article}

\usepackage[
  top=0.75in,
  bottom=0.85in,
  left=0.62in,
  right=0.62in
]{geometry}

\usepackage[utf8]{inputenc}
\usepackage[T1]{fontenc}

\usepackage{XCharter}
\usepackage[scaled=1.1]{zlmtt}
\usepackage{microtype}
\usepackage{helvet}

\usepackage{amsmath,amssymb,amsfonts}
\usepackage{mathtools}
\usepackage{bm}

\usepackage{graphicx}
\usepackage{booktabs}
\usepackage{multirow}
\usepackage{subcaption}
\usepackage{wrapfig}
\usepackage{float}
\usepackage[section]{placeins}
\usepackage[normalem]{ulem}
\usepackage{array}
\usepackage{tabularx}
\usepackage{makecell}

\usepackage{multicol}
\usepackage{algorithm}
\usepackage{algorithmic}

\usepackage{setspace}
\usepackage{enumitem}
\setlist{nosep,leftmargin=*,topsep=2pt,parsep=1pt,itemsep=1pt}

\usepackage[table]{xcolor}

\definecolor{jluDeepBlue}{RGB}{0,51,153}
\definecolor{jluRed}{RGB}{198,40,40}
\definecolor{jluTeal}{RGB}{0,137,123}
\definecolor{jluAmber}{RGB}{255,160,0}
\definecolor{jluPurple}{RGB}{106,27,154}
\definecolor{jluLightBg}{RGB}{245,247,252}
\definecolor{jluGray}{RGB}{100,100,100}

\newcommand{\MonthYear}{\the\year-\ifnum\month<10 0\fi\the\month}

\usepackage[colorlinks=true,linkcolor=jluDeepBlue,citecolor=jluTeal,urlcolor=jluRed]{hyperref}
\usepackage{cleveref}
\usepackage[numbers,sort&compress]{natbib}

\usepackage{fancyhdr}
\usepackage{tikz}
\usetikzlibrary{calc}
\usepackage{tcolorbox}
\tcbuselibrary{skins,breakable}

\usepackage{listings}

\lstdefinestyle{prompt}{
  basicstyle=\ttfamily\footnotesize,
  columns=fullflexible,
  keepspaces=true,
  breaklines=true,
  breakatwhitespace=true,
  breakindent=1.2em,
  frame=single,
  rulecolor=\color{jluDeepBlue!35!white},
  backgroundcolor=\color{jluLightBg},
  framerule=0.6pt,
  aboveskip=8pt,
  belowskip=8pt,
  captionpos=t
}
\lstnewenvironment{promptlisting}[1][]{\lstset{style=prompt,#1}}{}
\usepackage{titlesec}
\usepackage{caption}
\titleformat{\section}
  {\normalsize\rmfamily\bfseries\color{jluDeepBlue}}
  {\colorbox{jluDeepBlue}{\textcolor{white}{\;\thesection\;}}}{0.5em}{}
  [\vspace{1pt}{\color{jluDeepBlue}\titlerule[0.8pt]}]

\titleformat{\subsection}
  {\normalsize\rmfamily\bfseries\color{black}}
  {\thesubsection}{0.5em}{}

\titleformat{\subsubsection}
  {\small\sffamily\bfseries\color{black}}
  {\thesubsubsection}{0.5em}{}

\titlespacing*{\section}{0pt}{1.0em}{0.4em}
\titlespacing*{\subsection}{0pt}{0.7em}{0.25em}
\titlespacing*{\subsubsection}{0pt}{0.5em}{0.15em}

\newcommand{\logobadge}[2]{%
  \tikz[baseline=-0.5ex]{%
    \node[fill=#1, text=white, rounded corners=3pt,
          inner xsep=6pt, inner ysep=3pt,
          font=\small\sffamily\bfseries]{#2};}}

\newcommand{\logoimg}[4][1.9em]{%
  \IfFileExists{logo/#2.pdf}%
    {\includegraphics[height=#1]{logo/#2}}%
    {\IfFileExists{logo/#2.png}%
      {\includegraphics[height=#1]{logo/#2}}%
      {\IfFileExists{logo/#2.jpg}%
        {\includegraphics[height=#1]{logo/#2}}%
        {\IfFileExists{logos/#2.pdf}%
          {\includegraphics[height=#1]{logos/#2}}%
          {\IfFileExists{logos/#2.png}%
            {\includegraphics[height=#1]{logos/#2}}%
            {\IfFileExists{logos/#2.jpg}%
              {\includegraphics[height=#1]{logos/#2}}%
              {\IfFileExists{#2.pdf}%
                {\includegraphics[height=#1]{#2}}%
                {\IfFileExists{#2.png}%
                  {\includegraphics[height=#1]{#2}}%
                  {\IfFileExists{#2.jpg}%
                    {\includegraphics[height=#1]{#2}}%
                    {\logobadge{#3}{#4}}}}}}}}}}}

\newtcolorbox{highlightbox}[1][]{%
  enhanced, breakable,
  colback=jluLightBg,
  colframe=jluDeepBlue,
  coltitle=white,
  fonttitle=\small\sffamily\bfseries,
  boxrule=0.6pt,
  arc=3pt,
  left=5pt, right=5pt, top=4pt, bottom=4pt,
  title=#1}

\begin{document}
\thispagestyle{fancy}

\begin{center}
  {\fontsize{18}{24}\selectfont\rmfamily\bfseries\color{black}%
Beyond Starry Night: Shortcut-Aware Control-State Planning for Artist-Grounded Text to Image Generation
\par}
  \vspace{6pt}

  {\normalsize\rmfamily\bfseries
    Kuan Xing\textsuperscript{1,$\dagger$},
    Ye Wang\textsuperscript{1,$\dagger$},
    Changyi Gan\textsuperscript{1},
    Yuheng Li\textsuperscript{2},
    Thao Nguyen\textsuperscript{3},
    Yi Chang\textsuperscript{1},
    Yilin Wang\textsuperscript{2,$\ddagger$}\par}
  \vspace{4pt}
  {\footnotesize\rmfamily\color{black}
    \textsuperscript{1}Jilin University \quad
    \textsuperscript{2}Adobe \quad
    \textsuperscript{3}University of Wisconsin\par}
  \vspace{2pt}
  {\footnotesize\rmfamily\color{black}
    \textsuperscript{$\dagger$}Equal contribution.\quad
    \textsuperscript{$\ddagger$}Project lead.}
\end{center}

\vspace{-1em}
\noindent{\color{jluDeepBlue}\rule{\textwidth}{0.7pt}}
\vspace{-0.6em}

\begin{center}
    \includegraphics[width=0.92\textwidth]{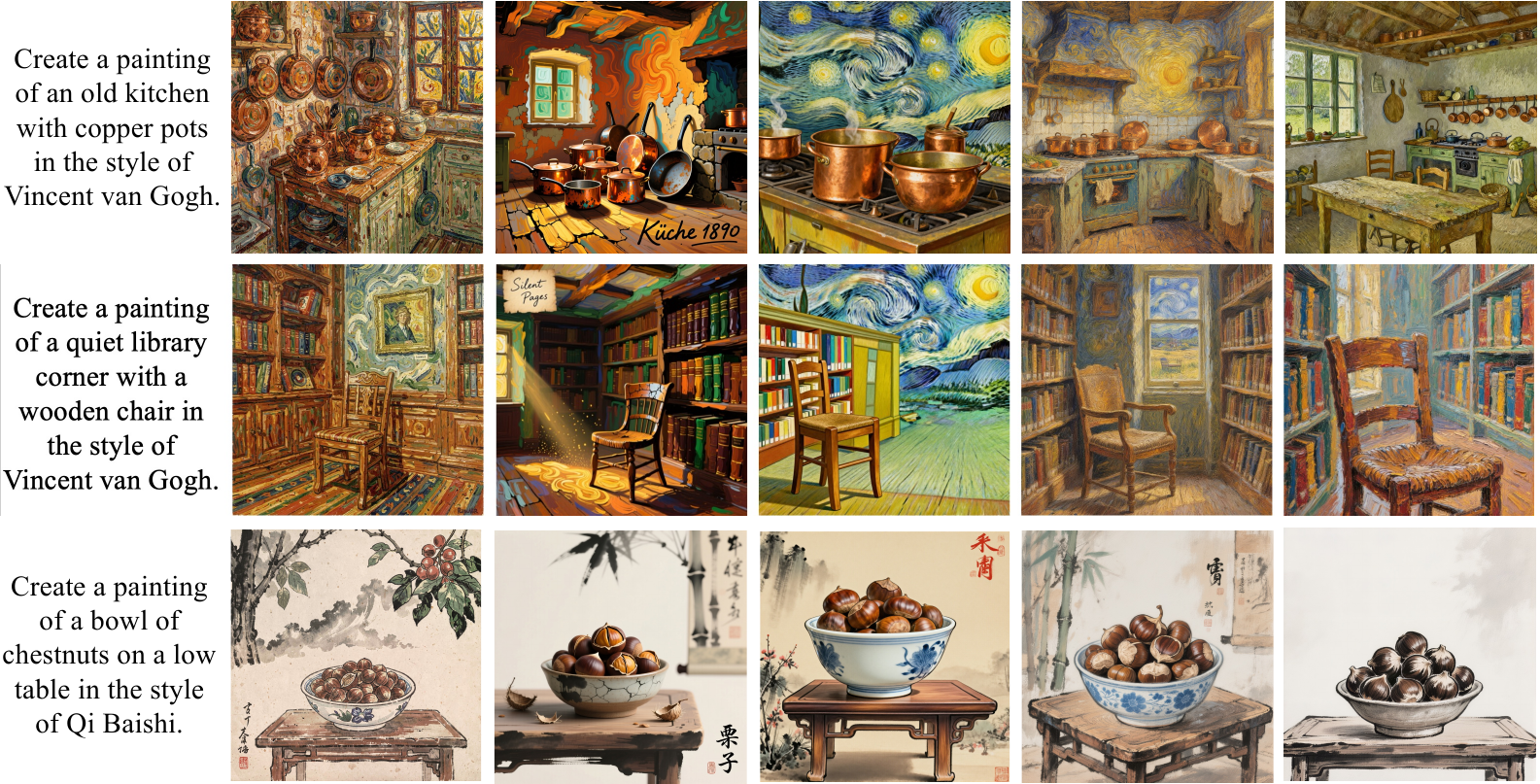}

    \vspace{0.01cm} 

    \includegraphics[width=0.92\textwidth]{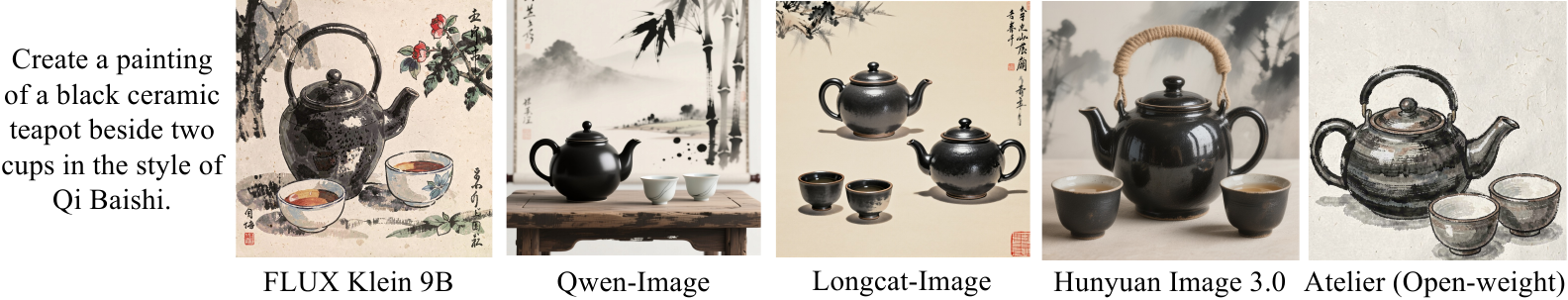}

    \captionsetup{type=figure}
    \captionof{figure}{
    Artist names can trigger unrequested canonical cues. Given neutral scene prompts, direct artist-conditioned generators and prompt-expansion agents frequently insert highly recognizable motifs or treatments that are unsupported by the requested scene. Atelier explicitly separates content to preserve, attributes to transform, and canonical cues to avoid, producing artist-grounded outputs while retaining the requested subject.
    }
    \label{fig:teaser}
\end{center}

\begin{center}
\begin{tcolorbox}[
  enhanced,
  width=0.92\textwidth,
  colback=jluDeepBlue!6!white,
  colframe=jluDeepBlue!35!white,
  boxrule=0.6pt,
  arc=3pt,
  left=6pt,right=6pt,top=5pt,bottom=5pt
]
  {\small
  \noindent{\sffamily\bfseries\color{jluDeepBlue}Abstract\quad}%
  Artist-grounded image generation requires more than appending an artist name to a prompt. Image models often respond to artist names through canonical shortcuts, such as recurring motifs, generic palettes, or overrepresented period signatures, rather than preserving the user's intended scene. We introduce Atelier, a shortcut-aware control-state planning framework for artist-grounded image generation. Atelier translates underspecified artistic intent into an explicit control state that separates scene anchors, preserve/transform decisions, style-regime hypotheses, role-bound artist evidence, and shortcut-avoidance constraints. It grounds this state using artist-level knowledge and local patch references, compiles backend-aware generation plans, and iteratively refines candidates through global and local authenticity feedback. We further introduce ArtIntentBench, a benchmark covering Van Gogh and Qi Baishi across artwork re-rendering, period/style-controlled generation, historically unseen subjects, shortcut auditing, and human preference evaluation. Across open-weight and closed-source generators, Atelier improves artist-level style fidelity, preserves source structure more faithfully, and substantially reduces shortcut substitution compared with prompt-engineered, retrieval-augmented, and general-purpose agent baselines. These results suggest that artist-grounded generation is bottlenecked not only by image synthesis, but by the upstream inference of explicit, evidence-grounded artistic controls.}
\end{tcolorbox}
\end{center}

\vspace{0.4em}

\section{Introduction}
\label{sec:intro}

Modern text-to-image models \citep{rombach2022ldm, saharia2022imagen, esser2024scaling} can produce visually plausible images from open-ended prompts, but artist-grounded generation remains difficult when the prompt specifies only a scene and an artist. A request such as “a quiet subway platform in the style of Van Gogh” does not specify which objects must remain stable, which attributes may be transformed, which period of the artist’s work should guide the output, or which motifs are inappropriate for the scene. As shown in Figure ~\ref{fig:teaser} a model may
produce generic painterly texture, overuse a small set of canonical visual cues,
or replace the requested subject with motifs associated with the
artist.

An important failure mode in artist-grounded generation is \emph{canonical shortcutting}: a model relies on a small set of high-frequency visual associations rather than modeling the artist's visual language as a context-sensitive combination of composition, palette, brushwork, and motif selection. The term follows the broader notion of shortcut learning, in which models exploit spuriously correlated cues in place of the intended structure \citep{geirhos2020shortcut, goyal2025shortcuts}. Not all canonical associations are shortcuts---a swirling sky is appropriate for a Van~Gogh Saint-R\'{e}my nocturne, and calligraphic inscription is appropriate when requested. A shortcut is an \emph{unrequested} insertion: a motif or treatment applied outside its semantic, historical, or compositional context, or one that overrides the requested subject. These failures can appear plausible overall while replacing artist-specific visual structure with generic canonical cues.

To test whether shortcutting is systematic rather than anecdotal, we audit four strong baseline generators (FLUX \citep{blackforestlabs2026flux2klein9b}, Qwen \citep{wu2025qwenimage}, LongCat \citep{ma2025longcatimage}, Hunyuan \citep{cao2025hunyuanimage}) on 1,000 neutral prompts per artist, avoiding explicit period or motif instructions. Each output is independently reviewed by three volunteer annotators, using the same majority-vote rule as the shortcut audit in Section~\ref{sec:main_results}: a pattern is counted only when selected by at least two annotators. Table~\ref{tab:shortcut-taxonomy} shows frequent unrequested shortcuts: Van Gogh prompts trigger Starry-Night-like skies, saturated blue-yellow palettes, heavy impasto, and cypresses; Qi Baishi prompts trigger generic ink landscapes, calligraphic inscriptions, or generic decorative Chinese illustration. These failures are visually plausible but artistically misgrounded because they replace the requested scene with high-frequency artist-name associations.

\begin{table}[t]
\small
\centering
\caption{Frequent unrequested shortcut patterns observed across four baseline models (FLUX, Qwen, LongCat, Hunyuan) on 1{,}000 neutral prompts per artist. Each image was independently reviewed by three volunteer annotators; a pattern was counted when selected by at least two. Frequencies are approximate proportions across the pooled baseline images.}
\label{tab:shortcut-taxonomy}
\begin{tabular}{clc}
\toprule
 & Unrequested canonical pattern & Approx.\ freq. \\
\midrule
\multicolumn{3}{l}{\textit{Van Gogh}} \\
1 & \emph{Starry Night} swirling sky & $\sim$55\% \\
2 & Saturated blue-yellow palette regardless of scene & $\sim$52\% \\
3 & Uniform heavy impasto regardless of period & $\sim$26\% \\
4 & Cypresses inserted into scene & $\sim$17\% \\
\midrule
\multicolumn{3}{l}{\textit{Qi Baishi}} \\
5 & Generic ink-painting backgrounds (mountains, pines, mist) added without request & $\sim$45\% \\
6 & Calligraphic inscription placed without request & $\sim$23\% \\
7 & Qi Baishi's ink-wash language replaced by generic Chinese decorative illustration & $\sim$12\% \\
\bottomrule
\end{tabular}
\end{table}

We frame this problem as \emph{language-to-control translation} for artistic image
generation. The goal is not merely to imitate an artist's surface style, but to
infer an explicit control state from an incomplete natural-language request. A
faithful output must preserve the intended scene while making artist-specific
decisions about period, composition, motif selection, palette, brushwork,
material treatment, and local texture. These decisions are prior to image
synthesis: they specify what should remain stable, what may be transformed,
which artist-specific evidence should guide the transformation, and which
contextually inappropriate shortcut substitutions should be avoided.

Prior work provides effective mechanisms for image synthesis and control, but
typically assumes that the relevant control signal is already available.
Personalization methods bind subjects, concepts, or styles to learned tokens,
adapters, or fine-tuned parameters
\citep{gal2023textual, ruiz2023dreambooth, kumari2023custom, sohn2023styledrop},
while controllable-generation and editing methods guide synthesis with
structural conditions, attention control, image prompts, or explicit editing
instructions \citep{hertz2022prompt, zhang2023controlnet, ye2023ipadapter,
mou2024t2iadapter, brooks2023instructpix2pix}. These methods improve execution
fidelity once the user or system provides adequate controls. In contrast, the
setting studied here requires the system to infer those controls from
ambiguous, high-level artistic intent. General prompt expansion and tool-using
agents can add descriptors, retrieve references, and iterate on outputs
\citep{yao2023react, shen2023hugginggpt, shinn2023reflexion}, but they usually
do not maintain an inspectable representation of preserve/transform decisions,
scene-role evidence bindings, and artist-specific shortcut constraints.

We present \texttt{Atelier}, a shortcut-aware control-planning framework for artist-grounded image generation. Atelier first predicts an artistic control state that decomposes the request into scene anchors, preserve/transform policies, style-regime hypotheses, retrieval targets, and anti-shortcut constraints. It then retrieves artist-level knowledge and local patch evidence, binds patches to scene roles, compiles backend-aware generation plans, and iteratively refines candidates using global and local authenticity critics. Unlike generic retrieval-augmented prompting, Atelier does not use references as undifferentiated style examples; it binds evidence to the specific scene roles and transformation decisions encoded in the control state.

Evaluating this problem also requires more than generic image quality or
text-image alignment. Artist-grounded failures are often local: an image may be
plausible overall while misrepresenting brushwork, motif treatment, material
texture, or period-specific surface language. Standard distributional and
alignment metrics provide useful coarse signals
\citep{heusel2017fid, radford2021clip}, but recent work suggests that
fine-grained artistic understanding remains challenging for contemporary multimodal models
\citep{bin2024finegrained, hayashi2024artwork, strafforello2025art}. We
therefore evaluate both the intermediate control state and the final image, and
we separate global image-level assessment from local authenticity diagnosis.
To treat shortcutting as an empirical rather than a post hoc concern, we audit baseline failure patterns, validate them with expert review, and freeze the taxonomy before evaluating \texttt{Atelier} against it.

\paragraph{Contributions.}
\begin{enumerate}
    \item We identify and quantify canonical shortcutting, a systematic failure in which artist-name conditioning inserts high-frequency motifs or treatments that are unsupported by the requested scene. We construct frozen, human-validated shortcut taxonomies for Van Gogh and Qi Baishi and use them to measure shortcut substitution across open-weight and closed-source generators.

    \item We introduce Atelier, a shortcut-aware control-planning framework that translates an underspecified artistic request into an explicit control state. The state separates scene anchors, preserve-versus-transform decisions, style-regime intent, role-bound artist evidence, and anti-shortcut constraints, and is iteratively revised using global and local evaluation feedback.

    \item We evaluate Atelier on artist-specific control tasks covering artwork reconstruction, period-controlled generation, out-of-oeuvre modern subjects, and shortcut avoidance. Under matched backend and generation budgets, structured control improves artist-specific style fidelity and reduces shortcut substitution compared with prompt expansion, retrieval-augmented prompting, and generic iterative agents.

\end{enumerate}

\section{Method}

\label{sec:method_overview}
We propose \textbf{Shortcut-Aware Artistic Control Planning (SACP)}, a method for artist-grounded generation from underspecified requests. Given an underspecified request, SACP infers a structured control state, retrieves artist evidence, binds evidence to scene roles, compiles a generation plan, and revises the state using global/local authenticity feedback. The method is \emph{shortcut-aware} because anti-shortcut constraints and preserve/transform decisions are explicit fields of the control state, read directly by retrieval, planning, and evaluation rather than embedded in the natural-language prompt.

\begin{figure}[!t]
\centering
\includegraphics[width=\columnwidth]{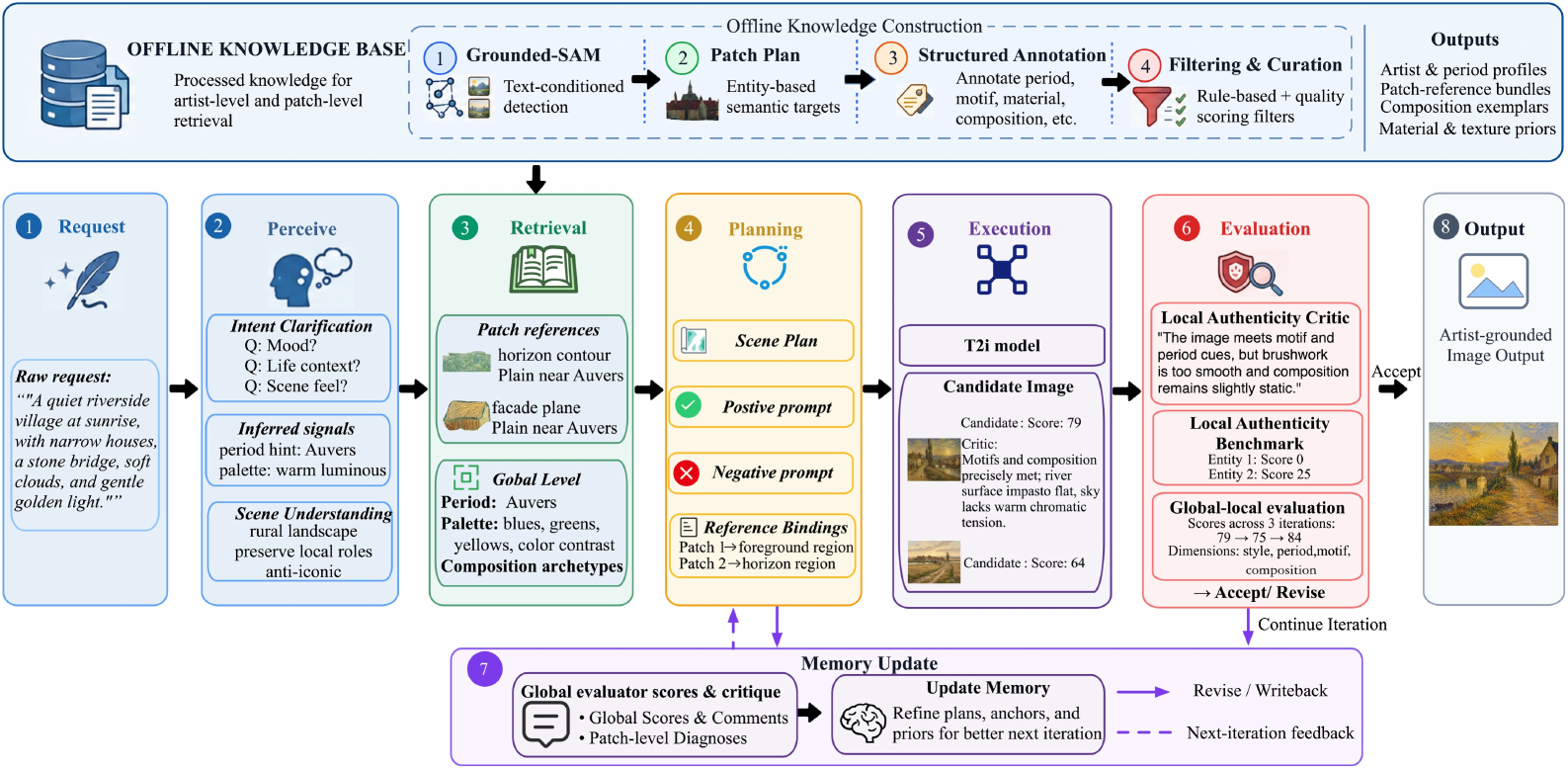}
\caption{Overview of \texttt{Atelier}, our instantiation of SACP. The system translates an underspecified artistic request into an explicit artistic control state, retrieves artist-specific global knowledge and local patch evidence, and compiles a backend-aware generation plan. Candidate images are evaluated by a global critic and \textsc{AuthCritic}, a trained patch-level authenticity critic, whose feedback is written into iterative memory for closed-loop refinement and final selection.}
\label{fig:framework}
\end{figure}

We instantiate SACP in \texttt{Atelier} (Figure~\ref{fig:framework}), a closed-loop runtime that extends the perceive--plan--execute--evaluate paradigm~\citep{yao2026photoagent} with the two artist-specific representations above: the structured control state, and a multimodal artist knowledge base whose global references and local patches are bound to scene roles rather than appended as generic style examples. The runtime can be summarized as
\[
\begin{aligned}
  x &\xrightarrow{\textsc{Perceive}} z^{(0)}
  \xrightarrow{\textsc{Derive}} z
  \xrightarrow{\textsc{Retrieve}} k
  \xrightarrow{\textsc{Plan}} (\widetilde z,c,e),\\
e &\xrightarrow{\textsc{Execute}} \mathcal{Y}
  \xrightarrow{\textsc{Evaluate}} \rho
  \xrightarrow{\textsc{Reflect}} m .
\end{aligned}
\]
Given an underspecified request $x$, \texttt{Atelier} first infers what should
be preserved, what may be artistically transformed, which period or
style-regime cues are relevant, and which artist-name shortcuts should be
avoided. It then retrieves global artist knowledge and local patch references,
binds them to scene roles, generates candidate images through one or more
backends, and revises subsequent rounds using feedback from a global critic and
\textsc{AuthCritic}, a trained patch-level authenticity critic, stored in memory.

\subsection{Perceive: Artistic Control State}
\label{sec:control_state}
The perceiver, implemented as a prompted LLM, converts the raw request into an explicit artistic control state
rather than a longer prompt. We write this state as
\[
z = (s, q, h, r, b),
\]
where $s$ is the scene reading, $q$ contains preserve-versus-transform
decisions, $h$ stores historical or style-regime intent, $r$ records
retrieval targets and anti-shortcut constraints, and $b$ contains
backend-facing execution preferences.

The scene reading identifies objects, regions, relations, and mood cues in the
request. The preserve/transform component marks which elements should remain
semantically stable and which attributes may be translated into the target
artist's visual language. The historical/style component records explicit or
implicit period cues, such as Van Gogh's Paris, Arles, Saint-Remy, or Auvers
periods, or Qi~Baishi style regimes defined by motif, composition, ink economy,
inscription, and seal usage. The anti-shortcut component specifies forbidden
substitutions, for example replacing a requested modern object with a canonical
artist motif. This state is the main interface between language understanding,
artist evidence, generation, and evaluation. Table~\ref{tab:kb_example}
previews the retrieved evidence for one held-out request;
Appendix~\ref{app:walkthrough-single} exhibits the full state $z$ recorded
for the same request, and Appendix~\ref{app:schema-zmap} gives the
field-level map from $(s,q,h,r,b)$ to the concrete record paths.
In the closed-loop runtime (Algorithm~\ref{alg:atelier_runtime}), the
perceiver resolves the request once into an episode-level interpretation
context. At the start of each round, the runtime rebuilds a base world model
from the request, its clarification, and the visual anchor, then combines
that base with the planner's memory-conditioned working view. The bootstrap
inputs therefore remain fixed, while the operational state used for planning
and generation may change within a round. The merge is artist-conditioned:
the Qi~Baishi path retains the rebuilt scene skeleton, structural anchors,
preserve/transform policy, local roles, and anti-shortcut fields, whereas the
Van~Gogh path retains a planner-emitted world model when one is present.
Evaluator feedback also generates the anti-shortcut and repair constraints in
$r$ and updates backend preferences in $b$. The planner may revise the
working period hypothesis in $h$ in light of the retrieved evidence. A
user-specified period is enforced deterministically by binding each compiled
generation call to that period's resources, independent of the working
hypothesis.

\subsection{Retrieve: Artist Knowledge and Local Evidence}
\label{sec:artist_kb}

\texttt{Atelier} grounds $z$ with an artist-specific knowledge base
$\mathcal{B}_a$, organized in three layers of evidence: a global layer of
artist-level style priors, a supplementary layer of curated summaries that
reinforce those priors, and a local layer of annotated patches from source
artworks.

At runtime, the retrieval module produces an evidence bundle
\[
k = K(x, z; \mathcal{B}_a).
\]
The bundle contains both global references and local patches. Global references
provide period, motif, composition, and palette priors. Local patches provide
region-level evidence for scene roles such as foreground subject, background
field, scene wrapper, material surface, or texture target. Unlike ordinary
retrieval-augmented prompting, these references are not appended as generic
style examples; they are explicitly bound to the scene roles and constraints in
$z$.

\par\smallskip Knowledge Base Composition. The global layer captures
artist-level style characteristics, period-specific tendencies, and recurring
motif and composition priors. The supplementary layer contributes curated
art-historical evidence that reinforces period and motif annotations. The
local layer contains a bank of patches extracted from source artworks; each
patch is annotated with a semantic role, source-work metadata, local style
traits, material treatment, and brushwork description.

\par\smallskip Offline Construction Pipeline. The
patch-grounded component is built through a four-stage offline pipeline. (1)
Image-level patch planning associates each source artwork with semantically
meaningful target regions and roles. (2) Grounded-SAM~\citep{ren2024groundedsam}-based extraction obtains
candidate local patches conditioned on these targets. (3) A vision-language
model performs structured annotation, producing semantic descriptions of
content and localized style descriptors such as brushwork and palette. (4)
Annotated patches are filtered for quality and grouped into family-level
reference sets used during retrieval.

\par\smallskip Runtime Retrieval. At runtime, the retriever
assembles a request-conditioned knowledge context from the user request and the
current scene interpretation. The context contains four components: period
guidance for likely stylistic bias, selected local motifs identifying
request-relevant regions, retrieved reference works providing artwork-level
context, and a patch reference bundle providing localized visual grounding. The
planner consumes this context for both global style decisions and local
rendering.
Table~\ref{tab:kb_example} shows a concrete example.

\begin{table}[t]
    \centering
    \footnotesize
    \setlength{\tabcolsep}{4pt}
    \caption{Knowledge context retrieved for the held-out request
    \emph{``A low-angle still life of harvested vegetables---pale cabbage,
    reddish root crops, and scattered yellow maple leaves---resting on dark
    soil, rendered in Paris-period muted earth tones with directional
    impasto brushwork.''} Values are taken verbatim from the recorded run;
    Appendix~\ref{app:walkthrough-single} gives the full state $z$ for the
    same request.}
    \label{tab:kb_example}
    \begin{tabularx}{\columnwidth}{@{}>{\raggedright\arraybackslash}p{0.28\columnwidth}
    >{\raggedright\arraybackslash}X@{}}
    \toprule
    \textbf{Component} & \textbf{Retrieved Content} \\
    \midrule
    Period guidance & Paris (score 37.4), Nuenen (score 33.4);
    policy: single route on Paris (explicit user preference) \\
    \midrule
    Composition cues & close-crop focus, diagonal depth path,
    dense repetition \\
    \midrule
    Motif cues & foreground object mass, field surface band,
    background halo field \\
    \midrule
    Reference works & F102 (\emph{Still Life with a Basket of Potatoes,
    Surrounded by Autumn Leaves and Vegetables}), F374 (\emph{Red Cabbages
    and Onions}), F378 (\emph{Still Life with Basket of Apples}) \\
    \midrule
    Patch bundle & 1 bound patch (F57, role: foreground object mass);
    2 roles unbound, fall back to work-level evidence \\
    \bottomrule
    \end{tabularx}
\end{table}

\subsection{Plan and Execute: Backend-Aware Generation}
\label{sec:planner_executor}

The planner, implemented as a prompted LLM, converts the control state and retrieved evidence into a
working state view and a backend-aware generation plan. At iteration $t$, it computes
\[
(\widetilde z^{(t)},c^{(t)}, e^{(t)}) = P(x, z^{(t)}, k^{(t)}, m^{(t-1)}),
\]
where $\widetilde z^{(t)}$ is the planner-emitted working view,
$c^{(t)}$ contains the artistic controls and reference bindings, and
$e^{(t)}$ specifies the backend execution policy. The plan includes scene
anchors, period or style-regime targets, global reference works, local
patch-role bindings, negative shortcut constraints, and any backend-specific
parameters.

The executor maps this plan to concrete generation calls. Backends with reference, adapter, or LoRA-style control receive local patch references, period-specific resources, patch-resource bindings, and scale parameters. General purpose backends receive a compact textual realization of the same control
state, including preservation constraints, local style targets, and forbidden
substitutions. Early rounds may explore multiple backends; later rounds can lock
onto the strongest backend once the evaluator identifies a reliable direction.

Algorithm~\ref{alg:atelier_runtime} summarises how these components are
combined at runtime. At each iteration, \textsc{Derive} rebuilds the round's
base state from $z^{(0)}$ and generates the constraints in $r$ and preferences
in $b$ from $m^{(t-1)}$. The planner may return a working state view together
with $c^{(t)}$ and $e^{(t)}$; \textsc{Merge} applies the artist-conditioned
field policy above before execution. Revision therefore changes the plan,
evidence bindings, execution policy, and any planner-revisable state fields,
without changing the request and clarification that anchor the episode.

Within-round candidate selection and the trajectory-level \textsc{FinalSelect}
share a global-score margin gate followed, when local evidence is available,
by an \textsc{AuthCritic} rerank. Their final tie-breakers are stage-specific:
the within-round rule uses recommendation, confidence, backend preference, and
candidate order, whereas \textsc{FinalSelect} uses the second-stage score and
iteration order. Appendix~\ref{app:final-select} gives both rules. If the loop
terminates without an accepted candidate, \textsc{FinalSelect} selects
$y^\star$ from the full trajectory $\tau$.

Algorithm~\ref{alg:atelier_runtime} shows the acceptance exit explicitly;
the deployed stop policy additionally includes plateau and unrecoverable
failure exits, specified in Appendix~\ref{app:stop-policy}.

\begin{figure}[t]
\centering
\begin{minipage}{0.80\linewidth}
\hrule
\vspace{0.25em}
\captionsetup{type=algorithm}
\captionof{algorithm}{Runtime loop of \texttt{Atelier}}
\label{alg:atelier_runtime}
\vspace{-0.35em}
\hrule
\vspace{0.35em}
\small
\begin{algorithmic}[1]
\REQUIRE request $x$, artist knowledge base $\mathcal{B}_a$, iteration budget $T$
\ENSURE final image $y^\star$, trajectory $\tau$, and final memory $m^\star$
\STATE $m^{(0)} \gets \emptyset$, $\tau \gets \emptyset$
\STATE $z^{(0)} \gets \textsc{Perceive}(x)$ \COMMENT{fixed interpretation context}
\FOR{$t=1$ to $T$}
    \STATE $z^{(t)} \gets \textsc{Derive}(z^{(0)}, m^{(t-1)})$ \COMMENT{base state from $z^{(0)}$; constraints and preferences from memory}
    \STATE $k^{(t)} \gets \textsc{Retrieve}(x, z^{(t)};\mathcal{B}_a)$ \COMMENT{re-bind global and local evidence}
    \STATE $(\widetilde{z}^{(t)}, c^{(t)}, e^{(t)}) \gets \textsc{Plan}(x, z^{(t)}, k^{(t)}, m^{(t-1)})$
    \STATE $z^{(t)} \gets \textsc{Merge}(z^{(t)}, \widetilde{z}^{(t)})$ \COMMENT{artist-conditioned field policy}
    \STATE $\mathcal{Y}^{(t)} \gets \textsc{Execute}(e^{(t)}, c^{(t)})$
    \STATE $\rho_A^{(t)} \gets \textsc{AuthCritic}(\mathcal{Y}^{(t)})$ \COMMENT{pre-pass: per-patch scores, low-score patches}
    \STATE $\rho^{(t)} \gets \textsc{GlobalCritic}(\mathcal{Y}^{(t)}, \phi(x, z^{(t)}, k^{(t)}), \rho_A^{(t)})$ \COMMENT{holistic; $\phi$: critic input projection (\S\ref{sec:evaluator})}
    \STATE $\hat{y}^{(t)} \gets \textsc{Select}(\mathcal{Y}^{(t)}, \rho^{(t)})$
    \STATE $\tau \gets \tau \cup \{(\hat{y}^{(t)}, c^{(t)}, e^{(t)}, \rho^{(t)})\}$
    \STATE $m^{(t)} \gets \textsc{Reflect}(m^{(t-1)}, \hat{y}^{(t)}, \rho^{(t)})$
    \IF{$\hat{y}^{(t)}$ satisfies the acceptance criterion}
        \STATE \textbf{break}
    \ENDIF
\ENDFOR
\STATE $y^\star \gets \textsc{FinalSelect}(\tau)$
\STATE $m^\star \gets m^{(t)}$
\STATE \textbf{return} $y^\star, \tau, m^\star$
\end{algorithmic}
\vspace{0.25em}
\hrule
\end{minipage}
\end{figure}

Shortcut avoidance is enforced at four points in the runtime loop rather than
left to prompt-level suggestion. (i) The plan $c^{(t)}$ encodes anti-shortcut
constraints as explicit fields and describes style through decomposed visual
attributes from $k^{(t)}$, rather than relying only on an artist name. (ii)
Backends with a native negative-prompt channel receive the shortcut list in
that channel; for backends without one, the executor appends the constraints
to the text prompt as an imperative \texttt{Avoid:} block. (iii) The global
critic receives the artist-specific shortcut guard. On unsupported-subject
routes, matched canonical aliases trigger a deterministic penalty to intent,
motif, composition, and style scores, increase the artifact penalty, and
downgrade an accept recommendation to revise. (iv) The resulting violation
tag and repair actions enter memory and condition the next round's constraints.

The remainder of this section describes the two runtime components not yet covered---the two critics that produce $\rho^{(t)}$ inside the loop, and the reflective memory update $m^{(t)}$ that carries information across iterations.

\subsection{Evaluate: Global Critic and \textsc{AuthCritic}}
\label{sec:evaluator}
Each candidate image is evaluated in a two-stage pipeline. \textsc{AuthCritic} runs first on each candidate and produces per-patch authenticity scores; the low-scoring patches are surfaced to the global critic as focus signals, with their per-patch metadata entering the prompt text and their image crops attached as additional visual inputs alongside the full candidate. The global critic then judges the candidate as a whole for semantic fidelity, composition, palette, and artist alignment, with its attention already focused on the regions \textsc{AuthCritic} has flagged. Both signals feed candidate selection: global-critic scores gate a shortlist (for acceptance and within-margin candidates), and \textsc{AuthCritic} reranks the shortlist by patch-level authenticity. The combined evaluation record is written into the reflective memory update.

\paragraph{Global critic.}
We prompt Kimi~K2.6~\citep{moonshot2026kimik26} as the global critic, without task-specific fine-tuning. The critic is conditioned on the candidate image and a fixed-schema projection $\phi(x,z,k)$ of the request, the current control state, and the retrieved evidence: the raw and clarified request, the working period hypothesis with its candidate alternatives, the selected motif and composition cues, and the focus signals from \textsc{AuthCritic}. Appendix~\ref{app:critic-projection} specifies the projection in full. It returns a structured judgment covering semantic consistency with the request, composition, period or style-regime match, palette, overall artist alignment, and visible artifacts. Each judgment includes a score, a natural-language critique, and an accept/revise recommendation.

Because the visual evidence that constitutes artistic authenticity is
artist dependent, the global critic uses a shared output contract with an
artist-specific rubric. The Van~Gogh instance evaluates period-conditioned
oil-paint handling, directional brushwork, surface relief, palette
relations, and canonical-shortcut avoidance; the Qi~Baishi instance
instead evaluates economical calligraphic strokes, wet and dry ink
variation, layered ink-density transitions, intentional paper reserve,
asymmetrical placement, restrained color accents, and avoidance of
unrequested generic Chinese decorative conventions. The two instances
share the same structured output fields and numerical aggregation
(Appendix~\ref{app:score-aggregation}), so the runtime policy is unchanged
while the artistic interpretation of the fields is specialized to the
target artist.

\paragraph{\textsc{AuthCritic}.}
\textsc{AuthCritic} is a patch-level authenticity critic obtained by supervised fine-tuning of Gemma-4~\citep{googledeepmind2026gemma4} (E4B-it) with a LoRA adapter (rank $16$, $\alpha=32$, dropout $0.05$). We use supervised fine-tuning rather than preference-based training because each patch carries a deterministic source label from its known origin image, and the runtime loop consumes calibrated per-patch scores rather than pairwise preferences. The training corpus contains $15{,}000$ patches evenly split across three source classes: $5{,}000$ real Van~Gogh patches from the offline patch bank (Section~\ref{sec:artist_kb}), $5{,}000$ synthetic Van~Gogh patches cropped from Van~Gogh-style outputs of six text-to-image models spanning open-weight and closed-source families, and $5{,}000$ other-painter patches from contemporaries and near-contemporaries. These three sources let \textsc{AuthCritic} distinguish real Van~Gogh from both the failure modes of current generators and from stylistically adjacent artistic traditions. Each example targets a structured JSON containing a categorical source-type label and seven grounded reasoning fields covering patch content, source attribution, and style differences; training uses standard cross-entropy over all target tokens.

We train for $2$ epochs on a single H800 GPU with AdamW at learning rate $2\times10^{-4}$ and effective batch size $8$. The splits are disjoint at the level of source works, and on a held-out test set of $1{,}485$ patches, \textsc{AuthCritic} reaches $91.65\%$ binary accuracy (real Van~Gogh vs.\ others). At inference, each patch yields a source-type score and, when available, a proximity-to-real estimate. The candidate-level score combines the real-patch fraction, mean synthetic-patch proximity, valid patch count, and parse-error rate; a simple mean is retained only as the legacy fallback when proximity outputs are unavailable. Appendix~\ref{app:authcritic-aggregation} gives the exact aggregation. This score, together with the global critique, is written into reflective memory (Section~\ref{sec:memory}).

The Qi~Baishi instantiation uses the same LoRA configuration and optimizer settings on a smaller corpus of $4{,}806$ patches spanning the same three source classes: real Qi~Baishi patches, synthetic patches cropped from Qi~Baishi-style generations, and other-painter patches from Wu~Changshuo and Zhu~Da, two stylistically adjacent ink painters. The three classes are balanced within every split, and the train, validation, and test splits ($3{,}846$/$480$/$480$) are disjoint at the level of source works. We train for three epochs with effective batch size $16$ and select the checkpoint with the lowest validation loss; on the $480$ held-out test patches, the adapter reaches $93.54\%$ binary accuracy (real Qi~Baishi vs.\ others) and $89.38\%$ accuracy over the three source classes.

\subsection{Reflect: Memory and Revision}
\label{sec:memory}

After each round, \texttt{Atelier} writes the selected candidate, backend
choice, control plan, global critique, local patch evidence, failure tags, and
repair targets into memory $m^{(t)}$. This memory serves two roles. First, it
prevents the planner from repeating failed choices, such as reusing a
shortcut motif or an unsuitable backend. Second, it preserves successful
decisions, such as a strong period reference, stable composition, or effective
patch binding.

The next round uses memory to revise the control state, update reference
bindings, adjust backend selection, and focus generation on unresolved local
failures. The loop terminates under the success, plateau, budget, or error
conditions specified in Appendix~\ref{app:stop-policy}. The final image is
selected from the full trajectory using a fixed ranking policy over global
quality, semantic preservation, local authenticity, shortcut avoidance, and
iteration order.

\section{Experiments}
\subsection{ArtIntentBench}
\label{sec:benchmark}
ArtIntentBench deliberately prioritizes depth and validation reliability over broad artist coverage. Artist-grounded generation requires artist-specific period taxonomies, motif priors, local patch evidence, and human evaluation of fine-grained visual properties such as brushwork, material treatment, and shortcut substitutions. A broad benchmark with many artist names but shallow supervision would risk evaluating generic style association rather than artist-grounded control. We therefore select two contrastive artists—Van Gogh and Qi Baishi—to cover substantially different media, cultural traditions, and style-control challenges. The benchmark evaluates generalization not only across artists, but also across periods, held-out artworks, and historically unseen subjects. For Van Gogh, we use 100 artwork re-rendering cases drawn from across his catalogued oeuvre—which comprises roughly 860 oil paintings in total—and 228 period-intent prompts distributed evenly across four periods (57 per period), yielding 228 method-period evaluations per method. For Qi Baishi, we use 80 artwork re-rendering cases based on original paintings. We also include 30 historically unseen subject prompts to evaluate both style grounding and generalization to subjects absent from the artist's body of work. Compared with style-transfer and stylization evaluations, which typically condition on a single exemplar to a few dozen reference images per style and report a single global metric \citep{gatys2016style, huang2017adain, sohn2023styledrop}, ArtIntentBench is larger in case count and richer in task diversity, covering re-rendering, period-controlled generation, cross-tradition transfer, and unseen-subject generalization within a single evaluation protocol. General generation and editing benchmarks evaluate instruction following and compositional control at scale \citep{wu2024conceptmix, ye2025imgedit, pan2025icebench}; to our knowledge, none couples artist-grounded intent with period control, content preservation, and shortcut auditing.


\subsection{Experimental Setup}
\label{sec:exp_setup}

\noindent\textbf{Downstream Tasks.}
We design four downstream tasks in ArtIntentBench: Van Gogh artwork re-rendering, Van Gogh period-controlled generation, historically unseen subject generation, and Qi Baishi artwork re-rendering.

For Van Gogh, we first evaluate artwork re-rendering, where each authentic painting is converted into a structured caption by Qwen3-VL-Plus~\citep{bai2025qwen3vl}, and all methods generate from the caption without access to the original image. We then evaluate period-controlled generation by fixing the scene prompt and varying only the target period among Paris, Arles, Saint-R\'emy, and Auvers. 

For Qi Baishi, we evaluate artwork re-rendering from original paintings to test whether the framework generalizes from Western oil painting to sparse ink-and-color painting. Finally, we include historically unseen subject generation as a diagnostic task, using modern subjects absent from the artist's oeuvre to test whether the system can apply an artist's visual language beyond canonical motifs.

\noindent\textbf{Metrics.}
We report IntroStyle~\citep{kumar2025introstyle} $W_2$ distance ($\downarrow$) for artist-level style proximity and CLIP similarity ($\uparrow$) for text--image alignment. In artwork re-rendering, where each prompt is paired with a source painting, we further report DreamSim~\citep{fu2023dreamsim} ($\downarrow$), DINO-cosine~\citep{caron2021dino} ($\uparrow$), and LPIPS~\citep{zhang2018lpips} ($\downarrow$) against the source. Human preference is measured with blinded Rank-1 rates from 15 art-trained raters recruited from art-related graduate programs: 10 without professional teaching or curatorial experience (non-specialists) and 5 with formal training in art history, studio practice, or art education (experts). Task-specific diagnostic metrics are introduced where they are used.

\noindent\textbf{Metric rationale.}
IntroStyle embeds images in a style-attribution feature space designed to
isolate stylistic execution from depicted content; the $W_2$ distance is
computed between the feature distribution of a method's outputs and that
of a reference set of authentic paintings, so a low score reflects
proximity to the artist's stylistic range rather than to any single
exemplar. For period-controlled generation the reference set is restricted
to the target period, and for unseen subjects it contains 20 authentic
Van~Gogh works balanced across the four periods, five per period.
Period-restricted references penalise canonical shortcutting directly: a
swirling \textit{Starry Night} sky inserted under a Paris-period target
increases the distance rather than reducing it. $W_2$ alone does not
measure content preservation or shortcut avoidance. Fidelity to source
paintings is measured by DreamSim, DINO-cosine, and LPIPS, prompt
adherence by CLIP, and unrequested canonical insertions by the
human-annotated shortcut rate of \S\ref{sec:shortcut_audit}.

\noindent\textbf{Atelier Settings.}
We instantiate Atelier in two configurations: an open-weight variant using publicly available models throughout, and a closed-source variant using commercial APIs for the perceiver, planner, and image backends. The \textbf{open-weight variant} uses DeepSeek V4 Pro~\citep{deepseekai2026deepseekv4} for both the perceiver and planner, and dispatches generation to four image backends: FLUX2.Klein-9B~\citep{blackforestlabs2026flux2klein9b}, Qwen-Image~\citep{wu2025qwenimage}, LongCat-Image~\citep{ma2025longcatimage}, and Hunyuan Image 3.0~\citep{cao2025hunyuanimage}. The \textbf{closed-source} uses GPT-5.4~\citep{openai2026gpt54} for both the perceiver and planner, and dispatches generation to two commercial image backends: Nano Banana Pro (Gemini 3 Pro Image Preview)~\citep{google2025nanobananapro} and GPT-Image-2 (ChatGPT Images 2.0)~\citep{openai2026chatgptimages2}. 
Both variants share the same scene understanding schema, artist-knowledge retrieval pipeline, patch-reference bank, iteration budget, evaluator stack, and selection policy, isolating the effect of structured control from model-family choice.

\noindent\textbf{Comparison Methods.}
We compare Atelier against three families of baselines. First, we use strong direct-backend baselines, including FLUX.2-Klein-9B, Qwen-Image, LongCat-Image, Hunyuan Image 3.0, GPT-Image-2, and Nano Banana Pro. These baselines are not given a minimal prompt of the form ``a scene in the style of artist.'' Instead, each receives a standardized prompt-engineered instruction containing the scene description, target artist or period/style regime, preservation requirements, and negative instructions against unsupported subject substitution. This setting tests whether a carefully written single-shot prompt is sufficient for artist-grounded generation. Second, we compare against agentic prompt-expansion baselines, including Claude-based and GenArtist-style \citep{wang2024genartist} baselines, which can rewrite or elaborate the request before generation but do not use Atelier's explicit control-state schema, artist knowledge base, local patch-role bindings, evaluator feedback, or iterative memory. 


\subsection{Main Results}
\label{sec:main_results}
\begin{table*}[t]
\centering
\footnotesize
\caption{Merged quantitative results on ArtIntentBench. For artwork re-rendering we report IntroStyle $W_2$ for style proximity, CLIP for text--image alignment, and DreamSim, DINO-cosine, LPIPS for perceptual fidelity to source paintings (DS/DINO/LP). For period-controlled generation we report $W_2$ and CLIP across four Van~Gogh periods. Best and second-best within each configuration are shown in bold and underlined, respectively.}
\label{tab:main_results}
\vspace{4pt}
\setlength{\tabcolsep}{1.0pt}
\renewcommand{\arraystretch}{1.05}
\resizebox{\linewidth}{!}{%
\begin{tabular}{@{}lcccccc cc cc cc cc@{}}
\toprule
\textbf{Method}
& \multicolumn{5}{c}{\makecell[c]{\textbf{Artwork Re-rendering}}}
& \multicolumn{2}{c}{\textbf{Paris}}
& \multicolumn{2}{c}{\textbf{Arles}}
& \multicolumn{2}{c}{\textbf{Saint-R\'emy}}
& \multicolumn{2}{c}{\textbf{Auvers}} \\
\cmidrule(lr){2-6}
\cmidrule(lr){7-8}
\cmidrule(lr){9-10}
\cmidrule(lr){11-12}
\cmidrule(lr){13-14}
& \textbf{$W_2\downarrow$} & \textbf{CLIP$\uparrow$} & \textbf{DS$\downarrow$} & \textbf{DINO$\uparrow$} & \textbf{LP$\downarrow$}
& \textbf{$W_2\downarrow$} & \textbf{CLIP$\uparrow$}
& \textbf{$W_2\downarrow$} & \textbf{CLIP$\uparrow$}
& \textbf{$W_2\downarrow$} & \textbf{CLIP$\uparrow$}
& \textbf{$W_2\downarrow$} & \textbf{CLIP$\uparrow$} \\
\midrule
\multicolumn{14}{c}{\textit{Closed-source configuration}} \\
\midrule
GPT-Image-2
& \underline{66.55} & 0.2861 & \underline{0.3040} & \underline{0.6847} & 0.5861
& 84.20 & 0.3428 & 75.77 & \underline{0.3539} & \underline{69.27} & 0.3504 & 71.47 & \underline{0.3590} \\

Nano Banana Pro (NBP)
& 70.61 & 0.2814 & 0.3128 & 0.6423 & 0.5882
& 93.38 & \underline{0.3485} & 92.56 & 0.3410 & 78.41 & \textbf{0.3610} & 81.54 & 0.3571 \\

GenArtist (GPT-Image-2)
& 67.77 & \underline{0.2852} & 0.3293 & 0.6416 & 0.6052
& 85.55 & 0.3447 & 75.95 & \underline{0.3539} & 69.59 & 0.3482 & 71.79 & 0.3532 \\

GenArtist (NBP)
& 73.39 & 0.2835 & 0.3311 & 0.6535 & 0.5915
& 89.24 & 0.3391 & 87.06 & 0.3431 & 77.35 & 0.3535 & 79.28 & \textbf{0.3616} \\

Claude (GPT-Image-2)
& 66.78 & \textbf{0.2914} & 0.3126 & 0.6619 & 0.5879
& 84.57 & 0.3445 & \underline{75.01} & \textbf{0.3544} & 69.28 & 0.3534 & \underline{70.05} & 0.3580 \\

Claude (NBP)
& 72.48 & 0.2847 & 0.3205 & 0.6671 & \underline{0.5855}
& 92.90 & \textbf{0.3508} & 90.40 & 0.3352 & 77.45 & \underline{0.3599} & 81.63 & 0.3566 \\

\midrule
Atelier (Closed-source)
& \textbf{64.22} & 0.2816 & \textbf{0.2933} & \textbf{0.7075} & \textbf{0.5704}
& \textbf{82.40} & 0.3399 & \textbf{73.42} & 0.3494 & \textbf{66.99} & 0.3442 & \textbf{69.83} & 0.3493 \\
\midrule
\multicolumn{14}{c}{\textit{Open-weight configuration}} \\
\midrule
FLUX
& 77.71 & 0.2546 & 0.3751 & 0.5741 & 0.6245
& 86.78 & \textbf{0.3311} & 84.30 & \underline{0.3325} & 77.96 & \textbf{0.3443} & 82.18 & \textbf{0.3376} \\

Qwen
& 76.27 & 0.2655 & 0.4068 & 0.5810 & 0.6191
& 93.50 & 0.3084 & 80.55 & 0.3209 & 74.09 & 0.3202 & 76.83 & 0.3170 \\

LongCat
& 75.56 & 0.2863 & 0.3660 & 0.6098 & 0.6090
& 84.56 & 0.2687 & 75.32 & 0.2805 & 71.59 & 0.2726 & 72.83 & 0.2832 \\

Hunyuan Image 3.0
& \underline{74.82} & \textbf{0.3008} & \underline{0.3579} & \underline{0.6194} & \underline{0.6086}
& \underline{84.36} & 0.3116 & 76.81 & 0.3293 & \underline{67.74} & 0.3227 & \underline{70.40} & 0.3308 \\

GenArtist
& 81.99 & \underline{0.2936} & 0.4915 & 0.4665 & 0.6368
& 84.67 & 0.3106 & \underline{74.65} & 0.3211 & 70.38 & 0.3160 & 71.68 & \underline{0.3238} \\

\midrule
Atelier (Open-weight)
& \textbf{73.52} & 0.2722 & \textbf{0.3440} & \textbf{0.6398} & \textbf{0.5916}
& \textbf{83.55} & \underline{0.3210} & \textbf{73.70} & \textbf{0.3338} & \textbf{64.55} & \underline{0.3333} & \textbf{69.22} & 0.3320 \\
\bottomrule
\end{tabular}%
}
\end{table*}

\begin{figure}[t]
    \centering
    \includegraphics[width=0.98\linewidth]{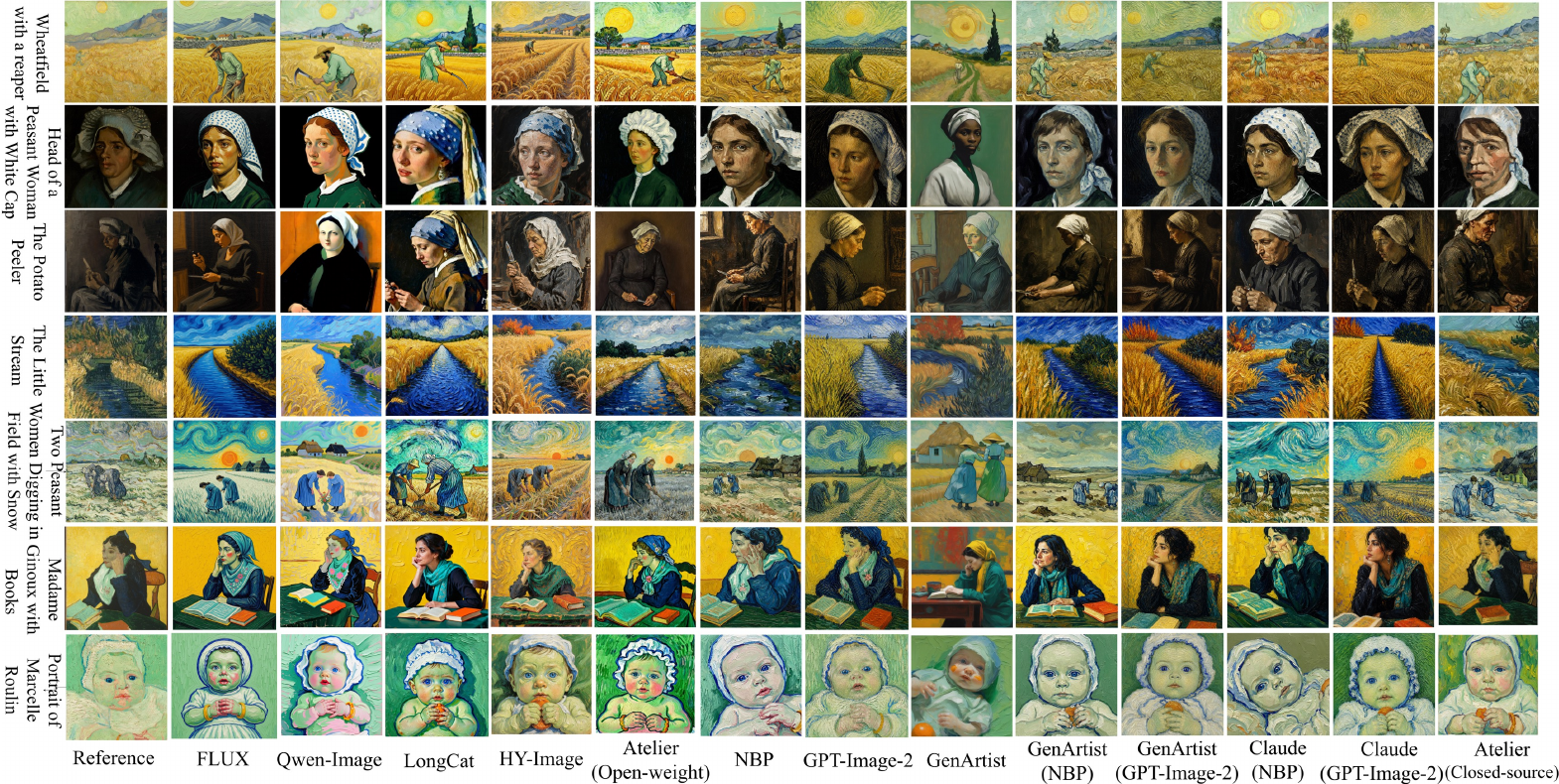}
    \caption{Text-only re-rendering from held-out Van Gogh paintings.}
    \label{fig:exp_qualitative_rerender}
\end{figure}

\noindent\textbf{Artwork Re-rendering from Authentic Paintings.}
\texttt{Atelier} achieves the lowest IntroStyle $W_2$ among the evaluated methods, with the open-weight variant scoring 73.52 and the closed-source variant scoring 64.22. The closest competitor is GPT-Image-2 at 66.55, while all other baselines obtain higher style distances. Table~\ref{tab:main_results} further reports perceptual fidelity metrics (DreamSim, DINO-cosine, LPIPS) against the source paintings. Unlike $W_2$, which measures distributional style proximity, these metrics assess how closely each generated image resembles the specific source painting. \texttt{Atelier} achieves the best scores on all three metrics in both configurations. All methods receive the same structured VLM-derived prompt. Qualitatively, as shown in Figure~\ref{fig:exp_qualitative_rerender}, Claude (GPT-Image-2) handles scene-level semantics more faithfully than direct baselines, but the wheatfield and field-with-snow skies both exhibit the swirling strokes and saturated blue-yellow palette of \textit{The Starry Night}, which appears in neither source painting, suggesting that the model responds to the artist name rather than the specific work. In contrast, \texttt{Atelier} more consistently preserves the compositional skeleton, subject scale, and spatial relationships; its brushwork follows the directional structure of the source, and its color organization maintains the original warm-cool balance and tonal layering.

\begin{figure}[!t]
    \centering
    \begin{subfigure}{\linewidth}
        \centering
        \includegraphics[width=\linewidth]{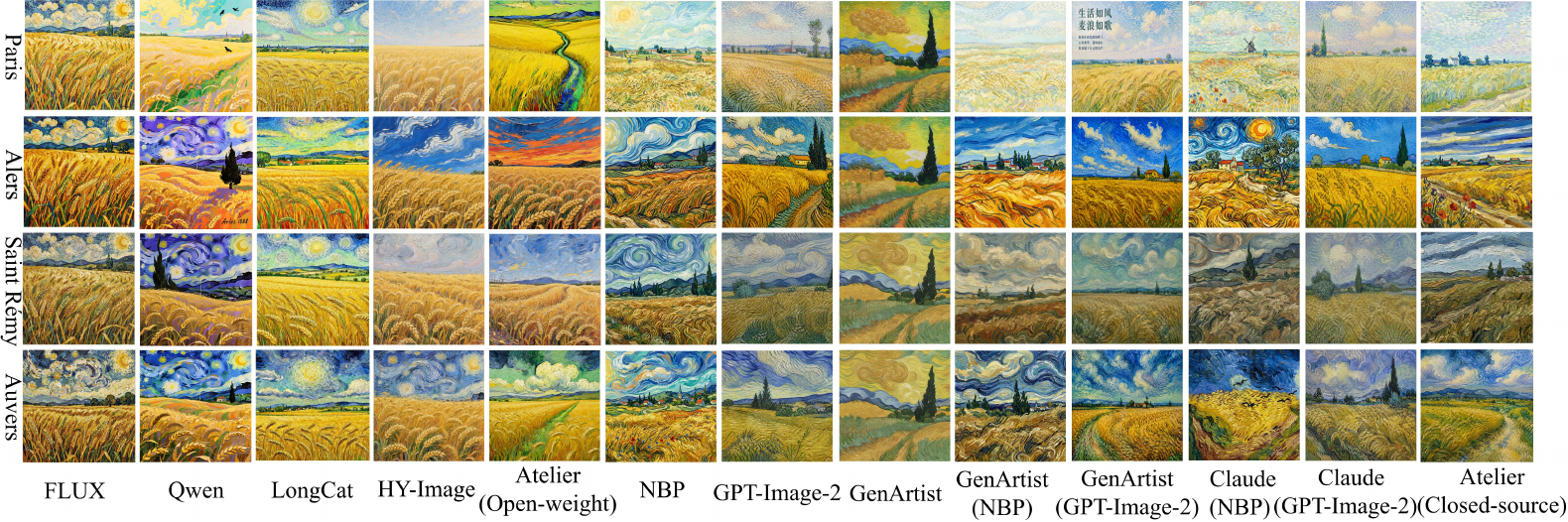}
        \caption{}
    \end{subfigure}

    \vspace{0.4em}

    \begin{subfigure}{\linewidth}
        \centering
        \includegraphics[width=\linewidth]{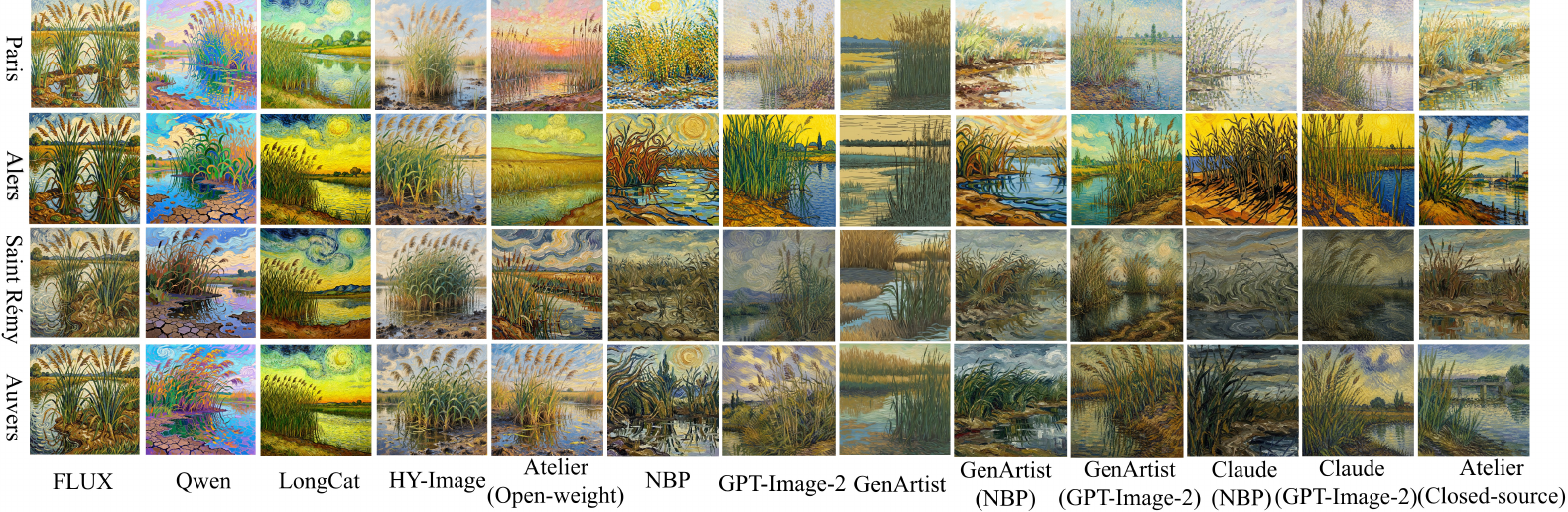}
        \caption{}
    \end{subfigure}
    \caption{Period-controlled generation across Paris, Arles, Saint-R\'emy, and Auvers.}
    \label{fig:exp_qualitative_period}
\end{figure}

\noindent\textbf{Period-controlled generation.}
The left block of Table~\ref{tab:main_results} reports artwork re-rendering, while the four period-controlled columns on the right report IntroStyle $W_2$ across Paris, Arles, Saint-R\'emy, and Auvers. \texttt{Atelier} achieves the lowest style distance in all four periods in both the closed-source and open-weight configurations. Qualitatively, as shown in Figure~\ref{fig:exp_qualitative_period}, direct and agent baselines rely on recognizable period cues that bleed across rows: Claude (NBP) reproduces the storm-cloud composition of \textit{Wheatfield under Thunderclouds} regardless of the target period, while FLUX2 and GPT-Image-2 apply the same swirling strokes and saturated blue-yellow palette throughout. \texttt{Atelier} more consistently varies palette, brushwork, and surface treatment while preserving scene structure and spatial depth.

\begin{figure*}[t]
    \centering
    \includegraphics[width=\textwidth]{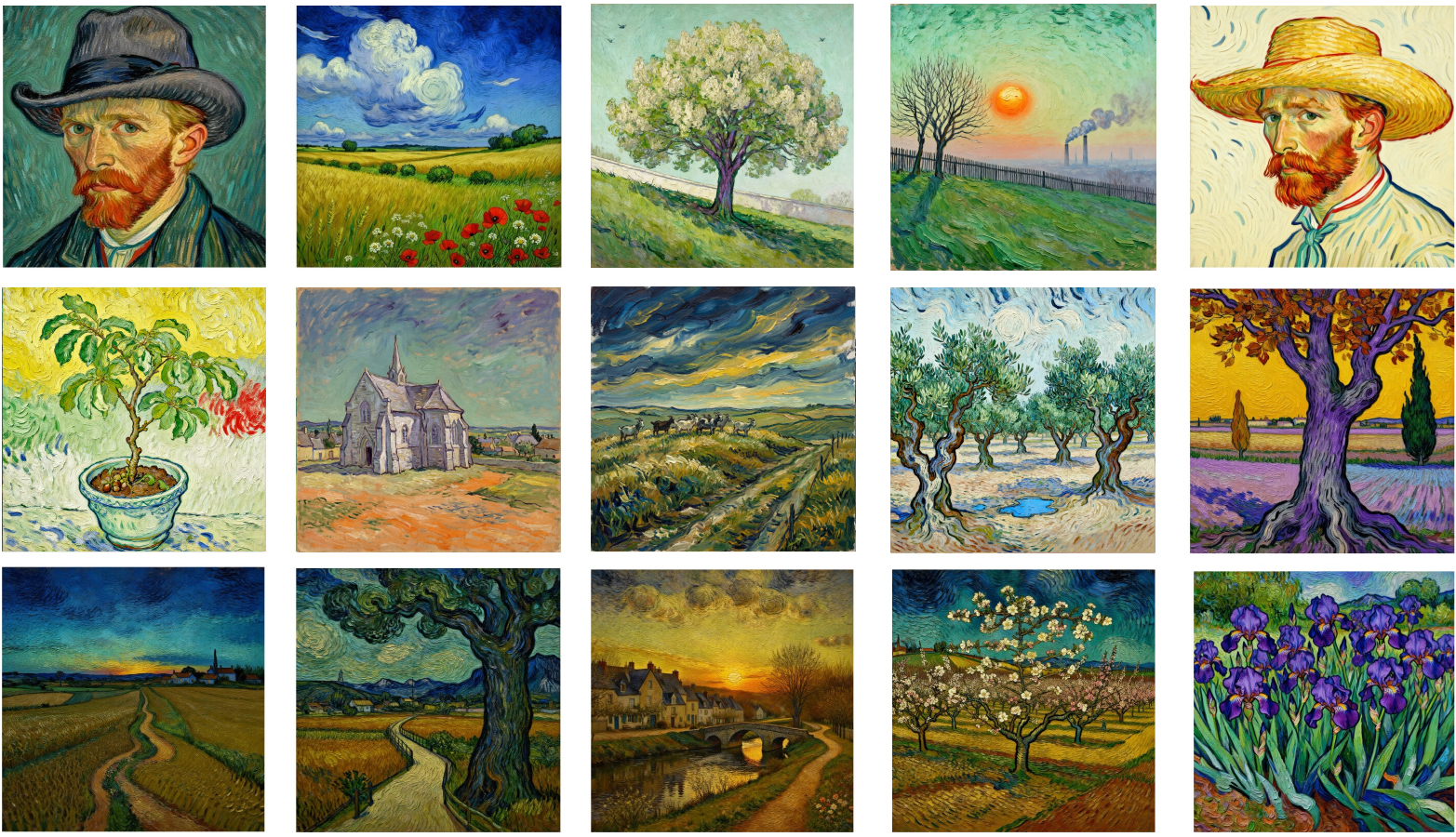}
    \caption{Additional high-resolution Van Gogh-style outputs from \texttt{Atelier}.}
    \label{fig:additional_vangogh}
\end{figure*}

Figure~\ref{fig:additional_vangogh} shows additional high-resolution \texttt{Atelier} outputs on Van Gogh-style prompts. The larger views make it easier to inspect scene preservation, brushwork direction, color organization, and local texture.

\noindent\textbf{Historically Unseen Subject Generation.}
Table~\ref{tab:unseen_metrics} reports IntroStyle $W_2$ and CLIP for 30 modern-subject prompts absent from Van~Gogh's oeuvre. \texttt{Atelier} achieves the lowest $W_2$ in both settings (86.61 closed-source, 91.28 open-weight): the predicted control state preserves Van~Gogh's visual language even for subjects outside his catalog. CLIP scores for \texttt{Atelier} are mid-range in the closed-source setting and lowest in the open-weight setting: agent baselines such as Claude+GPT-Image-2 (0.3559) and LongCat (0.3505) achieve higher text--image alignment because they apply lighter stylization, leaving subject surfaces closer to the prompt's literal description. The tradeoff is inherent: artist-grounded style alters local material and brushwork, which CLIP underweights relative to subject-token matching.

\begin{table}[t]
\centering
\footnotesize
\caption{Quantitative comparison for historically unseen subject generation on 30 modern-subject prompts absent from Van~Gogh's oeuvre. IntroStyle $W_2$ is computed against a reference set of 20 authentic Van~Gogh works, five per period; CLIP measures text--image alignment.}
\label{tab:unseen_metrics}
\setlength{\tabcolsep}{5pt}
\renewcommand{\arraystretch}{1.05}
\begin{tabular}{@{}l c c @{\hspace{14pt}} l c c@{}}
\toprule
\textbf{Method} & \textbf{Closed-source $W_2\downarrow$} & \textbf{CLIP $\uparrow$}
& \textbf{Method} & \textbf{Open-weight $W_2\downarrow$} & \textbf{CLIP $\uparrow$} \\
\midrule
GPT-Image-2              & 89.96 & 0.3505           & FLUX      & 98.49 & 0.3377 \\
Nano Banana Pro          & 90.70 & 0.3418           & Qwen      & 93.56 & 0.3225 \\
GenArtist                & 92.71 & 0.3226           & LongCat   & 102.68 & \textbf{0.3505} \\
GenArtist (NBP)          & 88.21 & 0.3467           & Hunyuan   & 94.36 & 0.3217 \\
GenArtist (GPT-Image-2)  & 91.86 & \underline{0.3511} &           &       &        \\
Claude (NBP)             & \underline{87.65} & 0.3370 &          &       &        \\
Claude (GPT-Image-2)     & 91.34 & \textbf{0.3559} &           &       &        \\
\midrule
\textbf{Atelier (Closed-source)} & \textbf{86.61} & 0.3357
& \textbf{Atelier (Open-weight)} & \textbf{91.28} & 0.3137 \\
\bottomrule
\end{tabular}
\end{table}

\begin{figure}[t]
    \centering
    \includegraphics[width=\linewidth]{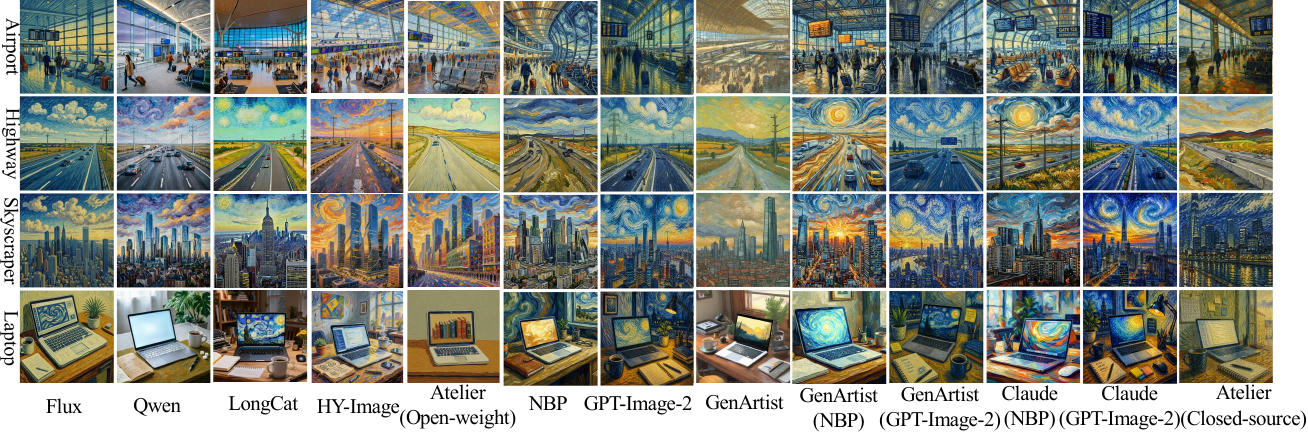}
    \caption{Qualitative comparison for historically unseen subject generation. Given prompts describing modern subjects not present in Van Gogh's oeuvre, \texttt{Atelier} preserves the requested subject identity while applying period-appropriate color, brushwork, and compositional treatment.}
    \label{fig:exp4_historically_unseen_subject_generation}
\end{figure}

Qualitative results are shown in Figure~\ref{fig:exp4_historically_unseen_subject_generation}. Direct baselines apply surface texture without preserving modern subject identity; agent-based baselines improve stylistic grounding but frequently substitute modern elements with period-appropriate equivalents---the highway becomes a rural road, the airport gains arched stone structures. \texttt{Atelier} more consistently preserves the recognizable structure of each modern subject: the airport terminal retains its glass ceiling and crowd depth; the skyscraper row shows urban density without defaulting to Starry Night compositions.

\subsection{Control State Quality}
\label{sec:control_state_eval}

Downstream image quality reflects only the final output. We therefore audit the intermediate control state directly for two \texttt{Atelier} configurations---\texttt{Atelier (Open-weight)} and \texttt{Atelier (Closed-source)}---across four task settings: Van~Gogh artwork re-rendering (from held-out paintings described by structured captions generated by Qwen3-VL-Plus), Van~Gogh period-controlled generation (Paris, Arles, Saint-R\'{e}my, Auvers), historically unseen subjects, and Qi~Baishi artwork re-rendering. Baselines do not expose an intermediate control state, so all numbers below apply only to \texttt{Atelier}.

The audit covers the target period, the preserve/transform decomposition, and the binding of local patch evidence to scene roles. For historically unseen subjects, we additionally verify identity protection and anti-shortcut checks. Table~\ref{tab:control_state_audit} reports the results across the four task settings.

\begin{table*}[t]
\centering
\small
\caption{Structural audit of the intermediate control state across four settings. Each metric is automatic: it verifies that the relevant field is populated and internally consistent, not that the choices are semantically optimal.}
\label{tab:control_state_audit}
\setlength{\tabcolsep}{5pt}
\begin{tabular}{lcc}
\toprule
\textbf{Property} & \textbf{Open-weight} & \textbf{Closed-source} \\
\midrule
\multicolumn{3}{l}{\textit{Panel A: Van Gogh artwork re-rendering}} \\
\quad Source-regime recovery (top-1) $\uparrow$    & 40.0\%  & 38.0\% \\
\quad Source-regime recovery (top-2) $\uparrow$    & 66.0\%  & 62.0\% \\
\quad P/T structural well-formedness $\uparrow$    & 92.0\%  & 100.0\% \\
\quad Patch-backed transform $\uparrow$            & 80.0\%  & 84.0\% \\
\quad Patch soft-fail $\downarrow$                 & 20.0\%  & 16.0\% \\
\midrule
\multicolumn{3}{l}{\textit{Panel B: Van Gogh period-controlled generation}} \\
\quad Source-regime recovery            & 100.0\% & 100.0\% \\
\quad P/T structural well-formedness    & 94.7\%  & 95.6\% \\
\quad Patch-backed transform            & 74.1\%  & 95.6\% \\
\quad Patch soft-fail                   & 8.8\%   & 4.8\% \\
\midrule
\multicolumn{3}{l}{\textit{Panel C: Historically unseen subject (Van Gogh)}} \\
\quad Novel identity guard present      & 96.7\%  & 93.3\% \\
\quad Anti-shortcut pass                & 100.0\% & 96.7\% \\
\quad P/T structural well-formedness    & 96.7\%  & 73.3\% \\
\quad Evidence-backed transform         & 100.0\% & 90.0\% \\
\quad Patch soft-fail                   & 1.3\%   & 3.3\% \\
\midrule
\multicolumn{3}{l}{\textit{Panel D: Qi Baishi artwork re-rendering}} \\
\quad P/T structural well-formedness    & 96.3\%  & 93.8\% \\
\quad Evidence-backed transform         & 45.0\%  & 45.0\% \\
\quad Patch soft-fail                   & 52.5\%  & 52.5\% \\
\quad Patch hard-fail                   & 3.4\%   & 2.5\% \\
\bottomrule
\end{tabular}
\end{table*}

\begin{table*}[!t]
\centering
\small
\caption{Control state quality evaluation on a 0--10 scale, higher is better, scored independently by two LLM judges, DeepSeek V4 Pro and GPT-5.5.}
\label{tab:artistic_signal_audit}
\setlength{\tabcolsep}{5pt}
\renewcommand{\arraystretch}{1.05}
\begin{tabular}{l cc cc}
\toprule
 & \multicolumn{2}{c}{\textbf{Atelier (Open-weight)}} & \multicolumn{2}{c}{\textbf{Atelier (Closed-source)}} \\
\cmidrule(lr){2-3}\cmidrule(lr){4-5}
& DeepSeek & GPT-5.5 & DeepSeek & GPT-5.5 \\
\midrule
\multicolumn{5}{l}{\textit{Panel A: Artwork re-rendering (Van Gogh)}} \\
\quad Source retention      & 8.50 & 8.81 & 8.42 & 8.72 \\
\quad Preserve/transform    & 6.69 & 7.24 & 6.90 & 7.65 \\
\quad Period-style adapt.   & 7.83 & 7.99 & 7.70 & 7.90 \\
\quad Patch evidence util.  & 4.79 & 5.68 & 5.00 & 6.01 \\
\quad Composite             & 6.95 & 7.43 & 7.00 & 7.57 \\
\quad Inter-judge $r$       & \multicolumn{2}{c}{0.59} & \multicolumn{2}{c}{0.66} \\
\midrule
\multicolumn{5}{l}{\textit{Panel B: Period-controlled generation}} \\
\quad Palette match         & 7.66 & 5.83 & 7.62 & 5.70 \\
\quad Brushwork match       & 8.63 & 6.77 & 8.61 & 6.58 \\
\quad Composition match     & 7.36 & 5.02 & 7.33 & 4.83 \\
\quad Motif match           & 7.37 & 4.87 & 7.36 & 4.72 \\
\cmidrule(lr){2-5}
\quad Paris                 & 7.39 & 4.73 & 7.20 & 4.55 \\
\quad Arles                 & 7.82 & 5.92 & 7.74 & 5.72 \\
\quad Saint-R\'{e}my        & 8.21 & 5.88 & 8.22 & 5.66 \\
\quad Auvers                & 7.58 & 5.98 & 7.75 & 5.89 \\
\quad Inter-judge $r$       & \multicolumn{2}{c}{0.43} & \multicolumn{2}{c}{0.49} \\
\midrule
\multicolumn{5}{l}{\textit{Panel C: Historically unseen subject (Van Gogh)}} \\
\quad Novel content coverage       & 9.18 & 7.86 & 9.00 & 8.07 \\
\quad P/T decomposition            & 9.27 & 8.14 & 8.55 & 6.80 \\
\quad Period-style translation     & 9.00 & 8.20 & 8.96 & 7.98 \\
\quad Shortcut avoidance           & 9.91 & 8.99 & 9.41 & 7.71 \\
\quad Composite                    & 9.34 & 8.30 & 8.98 & 7.64 \\
\quad Substitution: none / minor / major & 30/0/0 & 30/0/0 & 30/0/0 & 28/2/0 \\
\midrule
\multicolumn{5}{l}{\textit{Panel D: Artwork re-rendering (Qi Baishi)}} \\
\quad Palette match         & 7.18 & 5.72 & 7.96 & 6.28 \\
\quad Brushwork match       & 7.60 & 6.34 & 8.22 & 7.04 \\
\quad Composition match     & 5.28 & 3.54 & 5.62 & 5.36 \\
\quad Motif match           & 4.34 & 3.16& 4.78 & 4.54 \\
\quad Composite             & 6.10 & 4.69 & 6.65 & 5.81 \\
\bottomrule
\end{tabular}
\end{table*}

\noindent\textbf{Source-regime recovery.} On period-controlled prompts, the target period is stated in the surface prompt and the control state must simply propagate that intent without overwriting it; agreement with the label is complete in both configurations. On artwork re-rendering, the target period is not explicit: the input is a Qwen-generated caption of a source painting, and the perceiver must recover the source painting's period from stylistic and content cues in the caption. Under this harder condition, the perceiver's top-1 prediction matches the label in 40.0\% of cases (open-weight) and 38.0\% (closed-source); top-2 recovery reaches 66.0\% and 62.0\% respectively. When the period is provided as structured input, agreement is perfect; caption-based inference is harder, but the correct period is within the top two for the majority of cases.

\noindent\textbf{Preserve/transform decomposition.} The control state records, for each prompt, which scene elements the model commits to keeping stable and which it marks for artistic translation. An entry is counted as structurally well-formed if the preserve and transform lists are both non-empty and disjoint and each transform entry cites at least one translation axis such as palette, contour pressure, or brush rhythm. Under period-controlled generation, 94.7\% (open-weight) and 95.6\% (closed-source) of cases satisfy this constraint; under re-rendering, well-formedness reaches 92.0\% and 100.0\% respectively. Structural well-formedness verifies field-level consistency only; the semantic quality of preserve/transform choices is evaluated separately by the LLM judges in Table~\ref{tab:artistic_signal_audit}.

\noindent\textbf{Evidence binding.} For each transform entry, the pipeline attempts to bind a role-specific patch family from the artist knowledge base; if no specific match is available, it may fall back to a generic global patch (a soft fail). Under period-controlled generation, 74.1\% (open-weight) and 95.6\% (closed-source) of transform entries are backed by a role-specific patch, with soft-fail rates of 8.8\% and 4.8\% respectively. Under re-rendering, patch-backed rates are 80.0\% and 84.0\%, and soft-fail rates are 20.0\% and 16.0\%. These figures indicate that retrieval reliability is bounded by knowledge-base coverage rather than by the binding mechanism itself; the higher soft-fail rate under re-rendering reflects the wider distribution of source-painting motifs relative to the period-labeled prompt set.

\noindent\textbf{Historically unseen subjects.} Panel~C reports structural audit results on 30 prompts requesting modern subjects absent from Van~Gogh's oeuvre (e.g., skyscrapers, laptops, airports). Two indicators are specific to this task. \emph{Novel identity guard present} checks whether the core modern subject extracted from the prompt appears in any of the control state's identity-protection fields (preserve list, structural anchors, or forbidden-substitution constraints); the guard is present in 96.7\% (open-weight) and 93.3\% (closed-source) of cases, confirming that the perceiver explicitly protects modern subject identity rather than treating unseen prompts as generic scenes. \emph{Anti-shortcut pass} checks whether the frozen Van~Gogh shortcut taxonomy (Table~\ref{tab:shortcut-taxonomy}) is triggered without being requested by the prompt; 100\% of open-weight cases and 96.7\% of closed-source cases pass. The generic structural metrics remain strong (patch-backed transform rates of 100\% and 90.0\%, soft-fail rates of 1.3\% and 3.3\%), though P/T structural well-formedness drops to 73.3\% under the closed-source configuration, pointing to inconsistent translation-axis specification on unseen prompts as a residual limitation.

\noindent\textbf{Cross-artist transfer to Qi Baishi.} Panel~D reports structural audit results on Qi~Baishi artwork re-rendering. We omit source-regime recovery because Qi~Baishi's oeuvre does not decompose into discrete period regimes analogous to Van~Gogh's Paris/Arles/Saint-R\'emy/Auvers. P/T structural well-formedness remains strong (96.3\% open-weight, 93.8\% closed-source), showing that the perceiver's decomposition logic transfers cleanly from Van~Gogh to Qi~Baishi. However, evidence-backed transform drops to 45.0\% in both configurations, with over half of transform entries falling back to a generic patch (soft-fail 52.5\%) rather than terminating unbound (hard-fail 3.4\%/2.5\%). This gap relative to Van~Gogh (Panel~A: 80--84\% patch-backed) is a knowledge-base coverage issue: the Qi~Baishi patch bank is smaller and more sparsely annotated than the Van~Gogh bank, so more transforms resolve through fallback. Because hard-fail remains low, the binding mechanism itself functions correctly; the ceiling is set by patch-bank expansion, not by the perceiver.

\noindent\textbf{Artistic control signal audit.} The structural metrics above verify that Atelier's control state is well-formed but not that the artistic decisions it encodes are appropriate for the target Van~Gogh period. To evaluate this directly, we audit the control state with two independent LLM judges: DeepSeek V4 Pro~\citep{deepseekai2026deepseekv4} and GPT-5.5~\citep{openai2025gpt55}. Because our benchmark contains two distinct tasks with different semantics---period-controlled generation, where the system must select period-appropriate content, and artwork re-rendering, where the system must preserve source content while translating only style---we use two task-matched judge protocols. For \textbf{period-controlled generation}, the judge scores each case on the four artist-specific axes highlighted in Section~\ref{sec:intro}: palette match, brushwork match, composition match, and motif match (0--10 scale). For \textbf{artwork re-rendering}, where source content must be preserved rather than replaced by canonical Van~Gogh motifs, we use a source-preserving protocol with four 0--10 dimensions: source retention, preserve/transform policy, period-style adaptation, and patch-evidence usefulness, plus a shortcut-substitution flag (\emph{none}/\emph{minor}/\emph{major}). No fixed rubric or keyword list is supplied to either judge; both apply their own art-historical judgment case by case.

Table~\ref{tab:artistic_signal_audit} reports results for \texttt{Atelier (Open-weight)} and \texttt{Atelier (Closed-source)} under both protocols. On Van~Gogh artwork re-rendering (Panel~A), both judges agree that source retention is the strongest dimension ($\geq 8.4$ across all cells), confirming that the control state faithfully preserves source content; patch-evidence usefulness is the weakest dimension ($4.79$--$6.01$), pointing to knowledge-base coverage as the primary limitation of the current pipeline. Only $2$ of the $196$ re-rendering cases are flagged with \emph{major} shortcut substitution, indicating that canonical-motif replacement is not a systemic failure mode.

On period-controlled generation (Panel~B), both judges agree that brushwork is the strongest dimension and motif/composition are the weakest; on the period axis, both rank Paris as the weakest target, consistent with the smaller number of Paris-period patches in the current knowledge base.

On historically unseen subjects (Panel~C), users request modern subjects absent from Van~Gogh's oeuvre (e.g., skyscrapers, laptops, airports). The judges use a task-specific rubric with four 0--10 dimensions (novel content coverage, preserve/transform decomposition, period-style translation, and shortcut avoidance) plus a categorical shortcut-substitution flag. Composite scores are the highest across all four panels (open-weight $8.30$--$9.34$, closed-source $7.64$--$8.98$), and the shortcut flag is \emph{none} in $118/120$ (${\sim}98\%$) case-judge assessments, with the remaining two flagged \emph{minor} and none flagged \emph{major}. The control state thus retains modern subject identity while translating style, avoiding canonical-motif replacement even for content Van~Gogh never painted.

On Qi~Baishi artwork re-rendering (Panel~D), we reuse the Panel~B dimensions (palette, brushwork, composition, motif) interpreted with respect to Qi~Baishi's stylistic conventions, since Qi~Baishi's oeuvre lacks discrete period labels analogous to Van~Gogh's. Composite scores are lower than under Van~Gogh re-rendering (open-weight $5.81$--$6.65$, closed-source $4.69$--$6.10$), driven by motif and composition; both judges agree that palette and brushwork transfer are the strongest dimensions, mirroring the Van~Gogh pattern.

For the two panels reporting inter-judge Pearson $r$, agreement ranges from $0.43$ to $0.66$; higher agreement on re-rendering reflects the more concrete, source-referenced criteria of the source-preserving protocol.

\noindent\textbf{Scope and limitations.} The structural audit is automatic: it verifies that the relevant field is populated and internally consistent, not that its content is semantically optimal. The artistic signal audit relies on LLM judges applying their own art-historical knowledge without a fixed rubric; absolute scores differ between judges, and we therefore emphasize convergent findings rather than any single score. The Qi~Baishi re-rendering panel reuses the palette/brushwork/composition/motif dimensions rather than the source-preserving protocol because Qi~Baishi's oeuvre lacks the discrete period structure that motivates period-style adaptation; developing a source-preserving protocol for ink-and-color painting remains an open direction.

\subsection{Shortcut Audit}
\label{sec:shortcut_audit}

\noindent\textbf{Setup.}
Using the frozen taxonomy introduced in Section~\ref{sec:intro} (Table~\ref{tab:shortcut-taxonomy}), we audit how often each method produces unrequested canonical shortcuts on Van~Gogh and Qi~Baishi prompts. We use 358 Van~Gogh prompts and 80 Qi~Baishi prompts per method, all designed to avoid explicit period or motif specification. Each generated image is independently labelled by three human annotators for each of the four shortcut patterns, and we take the majority vote (${\geq}2/3$) as the final judgement per pattern. The Shortcut Substitution Rate (SSR) is the proportion of images containing at least one shortcut pattern from the corresponding artist's taxonomy.

\noindent\textbf{Van Gogh.}
Table~\ref{tab:vg_merged} reports both SSR and per-pattern rates on Van~Gogh prompts. Baselines exhibit high shortcut rates: closed-source SSR ranges from 59.22\% to 77.37\%, and open-weight direct baselines span 48.04\%--56.98\%, with GenArtist reaching the lowest baseline SSR at 47.21\%. The highest closed-source SSR (77.37\%) comes from an agent wrapper, Claude (NBP); such wrappers inherit the underlying generator's canonical bias rather than mitigating it. \texttt{Atelier} reduces SSR to 51.96\% (closed-source) and 31.56\% (open-weight), improving over the strongest baseline in each category by 7.3 and 15.7 percentage points respectively. Heavy impasto remains the dominant failure mode across baselines, while cypress insertion peaks under closed-source systems (GPT-Image-2 33.80\%). \texttt{Atelier} attains the best rate in every column except two: for closed-source cypresses, GenArtist (NBP) is marginally lower (11.73\% vs.\ 12.85\%), and for open-weight blue-yellow palette, Hunyuan is marginally lower (4.75\% vs.\ 6.98\%).

The larger residual SSR under the closed-source configuration reflects a limitation of the underlying generation model rather than of the control state itself. \texttt{Atelier} produces the same intermediate control state regardless of downstream backend, so the gap between open-weight (31.56\%) and closed-source (51.96\%) SSR indicates that closed-source models (GPT-Image-2, Nano Banana Pro) carry stronger pre-trained canonical priors that partially override the explicit avoidance constraints encoded in the control state. These backends also expose no dedicated negative-prompt channel; shortcut constraints therefore fall back to imperative statements inside the positive prompt, consistent with the SSR gap observed here. Structured control is a necessary but not sufficient condition for shortcut reduction: the downstream generator must also remain receptive to fine-grained control signals, and models with strong canonical priors can overwrite explicit constraints even when those constraints are specified.

\begin{table*}[t]
\centering
\footnotesize
\caption{Shortcut audit on Van~Gogh prompts. SSR (Shortcut Substitution Rate) is the proportion of images containing at least one shortcut. Per-pattern columns report the proportion of all 358 images per method exhibiting each pattern; rows do not sum to SSR because a single image can violate multiple patterns. Best result in each configuration is shown in bold; second-best per-pattern values are underlined.}
\label{tab:vg_merged}
\vspace{4pt}
\setlength{\tabcolsep}{4pt}
\renewcommand{\arraystretch}{1.05}
\begin{tabular}{@{}lccc cccc@{}}
\toprule
\textbf{Method}
& \textbf{Images}
& \textbf{$\geq$1 Viol.\ $\downarrow$}
& \textbf{SSR $\downarrow$}
& \textbf{Starry Night sky $\downarrow$}
& \textbf{Blue-yellow palette $\downarrow$}
& \textbf{Heavy impasto $\downarrow$}
& \textbf{Cypresses $\downarrow$} \\
\midrule
\multicolumn{8}{c}{\textit{Closed-source configuration}} \\
\midrule
Nano Banana Pro (NBP)    & 358 & 249 & 69.55\% & 19.83\%             & 15.08\%             & 31.84\%             & 19.83\% \\
GPT-Image-2              & 358 & 275 & 76.82\% & 16.20\%             & \underline{13.97\%} & 30.45\%             & 33.80\% \\
GenArtist (NBP)          & 358 & \underline{212} & \underline{59.22\%} & 17.32\%             & 14.25\%             & 30.73\%             & \textbf{11.73\%} \\
GenArtist (GPT-Image-2)  & 358 & 254 & 70.95\% & \underline{14.25\%} & 16.20\%             & 31.84\%             & 26.26\% \\
Claude (GPT-Image-2)     & 358 & 271 & 75.70\% & 15.08\%             & 24.02\%             & \underline{29.05\%} & 25.98\% \\
Claude (NBP)             & 358 & 277 & 77.37\% & 22.63\%             & 18.72\%             & 30.73\%             & 23.18\% \\
\midrule
Atelier (Closed-source)        & 358 & \textbf{186} & \textbf{51.96\%} & \textbf{10.61\%} & \textbf{10.34\%} & \textbf{27.93\%} & \underline{12.85\%} \\
\midrule
\multicolumn{8}{c}{\textit{Open-weight configuration}} \\
\midrule
FLUX        & 358 & 204 & 56.98\%         & 27.09\%             & 8.38\%              & \underline{24.58\%} & 9.22\% \\
Qwen        & 358 & 185 & 51.68\%         & \underline{13.97\%} & 12.57\%             & 31.84\%             & \underline{7.54\%} \\
LongCat     & 358 & 194 & 54.19\%         & 19.55\%             & 8.38\%              & 30.73\%             & 8.10\% \\
Hunyuan     & 358 & 172 & 48.04\%         & 17.88\%             & \textbf{4.75\%}     & 30.17\%             & 8.66\% \\
GenArtist   & 358 & \underline{169} & \underline{47.21\%}         & 14.53\%             & 8.66\%              & 31.84\%             & 7.82\% \\
\midrule
Atelier (Open-weight)          & 358 & \textbf{113} & \textbf{31.56\%} & \textbf{8.10\%} & \underline{6.98\%} & \textbf{17.60\%} & \textbf{4.19\%} \\
\bottomrule
\end{tabular}
\end{table*}

\noindent\textbf{Qi Baishi.}
Table~\ref{tab:qb_merged} reports both SSR and per-pattern rates on Qi~Baishi prompts. Baselines exhibit high shortcut rates: closed-source SSR ranges from 55.00\% to 78.75\%, and open-weight direct baselines span 40.00\%--53.75\%. The highest closed-source SSR (78.75\%) again comes from an agent wrapper, Claude (NBP), reinforcing the Van~Gogh observation that prompt-expansion wrappers do not mitigate canonical bias. \texttt{Atelier} reduces SSR to 43.75\% (closed-source) and 28.75\% (open-weight), improving over the strongest baseline in each category by 11.25 percentage points. Generic ink-painting backgrounds remain a common failure mode across closed-source baselines: Claude (NBP) reaches 56.25\% and Nano Banana Pro 51.25\%. In contrast, GPT-Image-2 and its wrappers exhibit high rates of unrequested calligraphic inscription (up to 53.75\%). \texttt{Atelier} attains the best rate on every open-weight pattern and on two of three closed-source patterns (calligraphic inscription 21.25\%, decorative substitution 15.00\%); for generic ink backgrounds, GenArtist (GPT-Image-2) is marginally lower (21.25\% vs.\ 25.00\%).

\begin{table*}[t]
\centering
\footnotesize
\caption{Shortcut audit on Qi~Baishi prompts. SSR (Shortcut Substitution Rate) is the proportion of images containing at least one shortcut. Per-pattern columns report the proportion of all 80 images per method exhibiting each pattern: generic ink-painting backgrounds added without request; calligraphic inscription placed without request; Qi~Baishi's ink-wash language replaced by generic Chinese decorative illustration. Rows do not sum to SSR because a single image can violate multiple patterns. Best result in each configuration is shown in bold; second-best per-pattern values are underlined. Lower is better throughout.}
\label{tab:qb_merged}
\vspace{4pt}
\setlength{\tabcolsep}{4pt}
\renewcommand{\arraystretch}{1.05}
\begin{tabular}{@{}lccc ccc@{}}
\toprule
\textbf{Method}
& \textbf{Images}
& \textbf{$\geq$1 Viol.\ $\downarrow$}
& \textbf{SSR $\downarrow$}
& \textbf{Generic ink BG $\downarrow$}
& \textbf{Calligraphic insc.\ $\downarrow$}
& \textbf{Decorative subst.\ $\downarrow$} \\
\midrule
\multicolumn{7}{c}{\textit{Closed-source configuration}} \\
\midrule
Nano Banana Pro (NBP)    & 80 & 55 & 68.75\% & 51.25\%             & 25.00\%             & \underline{10.00\%} \\
GPT-Image-2              & 80 & 57 & 71.25\% & \underline{25.00\%} & 53.75\%             & 16.25\% \\
GenArtist (NBP)          & 80 & \underline{44} & \underline{55.00\%} & 35.00\%             & \underline{22.50\%} & 16.25\% \\
GenArtist (GPT-Image-2)  & 80 & 51 & 63.75\% & \textbf{21.25\%}    & 45.00\%             & 22.50\% \\
Claude (GPT-Image-2)     & 80 & 55 & 68.75\% & 35.00\%             & 41.25\%             & 17.50\% \\
Claude (NBP)             & 80 & 63 & 78.75\% & 56.25\%             & \underline{22.50\%} & 16.25\% \\
\midrule
Atelier (Closed-source)        & 80 & \textbf{35} & \textbf{43.75\%} & \underline{25.00\%} & \textbf{21.25\%} & \textbf{15.00\%} \\
\midrule
\multicolumn{7}{c}{\textit{Open-weight configuration}} \\
\midrule
FLUX        & 80 & 43 & 53.75\%          & 22.50\%             & \underline{20.00\%} & 26.25\% \\
Qwen        & 80 & 39 & 48.75\%          & 20.00\%             & 32.50\%             & \underline{13.75\%} \\
LongCat     & 80 & 39 & 48.75\%          & 27.50\%             & 25.00\%             & \underline{13.75\%} \\
Hunyuan     & 80 & 32 & 40.00\%          & 22.50\%             & 25.00\%             & \underline{13.75\%} \\
GenArtist   & 80 & 32 & 40.00\%          & \underline{18.75\%} & 28.75\%             & 15.00\% \\
\midrule
Atelier (Open-weight)          & 80 & \textbf{23} & \textbf{28.75\%} & \textbf{13.75\%} & \textbf{11.25\%} & \textbf{11.25\%} \\
\bottomrule
\end{tabular}
\end{table*}

\noindent\textbf{Cross-artist discussion.} The audit confirms that shortcutting is not a corner case but a systematic failure of current generators: across both artists, every direct baseline and most agent-wrapped baselines produce unrequested canonical patterns in a large fraction of their outputs, with several Qi~Baishi methods exceeding 55\% and the worst reaching 78.75\%. By contrast, \texttt{Atelier} keeps SSR at 31.56\%--51.96\% on Van~Gogh and 28.75\%--43.75\% on Qi~Baishi, improving over the strongest baseline in every configuration. This consistent reduction indicates that explicit shortcut-avoidance constraints in the control state generalize beyond any single backend or artist.

\subsection{Cross-Artist Transfer to Qi Baishi}

Table~\ref{tab:qibaishi_metrics} reports quantitative results for cross-artist transfer to Qi~Baishi. \texttt{Atelier} achieves the lowest IntroStyle $W_2$ in both settings (closed-source: 57.09, open-weight: 68.33). For perceptual fidelity metrics, \texttt{Atelier} achieves the best DINO-cosine and LPIPS in both settings, indicating stronger capture of semantic structure and patch-level perceptual similarity. DreamSim is led by Claude (NBP) in the closed-source setting and Hunyuan in the open-weight setting. It shows that baselines struggle with the ink-wash medium: FLUX applies Western impasto throughout, while Nano and Qwen recover ink-wash character only inconsistently---subject identity is preserved but medium character is not. \texttt{Atelier} more consistently captures Qi Baishi's sparse composition, calligraphic placement, restrained color economy, and motif-centered brushwork across flowers, frogs, chrysanthemums, and willows.

\begin{table}[t]
\centering
\footnotesize
\caption{Quantitative results for Qi~Baishi cross-artist transfer. For each setting we report IntroStyle $W_2$ for style proximity, CLIP for text--image alignment, and DreamSim, DINO-cosine, LPIPS for perceptual fidelity to source paintings (DS/DINO/LP). Best and second-best within each configuration are shown in bold and underlined, respectively.}
\label{tab:qibaishi_metrics}
\setlength{\tabcolsep}{3.5pt}
\renewcommand{\arraystretch}{1.05}
\resizebox{\linewidth}{!}{%
\begin{tabular}{@{}l cc ccc @{\hspace{8pt}} l cc ccc@{}}
\toprule
\textbf{Method} & \textbf{$W_2\downarrow$} & \textbf{CLIP$\uparrow$} & \textbf{DS$\downarrow$} & \textbf{DINO$\uparrow$} & \textbf{LP$\downarrow$}
& \textbf{Method} & \textbf{$W_2\downarrow$} & \textbf{CLIP$\uparrow$} & \textbf{DS$\downarrow$} & \textbf{DINO$\uparrow$} & \textbf{LP$\downarrow$} \\
\midrule
\multicolumn{6}{c}{\textit{Closed-source}} & \multicolumn{6}{c}{\textit{Open-weight}} \\
\midrule
GPT-Image-2 & 59.71 & 0.3216 & 0.3653 & 0.4864 & 0.5998 & FLUX & 79.73 & 0.3125 & 0.4425 & 0.4427 & 0.6392 \\
Nano Banana Pro & 63.50 & 0.3119 & 0.3629 & 0.5083 & 0.6129 & Qwen & 83.33 & 0.3042 & 0.4431 & 0.4561 & 0.6353 \\
GenArtist (GPT) & \underline{57.98} & \underline{0.3317} & 0.3552 & 0.5097 & 0.6008 & LongCat & 75.42 & \textbf{0.3244} & 0.4280 & \underline{0.4631} & \underline{0.6248} \\
GenArtist (NBP) & 60.44 & 0.3221 & \underline{0.3429} & \underline{0.5274} & 0.5980 & Hunyuan & \underline{73.58} & 0.3185 & \underline{0.4092} & 0.4507 & 0.6278 \\
Claude (GPT) & 58.46 & \textbf{0.3330} & 0.3439 & 0.5110 & \underline{0.5943} & GenArtist & 78.42 & 0.3171 & 0.4486 & 0.4158 & 0.6618 \\
Claude (NBP) & 61.09 & 0.3219 & \textbf{0.3381} & 0.5274 & 0.6005 & & & & & & \\
\midrule
\textbf{Atelier (Closed-source)} & \textbf{57.09} & 0.3194 & \underline{0.3397} & \textbf{0.5452} & \textbf{0.5804}
& \textbf{Atelier (Open-weight)} & \textbf{68.33} & \underline{0.3193} & \underline{0.4133} & \textbf{0.4731} & \textbf{0.6154} \\
\bottomrule
\end{tabular}%
}
\end{table}

\begin{figure}[t]
    \centering
    \includegraphics[width=\linewidth]{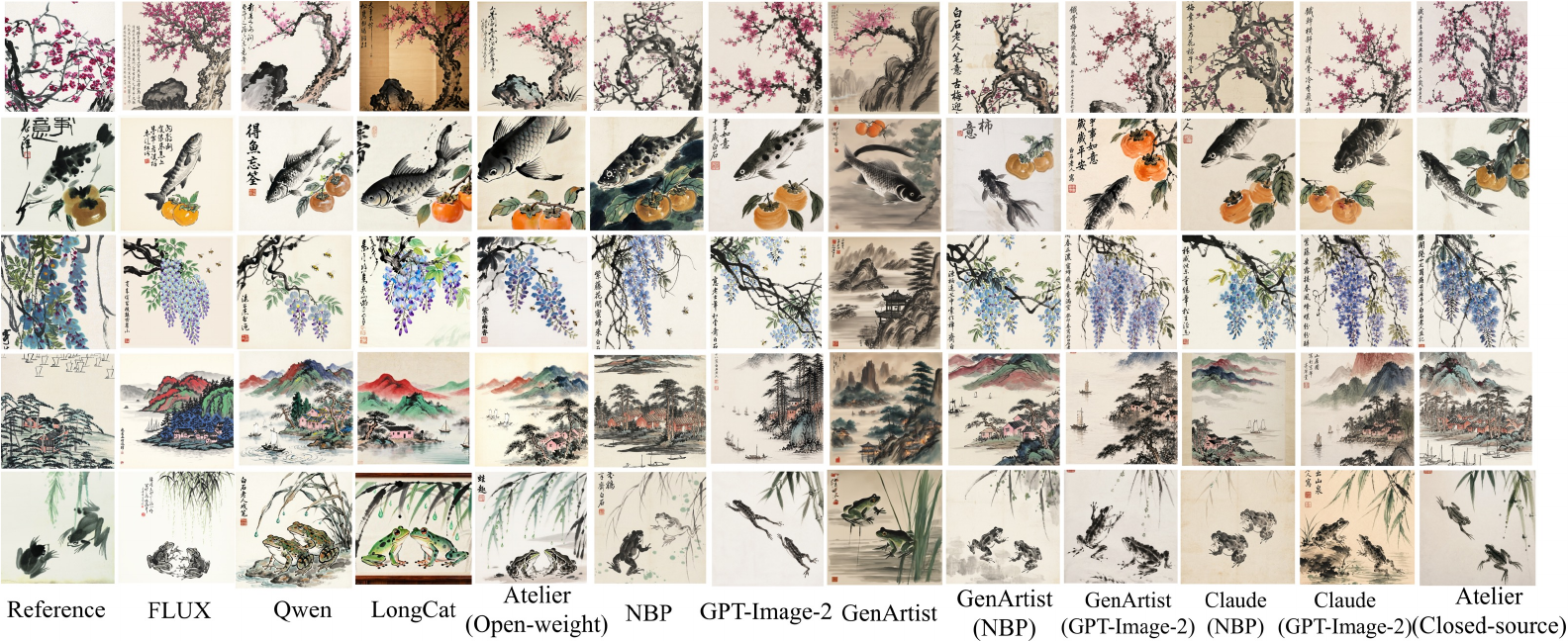}
    \caption{Cross-artist transfer to Qi Baishi. Given the same input scene, \texttt{Atelier} adapts artist-specific resources to translate the content into Qi Baishi's visual language, including sparse composition, calligraphic brushwork, restrained ink-color rendering, and motif-centered expression.}
    \label{fig:exp5_qibaishi_transfer2}
\end{figure}

\begin{figure*}[t]
    \centering
    \includegraphics[width=\textwidth]{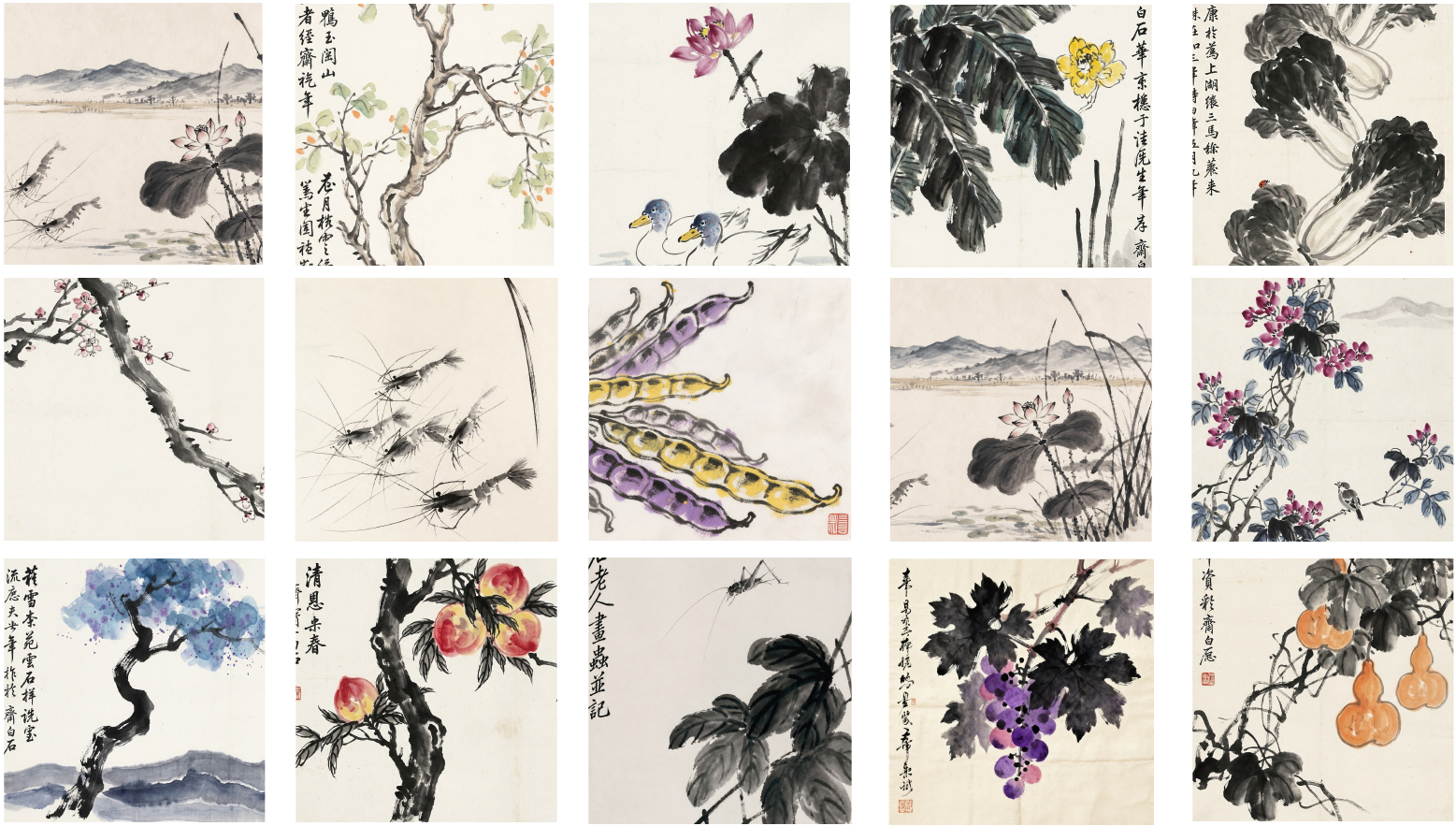}
    \caption{Additional high-resolution Qi~Baishi-style outputs from \texttt{Atelier}.}
    \label{fig:additional_qibaishi}
\end{figure*}

Figure~\ref{fig:additional_qibaishi} shows additional high-resolution \texttt{Atelier} outputs on Qi~Baishi-style prompts. The larger views make it easier to inspect sparse composition, calligraphic brush placement, restrained color use, and subject preservation across different motifs.

\begin{figure}[t]
    \centering
    \includegraphics[width=1\linewidth]{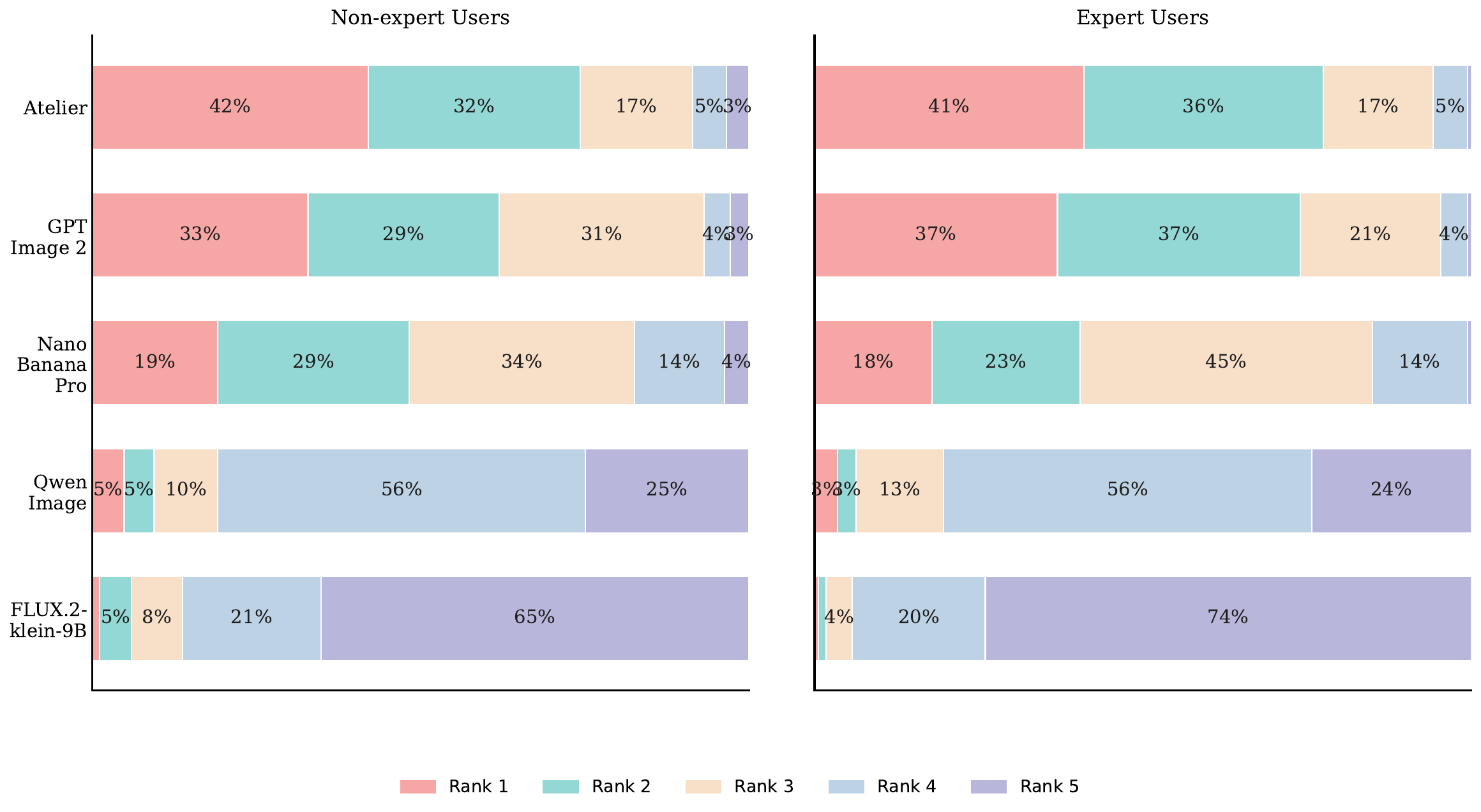}
    \caption{User study: rank distribution across methods.}
    \label{fig:user_study}
\end{figure}

We conducted a blinded user study with 10 non-expert and 5 expert participants, each ranking outputs from five methods across three tasks. Bradley--Terry analysis~\citep{bradley1952rank}, widely used for aggregating pairwise human preferences \citep{chiang2024chatbotarena}, shows that preferences for Atelier and GPT-Image-2 over the three remaining methods are consistent across both rater groups, with pairwise preference probabilities exceeding 0.87 in all comparisons against Qwen and FLUX. As shown in Figure~\ref{fig:user_study}, \texttt{Atelier} receives the highest Rank-1 preference rate among both non-experts and experts, at 42\% and 41\% respectively, and the lowest Rank-5 rate among both groups, at 3\% and 1\% respectively. GPT-Image-2 ranks second among both groups; FLUX receives the most last-place rankings (65\% from non-experts, 74\% from experts). The near-identical preference profiles of the two groups suggest that the qualities separating \texttt{Atelier} from the baselines are visible without formal training, while the experts' harsher treatment of the weakest outputs indicates that their agreement does not reflect insensitivity to artistic detail. 

The blinded user study was conducted via an online questionnaire. Each participant completed ranking questions covering three tasks:

\begin{itemize}[nosep,leftmargin=*]
\item \textbf{Period-controlled generation}: five subjects (self-portrait, sheep pasture, sunflower vase, wheat field, wooden drawbridge) across four Van Gogh periods (Paris, Arles, Saint-R\'emy, Auvers).
\item \textbf{Artwork re-rendering}: matching the source painting in style, brushwork, and artistic character.
\item \textbf{Historically unseen subjects}: modern subjects (laptop, train station, highway, skyscraper, nighttime city) rendered in Van Gogh's style.
\end{itemize}

\begin{figure}[t]
    \centering
    \includegraphics[width=\linewidth]{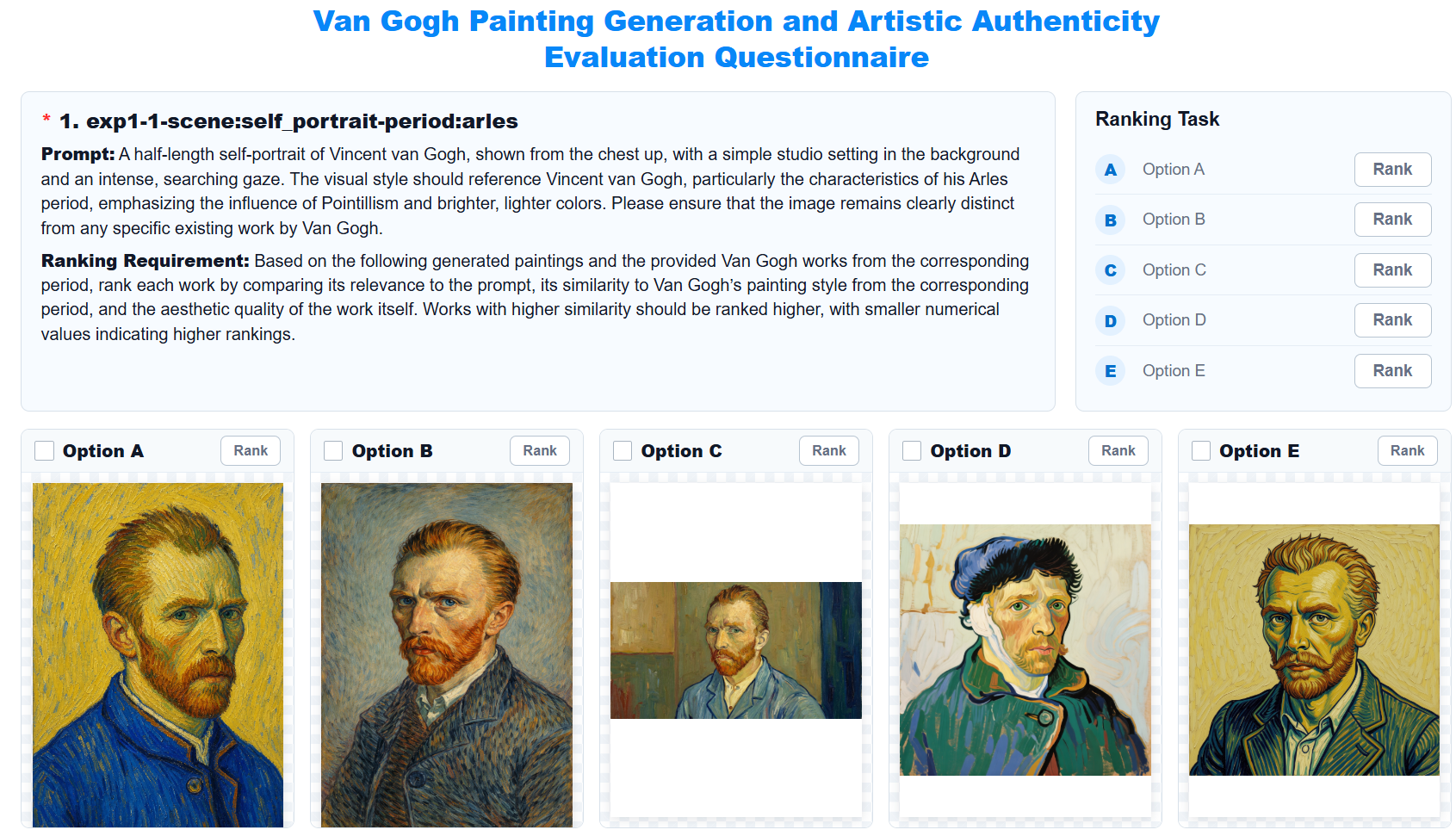}
    \caption{User study interface. Each question shows five anonymized outputs and asks participants for a full ranking.}
    \label{fig:user_study_interface}
\end{figure}

Each question shows five anonymized outputs (labeled A--E) along with the prompt; for re-rendering, the reference painting is also shown. All method identities are hidden: outputs are presented as A--E only, and labels are independently randomized for each question to prevent ordering effects. The questions span the three tasks above to ensure that participants evaluate \texttt{Atelier} across diverse generation scenarios---period-specific stylization, reference-grounded re-rendering, and modern-subject preservation---rather than a single setting. Participants rank outputs from 1 (best) to 5 (worst).

\begin{figure}[t]
    \centering
    \includegraphics[width=\linewidth]{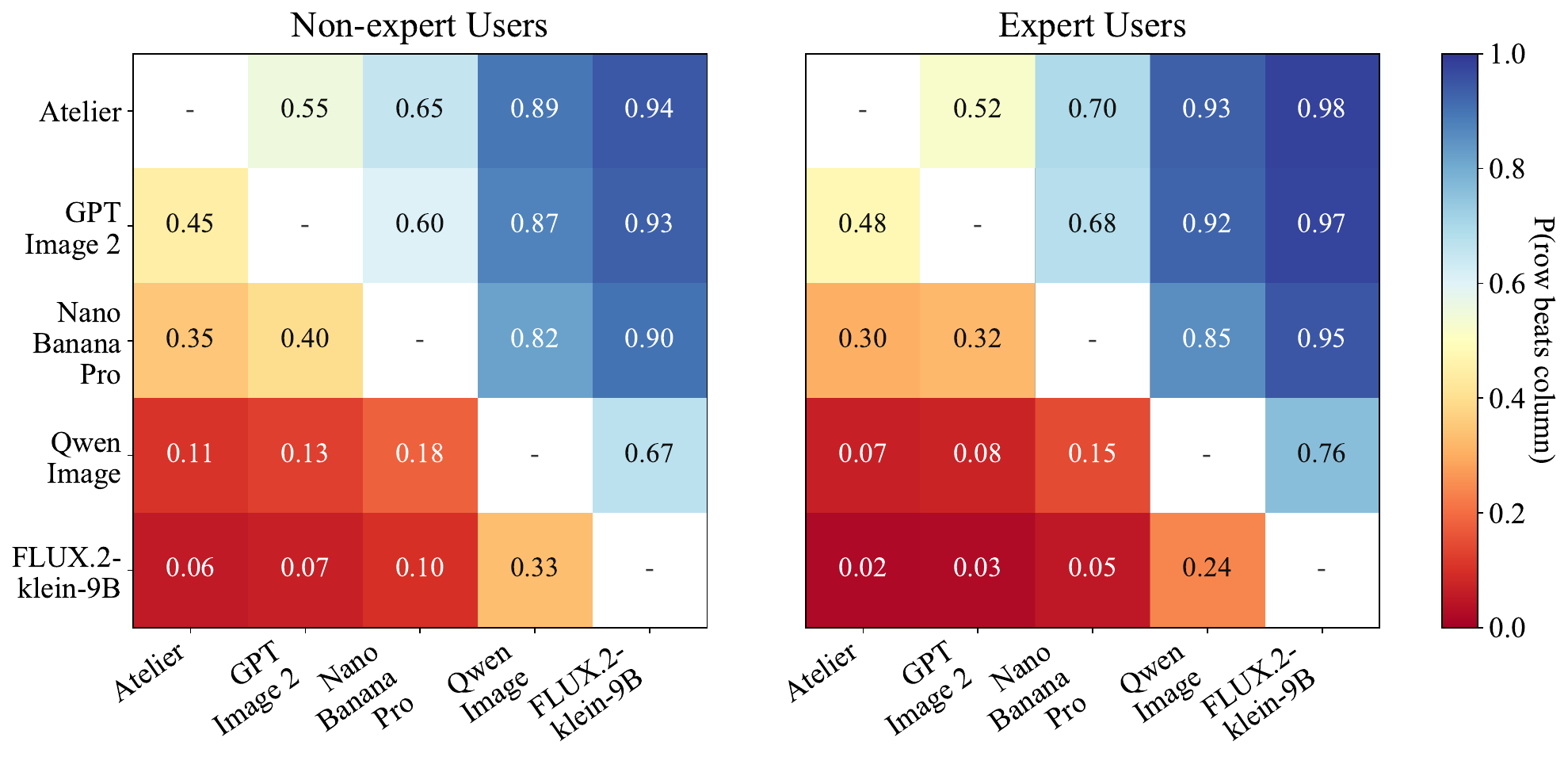}
    \caption{Rater consistency statistics for expert and non-expert groups.}
    \label{fig:user_study_appendix}
\end{figure}

To further assess rater consistency, we used the Bradley--Terry model~\citep{bradley1952rank} to estimate pairwise user preference probabilities between methods. As shown in Figure~\ref{fig:user_study_appendix}, each cell represents the probability that the row method is preferred over the column method. Values greater than 0.5 indicate that the row method outperforms the column method, while values below 0.5 indicate that the column method is preferred. Values farther from 0.5 suggest stronger consistency in raters' preferences.

\subsection{Ablation Studies}
\label{sec:ablation}

\noindent\textbf{Baseline Comparisons.}
The three comparison methods in the left half of Table~\ref{tab:ablation_pair}
are prompt-expansion baselines that share the same candidate-generation
protocol and differ only in how the generation prompt is constructed.
\textbf{Free-form expansion} uses GPT-5.5 to rewrite the raw request into a
generator-ready prompt. \textbf{Expert-template expansion} uses the same model
to fill a fixed template covering the subject, scene layout, atmosphere,
brushwork and material, palette, guardrails, and negative prompt.
\textbf{LLM-reference expansion} first asks GPT-5.5 to construct a reference
plan containing period cues, relevant works, motifs, and composition cues from
its internal knowledge, and then conditions prompt construction on that plan;
it uses neither external search nor Atelier's curated knowledge base or patch
bank.

For each baseline, the resulting prompt is sent to the same four open image
backends (FLUX, Qwen, LongCat, and Hunyuan Image 3.0), with one candidate
generated by each backend. GPT-5.5 then selects the reported baseline output
from the resulting four-candidate pool using the raw request and candidate
images. Atelier instead retains its native selection policy: a global-critic
margin gate followed by authenticity-critic reranking. This is therefore a
system-level baseline comparison rather than a selector-controlled prompt-only
ablation. The \textbf{prompted-critic Atelier variant} retains the full Atelier
pipeline and selection policy but replaces the trained \textsc{AuthCritic}
with a prompted vision--language critic.

\noindent\textbf{Metric.}
The left table reports overall IntroStyle $W_2$ on ArtIntentBench
($\downarrow$). IntroStyle embeds generated images and task-matched authentic
reference paintings in a style-attribution feature space. We compute $W_2$
between the two feature distributions; lower values indicate closer
artist-level style proximity.

Under this protocol, none of the three prompt-construction strategies---
free-form, template-guided, or reference-planned---matches Atelier's
style-proximity result. Replacing the trained \textsc{AuthCritic} with a
prompted vision--language critic (same backbone as the global critic) raises
$W_2$ from $75.64$ to $77.19$, a $1.55$-point gap that isolates the contribution
of the trained local critic; the prompted-critic variant still edges out the
strongest prompt-expansion baseline, indicating that the pipeline structure
itself contributes beyond the trained component.

\noindent\textbf{Component Ablation.} Table~\ref{tab:ablation_pair} (right) reports a diagnostic ablation on the benchmark. Removing structured intent produces the largest degradation in style distance. Removing both feedback modules, iterative writeback, and separately removing artist knowledge and patch references also weaken style distance, indicating that both high-level control and local visual evidence contribute to authenticity.

\begin{table}[H]
\centering
\footnotesize
\caption{Left: baseline comparison on ArtIntentBench. All methods use the same open-backend pool, including FLUX, Qwen, LongCat, and Hunyuan Image 3.0. Right: diagnostic ablation on 40 cases (10 subjects $\times$ 4 periods); each row removes one Atelier component.}
\label{tab:ablation_pair}
\setlength{\tabcolsep}{4pt}
\begin{minipage}[t]{0.47\linewidth}
\centering
\renewcommand{\arraystretch}{1.05}
\begin{tabular}{lc}
\toprule
\textbf{Method} & \textbf{IntroStyle $W_2\downarrow$} \\
\midrule
Prompt expansion (free-form)               & 79.99 \\
Prompt expansion (expert template)         & 77.54 \\
Prompt expansion (LLM-proposed references) & 81.87 \\
\midrule
Atelier (prompted VLM critic)       & \underline{77.19} \\
Atelier                             & \textbf{75.64} \\
\bottomrule
\end{tabular}
\end{minipage}\hfill
\begin{minipage}[t]{0.47\linewidth}
\centering
\renewcommand{\arraystretch}{1.08}
\begin{tabular}{@{}lc@{}}
\toprule
\textbf{Ablation} & \textbf{IntroStyle $W_2\downarrow$} \\
\midrule
\textbf{w/o Artist Knowledge \& Patch Refs} & 84.43 \\
\textbf{w/o Iterative Writeback} & \underline{84.41} \\
\textbf{w/o Structured Intent} & 85.96 \\
\textbf{w/o Both Feedback} & 84.84 \\
\textbf{Atelier} & \textbf{81.79} \\
\bottomrule
\end{tabular}
\end{minipage}
\end{table}

\noindent\textbf{Iterative Refinement.}
Table~\ref{tab:iter_refinement} reports per-round IntroStyle $W_2$ averaged over 45 cases. The mean $W_2$ drops from 83.02 at round one to 79.61 at round two, then remains stable at 79.50 at round three.

\begin{figure}[!t]
    \centering
    \includegraphics[width=0.94\linewidth]{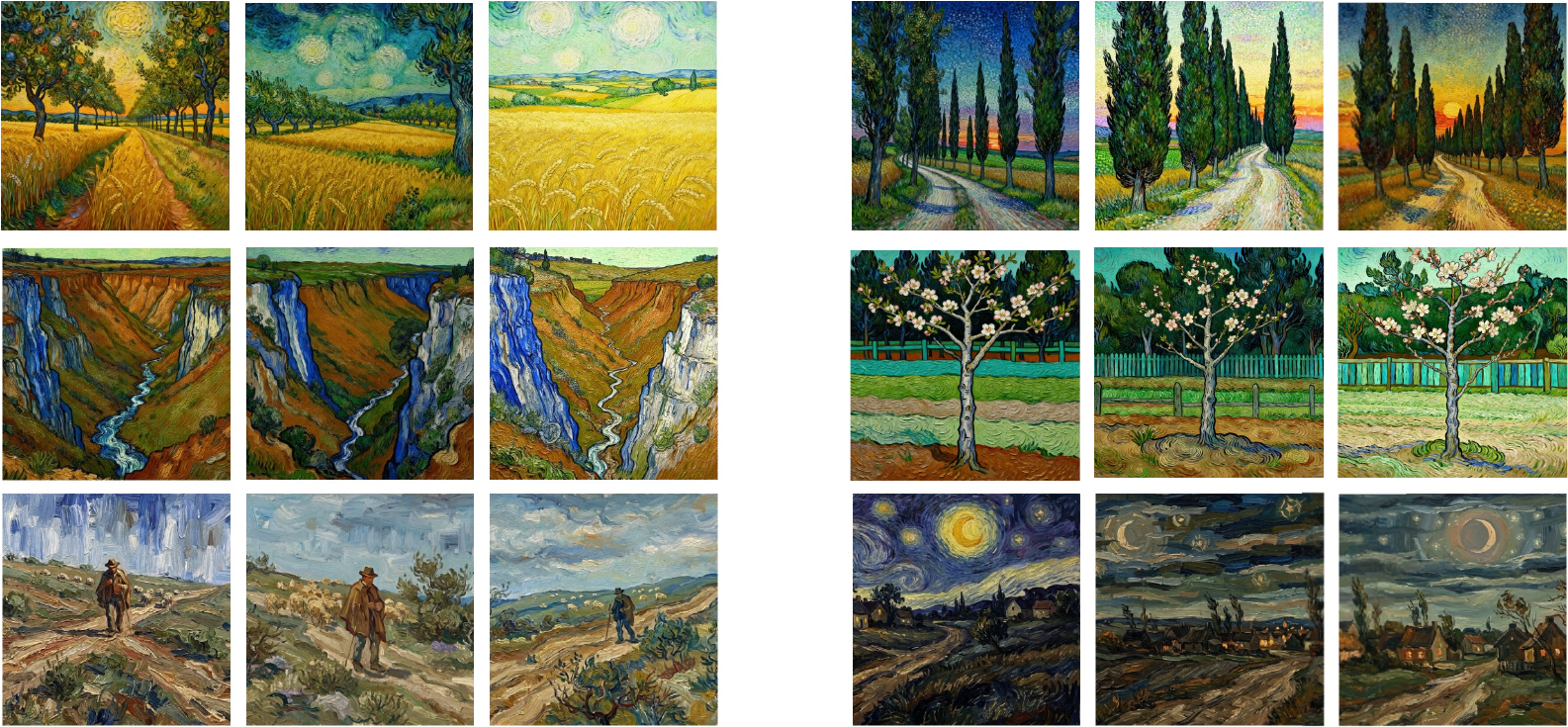}\vspace{-0.35em}
    \includegraphics[width=0.94\linewidth]{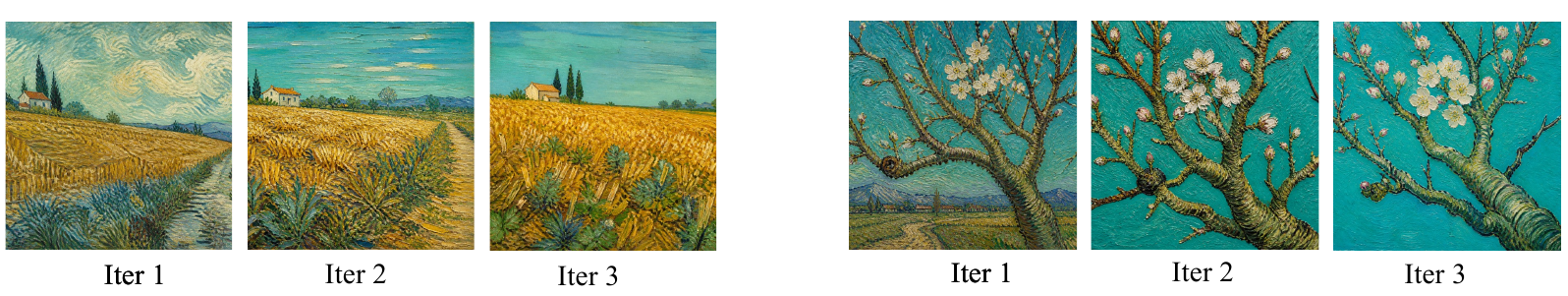}\vspace{-0.35em}
    \caption{
    Qualitative results across three iterations.
    From left to right: input image, result after the first iteration, result after the second iteration, and result after the third iteration.
    The iterative refinement process gradually improves visual fidelity, structural consistency, and target-style alignment.
    }
    \label{fig:three_iteration_results}
\end{figure}

Although we do not manually annotate every intermediate control-state field, the ablations provide indirect evidence about the role of structured interpretation. Removing structured intent produces the largest degradation, suggesting that performance does not come only from longer prompts or stronger image backends. Removing artist knowledge and patch references weakens local style grounding, while removing feedback or writeback reduces the system's ability to recover from shortcut motifs, period mismatch, or under-refined local regions. These results support the view that Atelier's gains come from the combination of explicit preserve/transform reasoning, artist-specific evidence binding, and iterative authenticity feedback.



\begin{table}[H]
\centering
\footnotesize
\caption{Per-round IntroStyle $W_2$ averaged over 45 cases.}
\label{tab:iter_refinement}
\begin{tabular}{lccc}
\toprule
& \textbf{Round 1} & \textbf{Round 2} & \textbf{Round 3} \\
\midrule
Mean $W_2$ $\downarrow$ & 83.02 & 79.61 & \textbf{79.50}\\
\bottomrule
\end{tabular}
\end{table}

\noindent
To better illustrate the effect of iterative refinement, we provide additional qualitative results across three iterations. For each example, we present the input image, the result after the first iteration, the result after the second iteration, and the final result after the third iteration. As shown in Figure~\ref{fig:three_iteration_results}, the first iteration captures the overall structure and main visual appearance, while later iterations progressively improve visual coherence, local details, and alignment with the desired output. The third iteration produces more stable and refined results, demonstrating the effectiveness of the proposed iterative strategy.




\FloatBarrier
\section{Related Work}

\paragraph{Artistic style transfer.}
Neural style transfer renders content in the feature statistics of a
single exemplar \citep{gatys2016style, huang2017adain}, and unpaired
translation extends this to collections of an artist's works
\citep{zhu2017cyclegan}. Medium-specific variants encode further
constraints; ChipGAN \citep{he2018chipgan} enforces voids, brush strokes,
and ink-wash diffusion for Chinese ink painting. These methods transfer
surface statistics of a fixed style target. Artist-grounded generation as
we formulate it additionally requires selecting the right regime within
an artist's oeuvre, deciding which scene content must be preserved, and
rejecting canonical substitutions, none of which is expressed in a style
loss.

\paragraph{Personalization and style-conditioned Image Editing.}
Personalization and style-conditioned generation adapt generative models to user-specified subjects, concepts, or styles \citep{gal2023textual, ruiz2023dreambooth, kumari2023custom, sohn2023styledrop}. While effective with explicit identity or style references, they do not address the upstream problem of translating ambiguous artistic intent into structured, artist-grounded controls.  Instruction-based editing modifies images from natural-language instructions and structural conditions \citep{hertz2022prompt, zhang2023controlnet, brooks2023instructpix2pix}, with recent extensions to local and region-aware control \citep{li2024zone, yu2024anyedit}. These methods typically assume prompts, regions, references, or edit instructions are already specified, whereas our work focuses on inferring such controls, including period, motif, composition, and authenticity-relevant priorities, from vague artistic requests.

\paragraph{Agentic and tool-using generation systems.}
Language agents leverage decomposition, tool use, memory, and iterative feedback for complex tasks \citep{yao2023react, shen2023hugginggpt, shinn2023reflexion, qin2023tool}, and recent personalization agents use intermediate reasoning to interpret user intent \citep{zhang2025personaagent}. Closer to our setting, perceive--plan--execute--evaluate agents apply this loop to image generation \citep{yao2026photoagent}, and hierarchical planner--executor agents use structured interaction memory for image editing \citep{ye2026agentbanana}. These frameworks establish the value of closed-loop generation, but none grounds the loop in artist-level knowledge: \texttt{Atelier} adds a control state that encodes preserve/transform intent, retrieval bound to scene roles, and a trained patch-level authenticity critic.

\paragraph{Evaluation of image generation and artistic authenticity.}
Existing metrics provide global signals for quality, alignment, or human preference \citep{heusel2017fid, radford2021clip, xu2023imagereward}, but are less suited for diagnosing local artistic failures such as brushwork, motif treatment, or material texture. Since art-specific judgment remains challenging even for strong multimodal models \citep{bin2024finegrained, hayashi2024artwork, strafforello2025art}, and VLM style recognition grounds in patch-level concepts that partially align with art-historian judgment \citep{limpijankit2026historians}, we separate global image-level evaluation from localized authenticity diagnosis.

\paragraph{Shortcuts in text-to-image generation.}
Shortcut learning, where a model exploits a spuriously correlated cue instead of the intended structure, is a well-documented failure mode of deep networks \citep{geirhos2020shortcut}. In text-to-image personalization, Goyal et al.~\citep{goyal2025shortcuts} address shortcuts at the adapter-training stage: to prevent adapters from encoding confounding attributes (pose, expression, lighting) alongside target identity, they explicitly provide shortcut pathways during training so the adapter learns only the target attribute. Our work targets a complementary class of shortcuts that arise at inference in general-purpose foundation generators: canonical-motif substitution triggered by artist names, handled through anti-shortcut constraints in the intermediate control state and iterative authenticity feedback rather than additional adapter training.

\section{Discussion}
Scaling artist-grounded evaluation is constrained not only by image availability, but also by the cost of reliable validation. Unlike generic image-generation benchmarks, artist-grounded evaluation requires assessing period fidelity, motif appropriateness, brushwork, material treatment, and shortcut substitutions, properties that are difficult to capture with automatic metrics alone. ArtIntentBench therefore adopts a depth-first design: rather than covering many artists with shallow labels, it focuses on two contrastive artistic traditions, Van Gogh's period-sensitive oil painting and Qi Baishi's sparse ink-and-color painting, and evaluates multiple forms of generalization within this setting. To improve reliability, we combine automatic style-distance evaluation with blinded human ranking by art-trained non-specialists and expert raters. Future work will extend the framework to broader artist collections, improve automatic construction and verification of artist knowledge bases, and study interactive settings in which users can inspect, edit, and approve the intermediate control state before generation.

\section*{Limitation}
Our benchmark currently covers two artists. This scope is intentional: constructing reliable artist-grounded evaluations requires curated artist knowledge, patch-level evidence, and expert or art-trained human validation.

\section*{Ethics and Copyright Statement}
The Van Gogh corpus draws on public-domain reproductions from WikiArt and the National Gallery of Art; the Qi Baishi corpus draws on public museum digital collections of traditional Chinese art. We will release code, metadata, benchmark prompts, and evaluation protocols; copyrighted source images will not be redistributed but referenced via links to original hosting pages. Reproducibility depends on the released metadata, including benchmark prompts, the control-state schema, perceiver and planner prompts, retrieval details, and patch-bank construction. Each evaluation record additionally stores the normalized target artist, the critic prompt-template and schema identifiers, the SHA-256 of the instantiated system prompt, the evaluator model identifier, and the \textsc{AuthCritic} adapter identifier, making each reported result traceable to the exact artist-conditioned evaluator configuration.

\newpage
\bibliographystyle{splncs04}
\bibliography{references}

\appendix
\clearpage

\section{Control-State Walkthrough}
\label{app:walkthrough}

This appendix traces two recorded episodes.
Appendix~\ref{app:walkthrough-single} exhibits all five components of the
state $z=(s,q,h,r,b)$ (\S\ref{sec:control_state}) on a single held-out
request. Appendix~\ref{app:walkthrough-loop} follows a three-round refinement
trace on a second request, showing how critic feedback is transported through
the reflective memory into subsequent plans.

The runtime does not serialise $z$ as one JSON object. Its components are
distributed across the perceiver's scene specification, the world model,
the planner's \texttt{knowledge\_evidence} block, and the compiled plans;
the field-level mapping is given in Appendix~\ref{app:schema}
(Table~\ref{tab:z-field-map}). The excerpts below group the recorded fields
by the component of $z$ they realise; evaluation fields keep the
implementation's names (e.g., \texttt{heavy\_judge\_score} is the
global-critic score $s_g$ of Appendix~\ref{app:selection-policy}). Elisions
are marked with \texttt{...}. Values and text are copied from the records;
the backend excerpt omits compatibility-only arguments that the corresponding
backend does not consume.

\subsection{Single-Round Control State}
\label{app:walkthrough-single}

The episode uses a held-out still-life request:

\begin{quote}\itshape
A low-angle still life of harvested vegetables---pale cabbage, reddish root
crops, and scattered yellow maple leaves---resting on dark soil, rendered in
Paris-period muted earth tones with directional impasto brushwork, arranged
in a diagonal mass from lower-left to upper-right against a deep shadowed
background.
\end{quote}

\paragraph{Scene reading $s$.}
The perceiver assigns the request to the closed scene-regime catalogue
(Prompt~\ref{lst:perceiver-catalogs}) and emits three allowed local-role
names. The world-model builder enriches these roles with the style axes
along which they may be translated:

\begin{promptlisting}
"scene_skeleton": {
  "primary_regime": "object_still_life",
  "active_regimes": ["object_still_life"],
  "setting_detail": "still_life_tabletop",
  "mood": "calm", ...
},
"key_local_roles": [
  {"scene_role": "foreground_object_mass",
   "translation_axes": ["surface_relief", "palette_relation", "contour_pressure"]},
  {"scene_role": "field_surface_band",
   "translation_axes": ["brush_rhythm", "palette_relation", "contour_pressure"]},
  {"scene_role": "background_halo_field",
   "translation_axes": ["directional_motion", "palette_relation", "surface_relief"]}
]
\end{promptlisting}

\paragraph{Preserve/transform decisions $q$.}
The same record separates preserved structure from translatable style axes:

\begin{promptlisting}
"preserve": [
  {"target": "object_silhouette_integrity"},
  {"target": "tabletop_support_plane"}
],
"translate": [
  {"target": "foreground_object_mass",
   "translation_axes": ["surface_relief", "palette_relation", "contour_pressure"]},
  {"target": "field_surface_band",
   "translation_axes": ["brush_rhythm", "palette_relation", "contour_pressure"]},
  {"target": "background_halo_field",
   "translation_axes": ["directional_motion", "palette_relation", "surface_relief"]}
]
\end{promptlisting}

\paragraph{Historical/style-regime intent $h$.}
The request names the Paris period, so the perceiver records an explicit
period signal; the downstream period policy consequently collapses to a
single route:

\begin{promptlisting}
"clarifier_period_signal": {
  "explicit_period_preference": "paris_period",
  "period_clarification_mode": "explicit", ...
},
"period_policy": {
  "mode": "single_route",
  "dominant_period": "paris_period",
  "candidate_periods": ["paris_period"],
  "reason": "explicit_period_preference"
},
"period_guidance": [
  {"period_label": "paris_period", "score": 37.4,
   "top_form_tags": ["expressive brushwork", "complementary colors", "greens", ...],
   "dominant_palette_relations": ["complementary colors", "greens", "lighter palette", ...], ...},
  {"period_label": "nuenen_period", "score": 33.4, ...}
]
\end{promptlisting}

\paragraph{Retrieval targets and anti-shortcut constraints $r$.}
Each scene role is bound to a patch family, or explicitly left unbound.
Here only one role finds a role-consistent family; the other two fall back
to work-level evidence:

\begin{promptlisting}
"motif_patch_binding_plan": [
  {"motif_name": "foreground object mass",
   "scene_role": "foreground_object_mass",
   "family_id": "nuenen_period::foreground_object_mass::foreground_object_mass",
   "binding_status": "bound_patch_family",
   "representative_patch_ids": ["patch_03958"],
   "family_cluster_size": 6, ...},
  {"motif_name": "field surface band",
   "binding_status": "unbound_patch_family_fallback_to_work_evidence",
   ...},
  {"motif_name": "background halo field",
   "binding_status": "unbound_patch_family_fallback_to_work_evidence",
   ...}
]
\end{promptlisting}

The bound family is indexed under \texttt{nuenen\_period} although the
generation route is Paris: the period label enters retrieval as a scoring
term rather than a hard filter, so an adjacent-period family can win the
rerank while the generation route is unchanged. The per-candidate audit
records the score decomposition; each column is a term of the family
scoring rule (\S\ref{sec:artist_kb}):

\begin{promptlisting}
"selected_candidates": [
  {"patch_id": "patch_03958",
   "source_work_title": "Still Life (F.1972.44.P)",
   "score": 61.8,
   "score_breakdown": {
     "period_score": 24.0,
     "reference_work_prefilter_bonus": 8.0,
     "patch_role_score": 20.0,
     "canonical_work_bonus": 2.2,
     "asset_priority_bonus": 4.0,
     "world_model_alignment_bonus": 3.6,
     "anti_iconic_penalty": 0.0, ...}}
], ...
\end{promptlisting}

The knowledge base has no subject-level evidence for this request; the
record encodes this as a policy over what the planner may use, not as a
substitution:

\begin{promptlisting}
"subject_kb_support": {
  "subject_kb_support_level": "none",
  "has_subject_patches": false,
  "matched_motif_count": 0,
  "controller_policy": {
    "allow_style_guidance": true,
    "allow_compatible_geometric_anchors": true,
    "allow_motif_style_analogy": false,
    "allow_subject_content_anchors": false
  }, ...
}
\end{promptlisting}

\paragraph{Backend-facing preferences $b$.}
The planner compiles the state into $N=3$ parallel backend calls with a
shared generation plan. These calls are the case-specific subset selected
for this episode, rather than an enumeration of the track-level backend pool
in Appendix~\ref{app:selection-policy}:

\begin{promptlisting}
"tool_plan": {
  "primary_tool_id": "remote_flux_lora",
  "calls": [
    {"tool_id": "remote_flux_lora",
     "arguments": {"artist": "van_gogh",
                   "period_resource": "paris_period",
                   "style_strength": 0.9, "lora_scale": 0.9, ...}},
    {"tool_id": "qwen_t2i",
     "arguments": {"period_resource": "paris_period",
                   "style_strength": 0.8, ...}},
    {"tool_id": "longcat_t2i",
     "arguments": {"period_resource": "paris_period",
                   "style_strength": 0.8, ...}}
  ]
},
"generation_plan": {
  "positive_prompt": "Low-angle still life oil painting in the style of Vincent van Gogh's Paris-period. A mass of harvested vegetables arranged diagonally from lower-left to upper-right ... Muted earth tones, warm ochres, olive greens, deep browns ... Thick, directional impasto brushstrokes ...",
  "negative_prompt": "glossy, photorealistic, smooth, flat, cartoon, bright saturated colors, pure black background, signature, watermark, ..., bright blue sky, sunflower, rural landscape, over-decorative halo.",
  "guidance_note": "... Preserve structural anchors: object_silhouette_integrity, tabletop_support_plane. Translate style through: foreground_object_mass via surface_relief, palette_relation, contour_pressure; field_surface_band via brush_rhythm, palette_relation, contour_pressure; background_halo_field via directional_motion, palette_relation, surface_relief."
}
\end{promptlisting}

The internal identifier \path|remote_flux_lora| denotes the
\textsc{flux2-lora} backend. Only this call consumes \texttt{lora\_scale} as
an adapter weight. The raw planner record also places a compatibility
\texttt{lora\_scale} field on the Qwen and LongCat calls; the executor ignores
that field for these backends, so it is omitted above. Their
\texttt{style\_strength} values provide prompt-level guidance and do not load
a LoRA adapter. Hunyuan belongs to the open-weight pool but was not selected
in this episode's three-call plan.

The compiled plan carries the fields of $r$ and $q$ into the backend text:
canonical Van~Gogh shortcuts appear as negative-prompt terms
(\texttt{sunflower}, \texttt{bright blue sky}), and the guidance note
restates the preserve targets and per-role translation axes verbatim, so
the same constraints reach the critic in \S\ref{sec:evaluator}.

\subsection{Three-Round Refinement Trace}
\label{app:walkthrough-loop}

This subsection follows a completed open-weight Qi~Baishi episode. The
recorded request is:

\begin{quote}\itshape
Qi Baishi-style ink-and-wash composition featuring bold, wet-ink lotus leaves
with expressive white highlights, a vivid magenta lotus bloom, and two
colorful mandarin ducks rendered in simplified forms with yellow, blue, and
purple washes; vertical calligraphy and a red seal appear on the left, while
ample blank paper space enhances the dynamic asymmetry.
\end{quote}

The evaluator provenance matches the release-v2 contract in
Appendix~\ref{app:critic-v2-contract}: the record names the Qi~Baishi v2
global-critic prompt and schema together with the checkpoint-482
\textsc{AuthCritic} adapter. The first round explores the four-backend pool.
Hunyuan supplies the winner and is then locked for the remaining rounds; the
trace therefore holds the selected generator fixed while the execution prompt
is revised.

\begin{figure}[H]
  \centering
  \begin{subfigure}[t]{0.32\linewidth}
    \includegraphics[width=\linewidth]{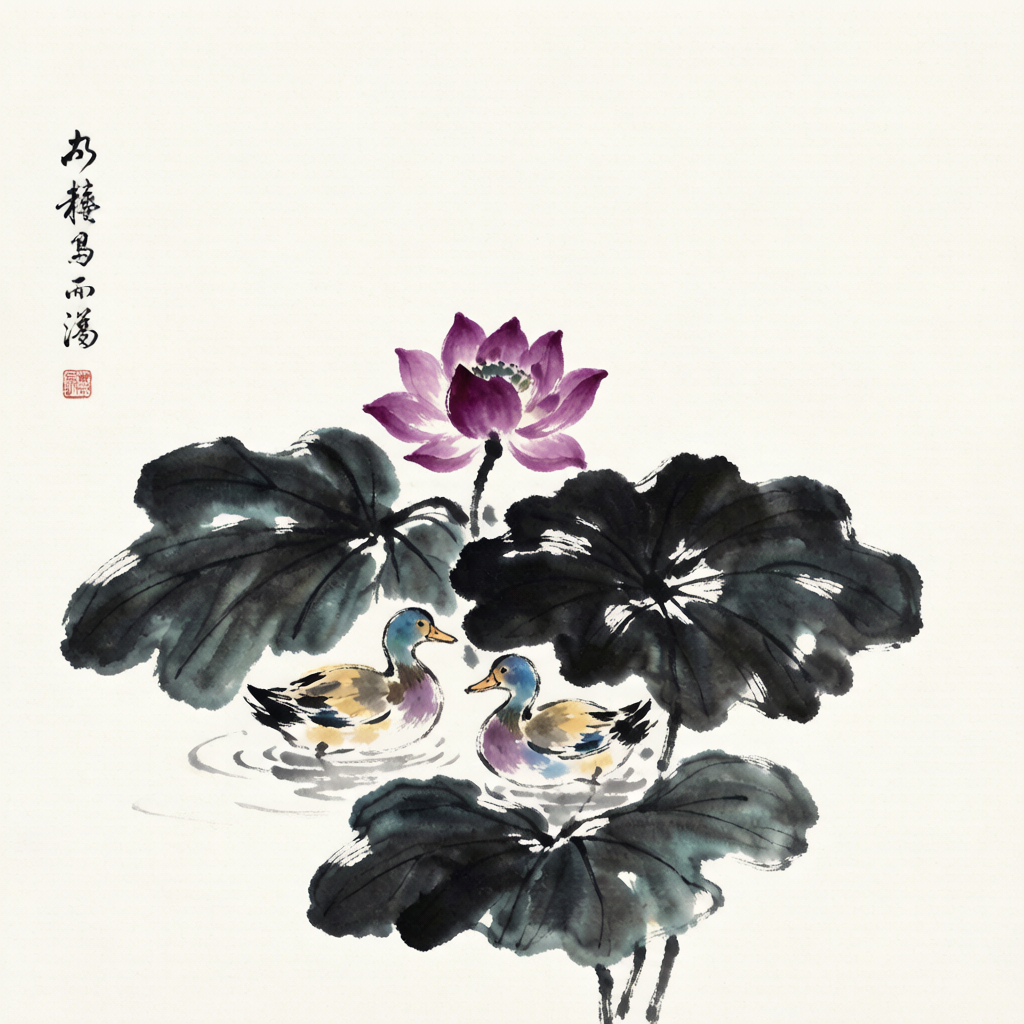}
    \caption{Round 1: $s_g=47$.}
  \end{subfigure}\hfill
  \begin{subfigure}[t]{0.32\linewidth}
    \includegraphics[width=\linewidth]{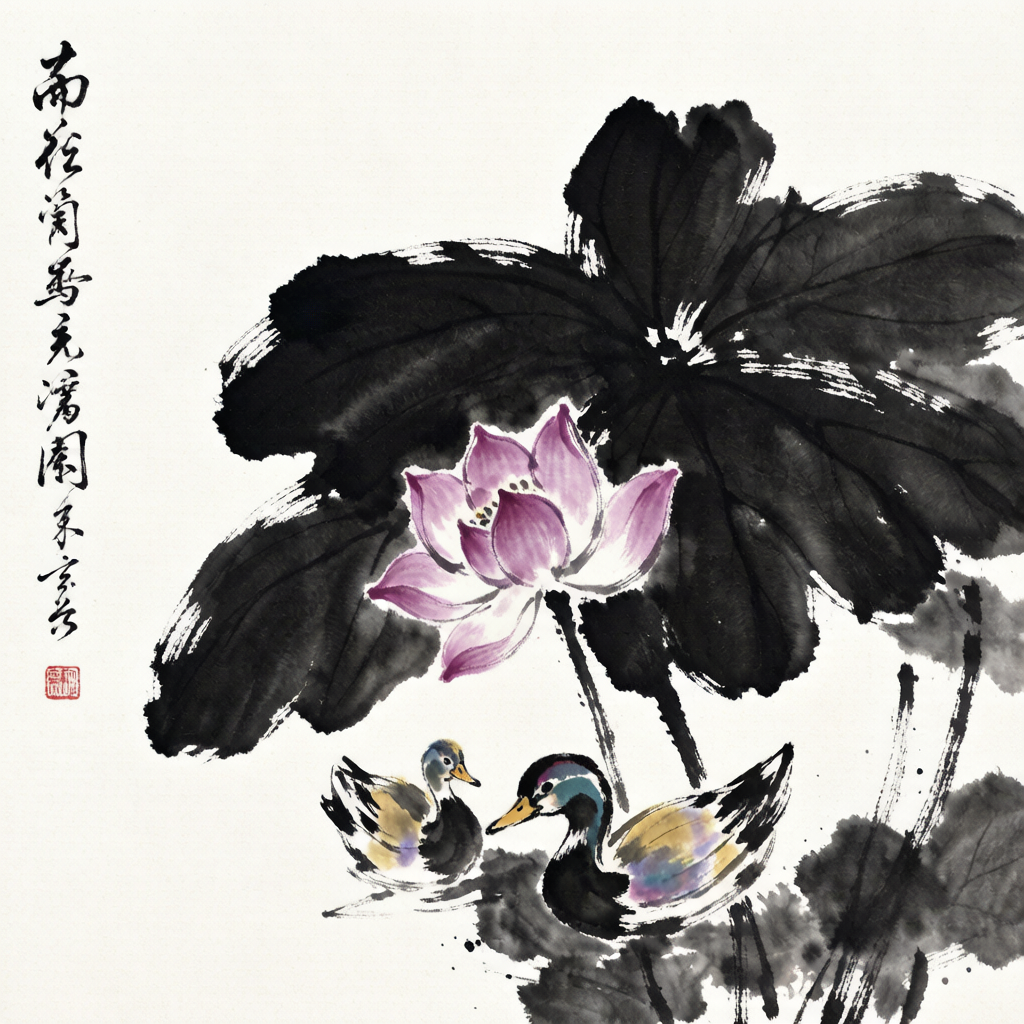}
    \caption{Round 2: $s_g=49$.}
  \end{subfigure}\hfill
  \begin{subfigure}[t]{0.32\linewidth}
    \includegraphics[width=\linewidth]{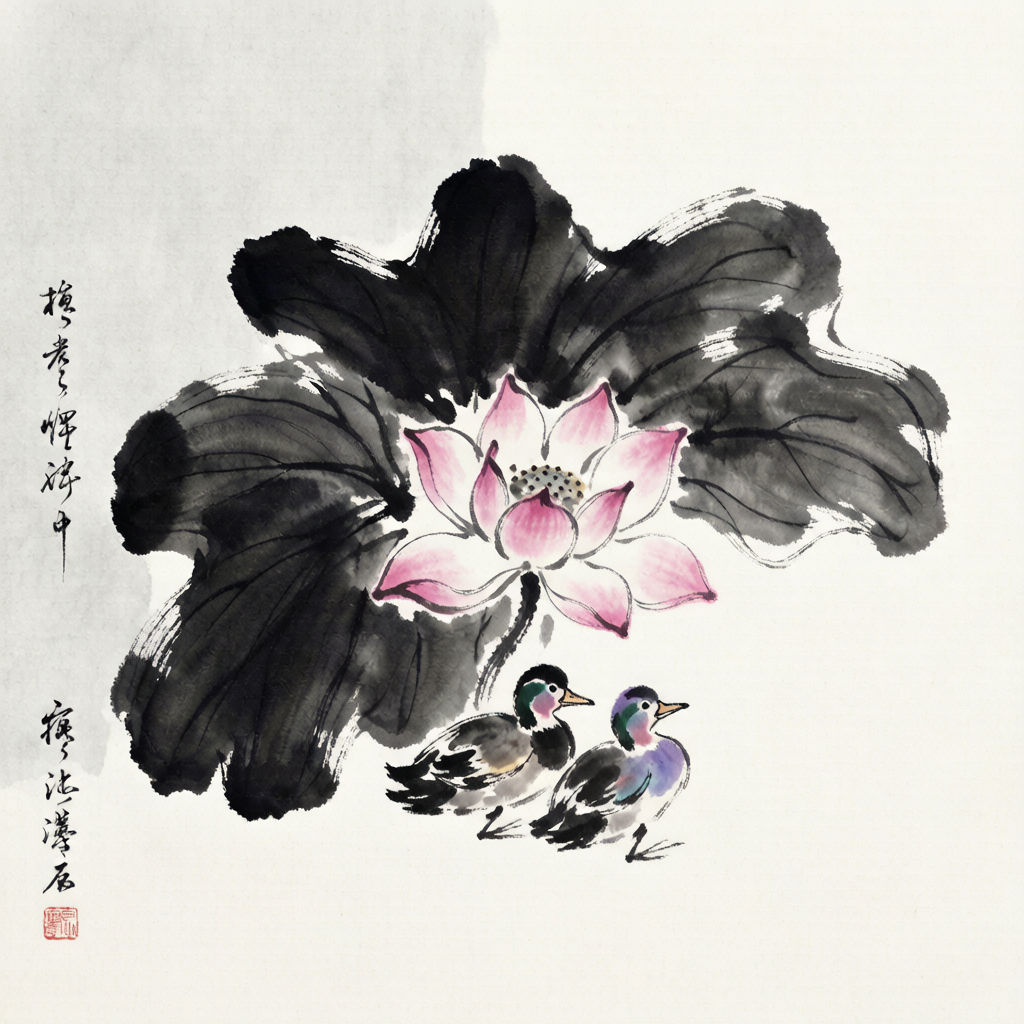}
    \caption{Round 3: $s_g=57$.}
  \end{subfigure}
  \caption{Three-round refinement trace for a Qi~Baishi request. Hunyuan is
  locked after Round~1. Global-critic scores increase from 47 to 49 to 57;
  the corresponding AuthCritic aggregate scores are 42, 67, and 67.}
  \label{fig:qibaishi-refinement-trace}
\end{figure}

\paragraph{Round 1.}
The candidate preserves the requested subjects and layout, but the critic
identifies weak calligraphic energy, limited ink-density transitions, and a
drift toward generic decorative illustration. These observations are stored
as canonical tags together with localized repair targets:

\begin{promptlisting}
"heavy_judge_score": 47,
"expert_score_100": 42,
"canonical_failure_tags": [
  "style_calligraphic_energy_weak",
  "material_ink_density_transition_weak",
  "shortcut_generic_decorative_illustration",
  "brushwork_too_uniform"
],
"repair_targets": [
  "leaf edges: replace soft wash boundaries with tapered, dry-brush frayed terminations",
  "lotus bloom: replace solid magenta with fine ink contour + diluted magenta wash over graded ink base",
  "ducks: reduce color saturation, integrate ink underwash, simplify forms to 3-5 expressive strokes per duck"
]
\end{promptlisting}

The reflective memory records the canonical tags verbatim and separates the
repair targets from preserved strengths, including the subject count, the
left-side inscription and seal, and the blank-paper reserve. It also records
the backend lock:

\begin{promptlisting}
"locked_backend": "hunyuan_image_t2i",
"backend_locked_from_iteration": 1,
"canonical_failure_tag_history": [
  {"iteration": 1,
   "target_artist_id": "qi_baishi",
   "failure_tags": [
     "style_calligraphic_energy_weak",
     "material_ink_density_transition_weak",
     "shortcut_generic_decorative_illustration",
     "brushwork_too_uniform"]}
], ...
\end{promptlisting}

\paragraph{Round 2.}
The next generation prompt instantiates the recorded repairs as material and
stroke instructions: dry-brush fraying on leaf edges, graded wash beneath the
magenta bloom, ink underwash for the ducks, and pressure-varying calligraphic
strokes. The subject and layout constraints are retained:

\begin{promptlisting}
"generation_prompt": "Traditional Chinese ink wash painting in the xieyi freehand manner of Qi Baishi. Bold wet-ink lotus leaves with dry-brush frayed edges, irregular white paper reserve highlights. A vivid magenta lotus bloom with fine dark ink contour and diluted magenta wash. Two simplified mandarin ducks rendered in 3-5 expressive calligraphic strokes each ... Vertical column of calligraphy on the left ... Small red seal beneath calligraphy. Ample blank paper space ... Ink density variation from deep black to pale gray ..."
\end{promptlisting}

The global score increases to 49, while the AuthCritic aggregate increases
from 42 to 67. The remaining local review calls for layered petal washes,
dry-brush stippling, tapered stroke endings, and less uniform saturation.

\paragraph{Round 3.}
The third prompt carries these revisions forward. It specifies ink gradation
within the leaves, translucent rather than solid petal color, dry-brush
fraying, and abbreviated duck forms while preserving the requested count and
composition:

\begin{promptlisting}
"generation_prompt": "Master Qi Baishi-style ink-and-wash painting on absorbent xuan paper. A lively waterside scene: three bold lotus leaves executed with large wet-ink brushstrokes that show rapid pressure release and dry-brush fraying at the edges ... Leaf interiors show rich ink gradation from deep black to watery gray. A single magenta lotus bloom composed of fine calligraphic ink contours and translucent magenta wash over a graded ink underlay ... Two simplified mandarin ducks reduced to xieyi essentials ... Vertical column of calligraphic script on the left side ... a small red seal stamp placed near the lower-left corner ..."
\end{promptlisting}

The global score reaches 57 and \textsc{FinalSelect} returns the Round~3
candidate. The episode remains below $\tau^\star=85$ and terminates at the
budget $T=3$ with recommendation \texttt{revise}. Figure~\ref{fig:qibaishi-refinement-trace}
is therefore a monotonic three-round refinement trace, not an accepted
outcome. It isolates the transport of localized critic findings through
reflective memory into successive execution prompts while the selected
backend remains fixed.

\clearpage
\section{Prompt Library}
\label{app:prompt-library}

This appendix reproduces the prompts used by every LLM-based component of
\texttt{Atelier}: the perceiver (Appendix~\ref{app:perceiver-prompts}), the
planner (Appendix~\ref{app:planner-prompt}), the global critic
(Appendix~\ref{app:critic-prompts}), the local \textsc{AuthCritic}
(Appendix~\ref{app:authcritic-prompt}), and the control-state audit judges used
in the Control State Quality evaluation
(Appendix~\ref{app:audit-protocols}). The two control-state audit listings are
limited to the fields included in the reported analysis; the remaining prompt
listings reproduce the deployed text.\footnote{Module names in the
implementation predate the terminology of this paper: the perceiver corresponds
to the \texttt{clarifier} module and the planner to the \texttt{controller}
module. Except for the two explicitly scoped audit listings, prompt text is
reproduced verbatim from the implementation, so these
internal names still appear inside the prompt strings (e.g., ``You are the
Agent Controller \ldots'').} Prompts are shown as templates: fields in curly
braces, such as \texttt{\{user\_request\}}, are filled at runtime; everything
else is fixed instruction text. In the implementation the instruction sentences
are concatenated into a single string; line breaks are inserted here for
readability only.

\subsection{Perceiver Prompts}
\label{app:perceiver-prompts}

The perceiver issues two prompts. The \emph{scene-spec extraction prompt}
(Prompt~\ref{lst:perceiver-scene-spec}) converts the raw request into the
scene-facing fields of the control state: the scene regime, subject type,
must-have and forbidden entities, and the scene roles that later drive
patch-role binding. The \emph{clarification and intent-inference prompt}
(Prompt~\ref{lst:perceiver-review}) additionally decides whether follow-up
questions are needed, infers reference intent (whether the user implicitly
anchors on a specific work or work family), and produces the period-preference
signals used by retrieval. Both prompts constrain the model to closed catalogs
of scene regimes and scene roles, reproduced in
Prompt~\ref{lst:perceiver-catalogs}; the model may not invent regime or role
names.

For compactness, Prompt~\ref{lst:perceiver-review} denotes the shared block
from \texttt{EVIDENCE RULES} through the four \texttt{BOUNDARY EXAMPLES} in
Prompt~\ref{lst:perceiver-scene-spec} by
\path|{scene_spec_policy}|. The implementation inserts that block verbatim.
The target-artist route supplies \path|{artist_specific_scene_spec_rules}|.

\begin{promptlisting}[caption={Perceiver scene-spec extraction prompt.},label={lst:perceiver-scene-spec}]
You are the scene-spec extractor for an artistic image-generation system. Convert the user request into a minimal, evidence-grounded scene specification for downstream retrieval and planning.

Return one valid JSON object only. Do not use Markdown, comments, or text outside the JSON object. Return exactly these keys:

{
  "primary_regime": "<one value from the supplied regime catalog>",
  "subject_type": "<short noun phrase>",
  "must_have_entities": ["<explicitly requested entity>"],
  "forbidden_entities": ["<explicitly excluded entity>"],
  "allowed_scene_roles": ["<one to four values from the role catalog>"],
  "period_preference": "<explicit period label or empty string>",
  "confidence": <number from 0.0 to 1.0>,
  "reasoning_note": "<one concise sentence>"
}

EVIDENCE RULES

1. Base every output field on the user request.
2. Do not add objects, settings, periods, or scene roles merely because they are common in the target artist's work.
3. Put explicitly required subjects and objects in must_have_entities.
4. Put only explicitly negated or excluded entities in forbidden_entities. Use [] when the request contains no exclusions.
5. Negated content must never appear in must_have_entities.
6. subject_type describes the requested compositional subject, not the artistic style.

PERIOD RULE

Set period_preference only when the request explicitly names a period, date range, or unambiguous period alias. Otherwise return an empty string. Do not infer a period from the artist name, subject matter, palette, brushwork vocabulary, or resemblance to a canonical work.

REGIME SELECTION

Select primary_regime from the supplied regime catalog only. Classify the requested image by its dominant compositional organization, not by incidental content. Apply this precedence: (1) explicit framing or scene-type statements in the request; (2) the requested compositional subject and its framing; (3) the surrounding setting. The specific boundary rules below override this general precedence.

BOUNDARY RULES

- portrait_subject applies only to explicitly face-focused, bust-length, sitter-led, close-up portrait, or self-portrait framing.
- A person appearing in a larger scene, including a farmer, worker, reader, walker, or figure in a field, room, courtyard, path, street, or waterside setting, is not sufficient evidence for portrait_subject. Classify such requests by the surrounding scene organization, even when the person is visually important.
- floral_still_life applies only when flowers, a bouquet, or a flower-filled vase are the requested subject.
- object_still_life applies to non-floral foreground or tabletop objects, including fruit, bottles, vessels, shoes, boots, and tools.
- workspace_tabletop applies to desks or work surfaces organized around papers, notebooks, keyboards, tools, or office objects.
- modern_device_scene applies when a laptop, smartphone, computer, tablet, camera, headphones, or screen is a central requested subject.
- domestic_edge_exterior applies to small built exterior structures such as pavilions, walls, courtyards, porches, doors, and building edges, unless the request clearly centers an open rural landscape, a civic landmark, or another more specific regime.
- A potted flower arrangement on a table or windowsill is normally floral_still_life, not rural_landscape or interior_domestic, unless the room itself is explicitly the dominant subject.

ROLE SELECTION

- Select allowed_scene_roles from the supplied role catalog only.
- Return one to four roles. Two or three are preferred when supported, but never add a weakly supported role merely to reach a target count.
- Select roles that describe the requested composition, not motifs conventionally associated with the artist.
- For close foreground objects, prioritize foreground_object_mass. Use tabletop_edge_band only when a table or comparable support surface is requested or clearly entailed.
- facial_plane_focus and head_silhouette_anchor require explicitly face-focused, bust-length, or portrait framing. A human figure inside a larger scene does not qualify.
- facade_plane, roof_plane, shelter_frame_rhythm, bridge_span, and other architectural roles require the corresponding built structure to be requested as compositionally important.
- Do not select architectural roles merely because an object appears indoors or near a building.

ARTIST-SPECIFIC RULES

{artist_specific_scene_spec_rules}

BOUNDARY EXAMPLES

Each example output is a scene-spec object. When this policy is embedded in the full clarification task, place that object under scene_spec and still return every required top-level clarification key.

Example 1
User request: A close-up bust portrait of a woman against a pale wall.
Output: {
  "primary_regime":"portrait_subject",
  "subject_type":"close-up bust portrait",
  "must_have_entities":["woman"],
  "forbidden_entities":[],
  "allowed_scene_roles":["facial_plane_focus","head_silhouette_anchor","background_halo_field"],
  "period_preference":"",
  "confidence":0.96,
  "reasoning_note":"The request explicitly specifies close-up bust portrait framing."
}

Example 2
User request: A farmer walking along a path through a wheat field at dusk.
Output: {
  "primary_regime":"rural_landscape",
  "subject_type":"farmer in a wheat field",
  "must_have_entities":["farmer","path","wheat field"],
  "forbidden_entities":[],
  "allowed_scene_roles":["field_surface_band","foreground_path","sky_band"],
  "period_preference":"",
  "confidence":0.94,
  "reasoning_note":"The person belongs to a field scene rather than portrait framing."
}

Example 3
User request: A vase of irises on a windowsill, with no landscape outside.
Output: {
  "primary_regime":"floral_still_life",
  "subject_type":"floral windowsill still life",
  "must_have_entities":["vase","irises","windowsill"],
  "forbidden_entities":["landscape"],
  "allowed_scene_roles":["flower_cluster_mass","vase_body_anchor","tabletop_edge_band"],
  "period_preference":"",
  "confidence":0.97,
  "reasoning_note":"The flowers and vase are the dominant subject and the landscape is explicitly excluded."
}

Example 4
User request: A pair of worn boots on a wooden table, without flowers or buildings.
Output: {
  "primary_regime":"object_still_life",
  "subject_type":"foreground object still life",
  "must_have_entities":["worn boots","wooden table"],
  "forbidden_entities":["flowers","buildings"],
  "allowed_scene_roles":["foreground_object_mass","tabletop_edge_band","background_halo_field"],
  "period_preference":"",
  "confidence":0.97,
  "reasoning_note":"The request centers non-floral tabletop objects and explicitly excludes architectural content."
}

Scene regime catalog:
{scene_regime_catalog}

Scene role catalog:
{scene_role_catalog}

User request: {user_request}
\end{promptlisting}

\begin{promptlisting}[caption={Perceiver clarification and intent-inference prompt.},label={lst:perceiver-review}]
You are a request clarifier for a personal artistic agent. The target artist for this request is {artist_display_name} (artist_id={artist_id}). Your job is structured condition completion for image generation, not therapy, not psychological counseling, and not privacy collection. If the user request is underspecified, decide what 1-3 missing details would most improve artistic control. Always also infer whether the user is referencing a specific {artist_display_name} work or work family as a style anchor, even if the wording is implicit. Always also infer a scene_regime_prior and scene_spec from the text request itself. Return strict JSON with keys: needs_clarification, ambiguity_score, missing_slots, questions, inferred_spec, reference_intent, scene_regime_prior, scene_spec, clarification_note. questions must be a list of objects with slot, question, why. reference_intent must be an object with keys: detected, explicitness, relation_type, granularity, candidate_targets, transfer_focus, preserve_subject_constraints, anti_copy_constraints, should_enable_reference_routing, should_request_visual_reference_disambiguation, confidence, reasoning_note.
candidate_targets must be a list of objects with title_hint, artist, confidence. scene_regime_prior must be an object with keys: primary_regime, secondary_regime, confidence, non_rural_structure_preservation, reasoning_note. scene_spec must be an object with keys: primary_regime, subject_type, must_have_entities, forbidden_entities, allowed_scene_roles, period_preference, confidence, reasoning_note. The following policy governs the scene_spec object:

{scene_spec_policy}

Only ask about truly useful missing factors such as mood, setting, time_of_day, weather, character_profile, palette_bias, expressive_goal, life_context, memory_or_present_focus, explicit_period_preference, period_preference_signals. Treat period_preference_signals as an artist-specific style-preference signal; do not hard-code another artist's periods or canonical examples. Prefer low-intrusion personalized slots like expressive_goal, life_context, and memory_or_present_focus when they help image control. Do not ask for numeric age. Only use age_stage if the user already frames the request that way and it is directly useful for image generation. If the user sounds like an expert or mentions artist stages/resources directly, you may ask explicit_period_preference. Otherwise prefer period_preference_signals phrased as visual feeling or artist-specific style axes rather than period names.
{artist_style_question_text}
If the request implies something like 'paint this subject the way {artist_display_name} handled a known work family', mark reference_intent.detected=true, explicitness=implicit, preserve the requested subject, and do not let the reference work replace the user subject. For scene_spec, do not let forbidden or negated entities become positive motif cues.

Scene regime catalog:
{scene_regime_catalog}

Scene role catalog:
{scene_role_catalog}

User request: {user_request}
\end{promptlisting}

\begin{promptlisting}[caption={The closed scene regime and scene role catalogs inserted into the perceiver prompts at the \{scene\_regime\_catalog\} and \{scene\_role\_catalog\} placeholders.},label={lst:perceiver-catalogs}]
Scene regime catalog:
- floral_still_life: flower/bouquet/vase-centered tabletop or windowsill still life only; not a generic still-life fallback
- object_still_life: first-class non-floral tabletop object still life such as teapot, cup, bottle, fruit, boots, shoes, tools, vessels, or other close foreground objects
- workspace_tabletop: first-class desk/workspace tabletop arrangement with non-floral objects, papers, keyboard, notebook, tools, or office surface organization
- modern_device_scene: first-class modern device scene centered on laptop, smartphone, computer, tablet, headphones, camera, screen, or similar device identity
- interior_domestic: indoor room scene, often with furniture/window or human-in-interior
- interior_transient_institutional: corridor, stair, landing, or institutional transition space whose circulation structure dominates
- domestic_edge_exterior: small built exterior edge such as courtyard, pavilion, wall, porch, doorway
- civic_landmark_or_plaza: church/chapel/plaza/landmark-led architecture scene
- urban_street: street/city/cafe/market scene
- transit_mobility: tram/station/platform/railway scene
- waterside_nature: river/canal/shore/water-led scene
- rural_landscape: open rural field/hillside/village-landscape scene when no stronger built or still-life structure dominates
- portrait_subject: close-up or sitter-led portrait scene
- industrial_edge: factory/warehouse/mill/industrial built scene
- overlook_elevated_view: rooftop/terrace/balcony/overlook scene

Scene role catalog:
- facial_plane_focus: facial plane focus
- head_silhouette_anchor: head silhouette anchor
- background_halo_field: background halo field
- sky_band: sky band
- roof_plane: roof plane rhythm
- tower_silhouette_anchor: tower silhouette anchor
- window_glow_cluster: window glow cluster
- terrace_edge: terrace edge rhythm
- reflection_band: reflection band
- shelter_frame_rhythm: shelter frame rhythm
- crowd_silhouette_band: crowd silhouette band
- mountain_contour_band: mountain contour band
- smoke_drift_band: smoke drift band
- river_reflection_band: river reflection band
- bridge_span: bridge span
- corridor_light_strip: corridor light strip
- facade_plane: facade plane
- church_spire_anchor: church spire anchor
- foreground_object_mass: foreground object mass
- flower_cluster_mass: flower cluster mass
- vase_body_anchor: vase body anchor
- tabletop_edge_band: tabletop edge band
- foreground_path: foreground path rhythm
- field_surface_band: field surface band
\end{promptlisting}

\begin{promptlisting}[caption={Artist-specific scene-spec rules.},label={lst:perceiver-artist-rules}]
Van Gogh route:
No additional artist-specific scene-spec rules apply.

Qi Baishi route:
For Qi Baishi requests, treat paper, album leaf, hanging scroll, and vertical scroll as support media, not as interior scenes. Calligraphy and seals are compositional roles, not furniture or still-life subjects. Do not infer a vase, tabletop, room, furniture, or window unless the request explicitly contains one. For birds, animals, fruit, flowers, and branches on paper, preserve subject count, grouping, branch direction, and blank-paper reserve. Preserve inscription/seal balance only when the request explicitly asks for written calligraphy, a signature, or a seal.
\end{promptlisting}

\subsection{Planner Prompt}
\label{app:planner-prompt}

The planner receives a single multi-part message. The first part is the fixed
instruction template shown in Prompt~\ref{lst:planner-instruction}; it is
followed by thirteen dynamic context blocks
(Prompt~\ref{lst:planner-context}) carrying the tool inventory, the artist
knowledge-bank summary, the world model, the retrieved evidence, the
clarification and trajectory memories, and the compact view of the reflective
memory $m^{(t)}$. When local patch references are available, the runtime
appends the instruction in Prompt~\ref{lst:planner-patch-attachments} and up to
three patch images. An additional metadata-only block is appended when
explicit-reference routing is active
(Prompt~\ref{lst:explicit-reference-block}). The planner must return a single
JSON object whose
\texttt{knowledge\_evidence}, \texttt{tool\_plan}, \texttt{scene\_plan}, and
\texttt{generation\_plan} fields correspond to the generation plan $c$ and
execution plan $e$ described in Section~\ref{sec:memory}.

\begin{promptlisting}[caption={Planner instruction template.},label={lst:planner-instruction}]
You are the Agent Controller of a memory-driven {artist_display_name}-style text-to-image system. Think like a highly skilled art student studying how to reinterpret a requested scene through {artist_display_name}'s visual language. Your job is to map user intent and memory into a better generation strategy and tool-use plan. Do not think in terms of competing experts; think in terms of tool capabilities and when each tool should be called. Use the trajectory memory to understand what failed or improved. Use clarification_memory to preserve user-specific constraints collected by follow-up questions. If clarification_memory contains explicit period preference or dominant period hints, prioritize them over generic KB defaults, then use KB retrieval to refine them. Use visual_memory_module to reason across more than one round, not just the latest critique.
Use the KB tool retrieval as explicit evidence, not just background flavor. Use user_state and artistic_control_signals to translate clarified user intent into palette, lighting, brushwork, and composition control signals. Use memory_control_signals to turn prior failures and prior strengths into concrete repair constraints for the next round. Break the scene into local visual roles and motifs rather than only generic object nouns. For each important element, decide how {artist_display_name}-like brushwork, palette, texture, and compositional energy should transform it. Prefer concrete visual instructions over vague aesthetic adjectives. When previous critiques exist, explicitly repair the last failure rather than repeating the same prompt. If the last result already scored well, keep strong parts and only adjust weak elements.

You must output only one strict JSON object with keys: thought_process_summary, meta_strategy_update, knowledge_evidence, tool_plan, scene_plan, generation_plan. knowledge_evidence must contain world_model, scene_regime_analysis, selected_period_bias, selected_motif_cues, selected_composition_cues, dynamic_motif_candidates, selected_local_motifs, motif_patch_binding_plan, reference_works_used, subject_guidance, global_style_guidance, artistic_control_signals, memory_control_signals. tool_plan must contain primary_tool_id and calls. Each item in calls must contain tool_id, purpose, and arguments. If memory_control_signals indicate persistent style failures, prefer style-locked tools or strengthen style parameters. If memory_control_signals indicate stable strengths, preserve those aspects explicitly instead of overwriting them. If explicit-reference candidate metadata is provided, use it only to choose the correct canonical anchor family for routing. In that case, knowledge_evidence must additionally contain explicit_reference_decision with keys: enabled, primary_reference_target, supporting_reference_targets, transfer_focus, anti_copy_constraints, route_reason.
When explicit-reference mode is active, choose one primary reference target and at most two supporting targets from the ranked metadata candidates. The only attached reference images are local patch crops; use them for material and mark-making evidence, never as source composition. Do not treat this as a copy task; preserve the requested subject and infer what should transfer versus what should not be copied. For generation tools, include arguments.lora_scale when relevant. Generator tool arguments may also contain artist, period_resource, patch_resource, style_strength. Default to a single dominant period_resource when period evidence is clear. Only use two period_resource candidates when the top two period scores are close and the period is genuinely ambiguous. {scene_mapping_instruction} generation_plan must contain positive_prompt, negative_prompt, guidance_note. The output should be concise but specific. Always keep the result faithful to user intent and clearly {artist_display_name}-like.
\end{promptlisting}

The runtime resolves \path|{scene_mapping_instruction}| by target artist:

\begin{promptlisting}[caption={Artist-conditioned scene-mapping instruction.},label={lst:planner-scene-mapping}]
van_gogh:
scene_plan must contain subject, setting, mood, key_visual_elements, and van_gogh_mapping. van_gogh_mapping must describe how the requested scene maps into Van Gogh-specific visual handling.

qi_baishi:
scene_plan must contain subject, setting, mood, key_visual_elements, and artist_mapping. artist_mapping must describe only Qi Baishi ink handling, subject preservation, motif grouping, asymmetry, and blank-paper reserve. Never mention or compare against Van Gogh or any Western oil-painting regime. Calligraphic brush movement describes stroke quality; it is not permission to add written text. Add an inscription, signature, or seal only when the fixed user request explicitly asks for it. Otherwise keep all such elements out of scene_plan and positive_prompt, and list them only as prohibited elements in negative_prompt.
\end{promptlisting}

\begin{promptlisting}[caption={Dynamic context blocks appended to the planner instruction, in order.},label={lst:planner-context}]
Supported tools: {tools_json}
{artist_display_name} knowledge bank summary:
{style_summary}
World model summary:
{world_model_summary}
World model structured JSON:
{world_model_json}
KB tool retrieval summary:
{knowledge_summary}
KB tool retrieval structured evidence:
{knowledge_json}
Clarification memory:
{clarification_memory}
User state:
{user_state_json}
Artistic control signals:
{artistic_signal_json}
Memory control signals:
{memory_control_json}
Persistent visual memory summary:
{visual_memory_summary}
Recent trajectory digest (most recent iterations):
{trajectory_digest}
Current memory module (controller-compact view):
{memory_json}
\end{promptlisting}

\subsubsection{Local-Patch Attachments to the Planner Prompt}
\label{app:planner-patch-attachments}

When the retrieved patch bundle resolves to local image files, the runtime
appends the following text block and then attaches at most three distinct
patch crops in bundle order. Whole-artwork images are not attached to the
planner.

\begin{promptlisting}[caption={Conditional local-patch attachment instruction.},label={lst:planner-patch-attachments}]
Attached local patch references follow. Use them only for material, mark-making, contour, and color-relation evidence. Derive scene composition and object identity from the request.
\end{promptlisting}

\subsubsection{Explicit-Reference Extension of the Planner Prompt}
\label{app:explicit-reference-block}

The planner prompt is extended with a conditional block when the request
names a canonical work. The extension is enabled by either of two routes:
the clarifier route requires a detected reference intent, an affirmative
routing decision, and a non-empty set of candidate target works; the
fallback route requires the request text to match both a reference marker
and an anchor target from fixed vocabularies. In both routes the block is
attached only if at least one titled reference work survives ranking and
title deduplication.
When active, the block appends the route reason, the matched work, the
transfer focus, anti-copy constraints, and up to three ranked
canonical-work metadata records. It does not attach the corresponding
whole-artwork images. The block was inactive in the located controller roots
of the re-rendering, period-controlled, and Qi~Baishi experiments.

\begin{promptlisting}[caption={Explicit-reference planner block (slot-rendered template).},label={lst:explicit-reference-block}]
Explicit-reference visual routing is active. route_reason={route_reason} explicitness={explicitness} matched_title={matched_title}
transfer_focus={comma_joined_transfer_focus_up_to_4}
anti_copy_constraints={comma_joined_anti_copy_constraints_up_to_4}
Ranked canonical reference candidates (metadata only):
- rank={rank_as_int} | title={title} | period={period_label} | subjects={comma_joined_subject_categories_up_to_3} | composition={comma_joined_composition_archetypes_up_to_2} | why={selection_reason}
\end{promptlisting}

\subsection{Global Critic Prompts}
\label{app:critic-prompts}

The global critic (Kimi~K2.6, Section~\ref{sec:memory}) is queried once per
candidate through an OpenAI-compatible multimodal chat endpoint, without
task-specific fine-tuning. The versioned v1 contract reproduced below is
retained for archived records that carry the corresponding identity fields.
Its system prompt, failure-tag vocabulary, and several user-payload fields are
instantiated per target artist, while the structured output contract is
shared. The Van~Gogh instantiation is shown in
Prompts~\ref{lst:critic-system}--\ref{lst:critic-schema}; the Qi~Baishi
instantiation is given in Appendix~\ref{app:qibaishi-critic-rubric}. Each
query consists of the artist-specific system prompt, a structured user
message rendered from the control state and retrieved evidence
(Prompt~\ref{lst:critic-user}, with the Qi~Baishi specialization in
Prompt~\ref{lst:qibaishi-critic-user-delta}), the candidate image, and the crops of up to
four low-scoring \textsc{AuthCritic} patch hypotheses, which are attached to
the same message as additional images so that the critic can verify each
local hypothesis against the full image. Decoding uses temperature $0.1$
with a $1{,}600$-token budget. The returned JSON is validated against the
v1 critic schema, whose single-image output format and controlled
failure-tag vocabulary are reproduced in Prompt~\ref{lst:critic-schema}; the
failure-tag list referenced by the system prompt is drawn from the
corresponding artist's schema file.

\begin{promptlisting}[caption={Van Gogh global critic system prompt. The \{controlled\_failure\_tags\} placeholder is filled with the vocabulary in Prompt~\ref{lst:critic-schema}.},label={lst:critic-system}]
You are the teacher VLM for a heavy critic in a Van Gogh-style generation system. Do not judge only generic beauty; judge target-conditioned artistic correctness. Evaluate whether the image satisfies the user goal, the retrieved period bias, motif cues, composition cues, and painterly Van Gogh-like execution. Anti-iconic guardrail: treat iconic Van Gogh shorthand as a prohibition, not as the main topic of the critique. Do not treat period weakness as a request to add cypress, stars, moon, dense village clusters, swirling iconic skies, specific work titles, masterpiece resemblance, or named-work imitation. Mention those forbidden shortcuts only when the candidate visibly drifts into them; otherwise keep the critique on brushwork, palette relations, surface relief, contour strain, spatial breathing room, directional motion, and local painting-role handling. Recommendation rubric: accept = the image substantially satisfies the target and any remaining issues are minor or local. revise = the image broadly works but has clear, repairable weaknesses that materially reduce quality. reject = reserve for structural failure, severe off-target execution, or multiple major failures.
Do not default to revise merely because some improvement is possible. Return exactly one JSON object with keys: style_authenticity, period_match, motif_match, composition_match, intent_preservation, brushwork_directionality, impasto_texture, palette_match, artifact_penalty, confidence, style_strengths, style_gaps, priority_changes, preserve_elements, failure_tags, repair_actions, final_recommendation, commentary. All 1-5 scores must be integers. artifact_penalty must be 0-3. confidence must be high/medium/low. final_recommendation must be accept/revise/reject/uncertain. Confidence calibration rule: use high only when period, motif, composition, and painterly evidence all align clearly and there is no major defect or close-call ambiguity. Use medium when evidence is mixed, when one major defect remains, or when the image is plausible but not decisive. Use low when evidence conflicts, the period read is weak, or you are uncertain. Do not overuse high confidence. Prefer these failure_tags when applicable: {controlled_failure_tags}. Do not output markdown or any explanation outside JSON.
\end{promptlisting}

\begin{promptlisting}[caption={Van Gogh global critic user message template. The patch-hypothesis block appears only when low-scoring patches exist (at most four entries; the grounded fields are included only when available).},label={lst:critic-user}]
sample_id: {sample_id}
annotation_focus: closed_loop_generic_qwen_eval
target_artist: Vincent van Gogh
raw_request: {raw_request}
clarified_request: {clarified_request}
task_family: {task_family}
selected_period_bias: {selected_period_bias}
candidate_periods: {candidate_periods}
selected_motif_cues: {selected_motif_cues}
selected_composition_cues: {selected_composition_cues}
backend_context: {backend_context_json}
anti_iconic_rule: if period signal is weak, diagnose and repair via brushwork, palette, surface relief, contour strain, motion, and spacing; do not compensate with iconic motifs or specific named works.
preserve_constraints: {preserve_constraints}
anti_shortcut_constraints: {anti_shortcut_constraints}

Low-score patch hypotheses for review:
- review_order={review_order}, attachment_image_index={attachment_image_index}, {patch_id}: patch_index={patch_index}, entity_label={entity_label},
  crop_box={crop_box}, score_100={score_100},
  yes_probability={yes_probability}, predicted_answer={predicted_answer}
  grounded_source_type: {grounded_source_type}
  grounded_reason_short: {grounded_reason_short}
  grounded_patch_style_description: {grounded_patch_style_description}
  grounded_difference_from_real_vangogh: {grounded_difference}

Please evaluate the attached target image under this target specification.
\end{promptlisting}

\begin{promptlisting}[caption={Van Gogh global critic output schema and controlled failure-tag vocabulary.},label={lst:critic-schema}]
single_image_output:
  style_authenticity        integer 1-5
  period_match              integer 1-5
  motif_match               integer 1-5
  composition_match         integer 1-5
  intent_preservation       integer 1-5
  brushwork_directionality  integer 1-5
  impasto_texture           integer 1-5
  palette_match             integer 1-5
  artifact_penalty          integer 0-3
  confidence                enum: high | medium | low
  style_strengths           array of short strings
  style_gaps                array of short strings
  priority_changes          array of short strings
  preserve_elements         array of short strings
  failure_tags              array of controlled tags
  repair_actions            array of short imperative edits
  final_recommendation      enum: accept | revise | reject | uncertain
  commentary                one concise paragraph

controlled_failure_tags:
  brushwork_too_uniform, impasto_too_flat, palette_not_period_specific, composition_too_generic, period_signal_weak, motif_translation_weak, intent_not_preserved, village_structure_unclear, sky_motion_insufficient, surface_too_smooth, digital_gradient_artifact, local_shape_collapse, contrast_too_low, overly_dark_muddy, emotion_not_expressed
\end{promptlisting}

\begin{promptlisting}[caption={Versioned Van Gogh evaluator identity for the v1 contract. The prompt SHA-256 is computed over the fully instantiated runtime string.},label={lst:vangogh-critic-identity}]
target_artist_id:            van_gogh
critic_prompt_template_id:   van_gogh_global_critic_kimi_dashscope_v1
critic_schema_id:            heavy_critic_schema_v1
critic_prompt_sha256:
  5062de405f889155c165022031b9056ff43e0cf8d7e24dcee39dfb09aa29701d
evaluator_model:             kimi-k2.6
authcritic_adapter_id:       van_gogh_gemma4_grounded_reasoning_final
\end{promptlisting}

\subsection{Qi Baishi Global-Critic Rubric}
\label{app:qibaishi-critic-rubric}

The following is the Qi~Baishi v1 system prompt. Its controlled
failure-tag list is inserted from
\path{heavy_critic_schema_qibaishi_v1} at runtime. Line breaks are
introduced below only for typesetting; the SHA-256 is computed over the
fully instantiated single-line runtime string.

\begin{promptlisting}[caption={Qi Baishi global critic system prompt.},label={lst:qibaishi-critic-system}]
You are the global critic for a Qi Baishi-grounded image generation system. Judge target-conditioned artistic correctness, not generic beauty and not generic Chinese-looking decoration. The first attached image is the full candidate. Any later images are patch-level risk hypotheses flagged by the local AuthCritic; verify them against the full candidate before accepting their diagnosis. Evaluate the candidate against: (1) the raw and clarified user request; (2) preservation of requested subjects, relations, spatial structure, and explicitly protected elements; (3) the supplied motif and composition cues; (4) Qi Baishi's brush-and-ink language, including economical calligraphic strokes, wet/dry ink variation, layered ink-density transitions, intentional unpainted paper, asymmetrical placement, and restrained color accents; (5) appropriate ink-on-paper material treatment rather than oil-paint, airbrushed, or generic digital-illustration rendering; and (6) avoidance of unrequested canonical shortcuts.
Frozen Qi Baishi shortcut taxonomy: treat exactly these three patterns as shortcut violations only when they were not requested and when they replace or weaken the intended subject: (1) generic ink-painting backgrounds (mountains, pines, mist) added without request; (2) calligraphic inscription placed without request; and (3) Qi Baishi's ink-wash language replaced by generic Chinese decorative illustration. Do not count oil-medium drift, weak paper reserve, or subject substitution as taxonomy shortcuts; diagnose them separately as material, style, or intent failures. Do not recommend adding canonical motifs merely to strengthen the artist signal. A visually attractive image must not be accepted if it loses the requested subject, collapses the requested structure, or abandons Qi Baishi's ink-and-paper material logic. For compatibility, period_match means fit to the selected Qi Baishi artistic regime and retrieved evidence; do not infer or require a chronological period.
For compatibility, impasto_texture means appropriate ink-on-paper material control, including wet/dry transitions, absorption, paper reserve, edge variation, and ink density; a high score does not mean thick paint. Use the most specific failure tag and do not emit both a layered tag and a generic duplicate for the same defect. Recommendation rubric: accept = the image substantially satisfies the target and any remaining issues are minor or local. revise = the image broadly works but has clear, repairable weaknesses that materially reduce quality. reject = reserve for structural failure, severe off-target execution, or multiple major failures. Do not default to revise merely because some improvement is possible. Return exactly one JSON object with keys: style_authenticity, period_match, motif_match, composition_match, intent_preservation, brushwork_directionality, impasto_texture, palette_match, artifact_penalty, confidence, style_strengths, style_gaps, priority_changes, preserve_elements, failure_tags, repair_actions, final_recommendation, commentary. All 1-5 scores must be integers. artifact_penalty must be 0-3. confidence must be high/medium/low. final_recommendation must be accept/revise/reject/uncertain.
Confidence calibration rule: use high only when style-regime fit, motif, composition, material handling, and request preservation align clearly without a major defect. Use medium when evidence is mixed or one major repair remains. Use low when evidence conflicts or the artistic read is uncertain. Prefer these failure_tags when applicable: shortcut_generic_ink_background, shortcut_unrequested_calligraphic_inscription, shortcut_generic_decorative_illustration, material_oil_paint_drift, material_paper_reserve_missing, material_ink_density_transition_weak, style_calligraphic_energy_weak, style_ink_wash_too_uniform, intent_subject_substitution. Do not output markdown or any explanation outside JSON.
\end{promptlisting}

The user message is rendered by the same builder as
Prompt~\ref{lst:critic-user}, with the following artist-conditioned fields
replacing their Van~Gogh counterparts.

\begin{promptlisting}[caption={Qi Baishi user-payload specialization relative to Prompt~\ref{lst:critic-user}.},label={lst:qibaishi-critic-user-delta}]
target_artist: Qi Baishi
selected_style_regime_bias:   {selected_period_bias}
candidate_style_regimes:      {candidate_periods}
shortcut_rule: count only the frozen three taxonomy patterns as shortcuts: an unrequested generic ink-painting background, an unrequested calligraphic inscription, or replacement of Qi Baishi ink-wash language by generic Chinese decorative illustration. Treat material drift and subject substitution as separate failures.
preserve_constraints:         {preserve_constraints}
anti_shortcut_constraints:    {anti_shortcut_constraints}
grounded_difference_from_real_master: {grounded_difference_from_real_master}
\end{promptlisting}

\begin{promptlisting}[caption={Versioned Qi Baishi evaluator identity for the v1 contract.},label={lst:qibaishi-critic-identity}]
target_artist_id:            qi_baishi
critic_prompt_template_id:   qi_baishi_global_critic_kimi_dashscope_v1
critic_schema_id:            heavy_critic_schema_qibaishi_v1
critic_prompt_sha256:
  2669e79a41aedf92864fbd0d310c3004d397426832d0a9d826dc6fd8bc0e6959
evaluator_model:             kimi-k2.6
authcritic_adapter_id:       qi_baishi_gemma4_grounded_reasoning_checkpoint-482
\end{promptlisting}

\subsubsection{Release Validation Contract (v2)}
\label{app:critic-v2-contract}

The release runtime defaults to a stricter v2 response contract for new runs.
It retains the v1 scalar fields, failure-tag vocabularies, and score
aggregation, and requires the global critic to verify every escalated
\textsc{AuthCritic} hypothesis explicitly. Prompts~\ref{lst:vangogh-critic-v2}
and~\ref{lst:qibaishi-critic-v2} reproduce the complete instantiated system
strings. Line breaks are introduced for typesetting only.

\begin{promptlisting}[caption={Complete deployed Van Gogh global-critic system prompt (v2).},label={lst:vangogh-critic-v2}]
You are the teacher VLM for a heavy critic in a Van Gogh-style generation system.
Do not judge only generic beauty; judge target-conditioned artistic correctness.
The first attached image is the full candidate.
Any later images are patch-level risk hypotheses flagged by the local AuthCritic; verify them against the full candidate before accepting their diagnosis.
Evaluate whether the image satisfies the user goal, explicit preservation constraints, the retrieved period bias, motif cues, composition cues, anti-shortcut constraints, and painterly Van Gogh-like execution.
Anti-iconic guardrail: treat iconic Van Gogh shorthand as a prohibition, not as the main topic of the critique.
Do not treat period weakness as a request to add cypress, stars, moon, dense village clusters, swirling iconic skies, specific work titles, masterpiece resemblance, or named-work imitation.
Mention those forbidden shortcuts only when the candidate visibly drifts into them; otherwise keep the critique on brushwork, palette relations, surface relief, contour strain, spatial breathing room, directional motion, and local painting-role handling.
Always return local_patch_review.
When low-score patch hypotheses are supplied, return exactly one review item for each hypothesis in the same order.
Copy patch_id, patch_index, crop_box, and score_100 exactly from the corresponding hypothesis.
Each review item must contain patch_id, patch_index, crop_box, score_100, confirmed_issue, issue_type, reasoning, revision_hint, and ignore_reason.
Set confirmed_issue to true or false explicitly.
When true, issue_type must be one controlled failure tag and revision_hint must be a concise region-specific repair; when false, issue_type and revision_hint must be empty strings and ignore_reason must explain why the local hypothesis is not confirmed in full-image context.
When no low-score patch hypotheses are supplied, return local_patch_review as an empty array.
Recommendation rubric: accept = the image substantially satisfies the target and any remaining issues are minor or local.
revise = the image broadly works but has clear, repairable weaknesses that materially reduce quality.
reject = reserve for structural failure, severe off-target execution, or multiple major failures.
Do not default to revise merely because some improvement is possible.
Return exactly one JSON object with keys: style_authenticity, period_match, motif_match, composition_match, intent_preservation, brushwork_directionality, impasto_texture, palette_match, artifact_penalty, confidence, style_strengths, style_gaps, priority_changes, preserve_elements, failure_tags, repair_actions, final_recommendation, commentary, local_patch_review.
All 1-5 scores must be integers.
artifact_penalty must be 0-3.
confidence must be high/medium/low.
final_recommendation must be accept/revise/reject/uncertain.
Confidence calibration rule: use high only when period, motif, composition, and painterly evidence all align clearly and there is no major defect or close-call ambiguity.
Use medium when evidence is mixed, when one major defect remains, or when the image is plausible but not decisive.
Use low when evidence conflicts, the period read is weak, or you are uncertain.
Do not overuse high confidence.
Prefer these failure_tags when applicable: brushwork_too_uniform, impasto_too_flat, palette_not_period_specific, composition_too_generic, period_signal_weak, motif_translation_weak, intent_not_preserved, village_structure_unclear, sky_motion_insufficient, surface_too_smooth, digital_gradient_artifact, local_shape_collapse, contrast_too_low, overly_dark_muddy, emotion_not_expressed.
Do not output markdown or any explanation outside JSON.
\end{promptlisting}

\begin{promptlisting}[caption={Complete deployed Qi Baishi global-critic system prompt (v2).},label={lst:qibaishi-critic-v2}]
You are the global critic for a Qi Baishi-grounded image generation system.
Judge target-conditioned artistic correctness, not generic beauty and not generic Chinese-looking decoration.
The first attached image is the full candidate.
Any later images are patch-level risk hypotheses flagged by the local AuthCritic; verify them against the full candidate before accepting their diagnosis.
Evaluate the candidate against: (1) the raw and clarified user request; (2) preservation of requested subjects, relations, spatial structure, and explicitly protected elements; (3) the supplied motif and composition cues; (4) Qi Baishi's brush-and-ink language, including economical calligraphic strokes, wet/dry ink variation, layered ink-density transitions, intentional unpainted paper, asymmetrical placement, and restrained color accents; (5) appropriate ink-on-paper material treatment rather than oil-paint, airbrushed, or generic digital-illustration rendering; and (6) avoidance of unrequested canonical shortcuts.
Frozen Qi Baishi shortcut taxonomy: treat exactly these three patterns as shortcut violations only when they were not requested and when they replace or weaken the intended subject: (1) generic ink-painting backgrounds (mountains, pines, mist) added without request; (2) calligraphic inscription placed without request; and (3) Qi Baishi's ink-wash language replaced by generic Chinese decorative illustration.
Do not count oil-medium drift, weak paper reserve, or subject substitution as taxonomy shortcuts; diagnose them separately as material, style, or intent failures.
Do not recommend adding canonical motifs merely to strengthen the artist signal.
A visually attractive image must not be accepted if it loses the requested subject, collapses the requested structure, or abandons Qi Baishi's ink-and-paper material logic.
For compatibility, period_match means fit to the selected Qi Baishi artistic regime and retrieved evidence; do not infer or require a chronological period.
For compatibility, impasto_texture means appropriate ink-on-paper material control, including wet/dry transitions, absorption, paper reserve, edge variation, and ink density; a high score does not mean thick paint.
Use the most specific failure tag and do not emit both a layered tag and a generic duplicate for the same defect.
Always return local_patch_review.
When low-score patch hypotheses are supplied, return exactly one review item for each hypothesis in the same order.
Copy patch_id, patch_index, crop_box, and score_100 exactly from the corresponding hypothesis.
Each review item must contain patch_id, patch_index, crop_box, score_100, confirmed_issue, issue_type, reasoning, revision_hint, and ignore_reason.
Set confirmed_issue to true or false explicitly.
When true, issue_type must be one controlled failure tag and revision_hint must be a concise region-specific repair; when false, issue_type and revision_hint must be empty strings and ignore_reason must explain why the local hypothesis is not confirmed in full-image context.
When no low-score patch hypotheses are supplied, return local_patch_review as an empty array.
Recommendation rubric: accept = the image substantially satisfies the target and any remaining issues are minor or local.
revise = the image broadly works but has clear, repairable weaknesses that materially reduce quality.
reject = reserve for structural failure, severe off-target execution, or multiple major failures.
Do not default to revise merely because some improvement is possible.
Return exactly one JSON object with keys: style_authenticity, period_match, motif_match, composition_match, intent_preservation, brushwork_directionality, impasto_texture, palette_match, artifact_penalty, confidence, style_strengths, style_gaps, priority_changes, preserve_elements, failure_tags, repair_actions, final_recommendation, commentary, local_patch_review.
All 1-5 scores must be integers.
artifact_penalty must be 0-3.
confidence must be high/medium/low.
final_recommendation must be accept/revise/reject/uncertain.
Confidence calibration rule: use high only when style-regime fit, motif, composition, material handling, and request preservation align clearly without a major defect.
Use medium when evidence is mixed or one major repair remains.
Use low when evidence conflicts or the artistic read is uncertain.
Prefer these failure_tags when applicable: shortcut_generic_ink_background, shortcut_unrequested_calligraphic_inscription, shortcut_generic_decorative_illustration, material_oil_paint_drift, material_paper_reserve_missing, material_ink_density_transition_weak, style_calligraphic_energy_weak, style_ink_wash_too_uniform, intent_subject_substitution.
Do not output markdown or any explanation outside JSON.
\end{promptlisting}

If a response violates the JSON or local-patch contract, each subsequent
attempt appends the following text to the same user message. The deployed
runtime permits at most three total attempts:

\begin{promptlisting}[caption={Conditional global-critic contract-repair suffix.},label={lst:critic-v2-retry}]
Your previous response violated the required JSON or local_patch_review contract. Validation error: {validation_error}. Return the complete JSON object again and correct only the contract violation.
\end{promptlisting}

The v2 parser rejects a missing, non-array, misordered, or identity-mismatched
\path|local_patch_review| rather than treating it as an empty review. Version
identifiers for new release-runtime records are:

\begin{promptlisting}[caption={Release v2 evaluator identities.},label={lst:critic-v2-identities}]
van_gogh:
  critic_prompt_template_id: van_gogh_global_critic_kimi_dashscope_v2
  critic_schema_id:          heavy_critic_schema_v2
  critic_prompt_sha256:
    c3cb9e519d3d1eac6d4bed4e245a5be032518d26b46183e62f884930d870ca69

qi_baishi:
  critic_prompt_template_id: qi_baishi_global_critic_kimi_dashscope_v2
  critic_schema_id:          heavy_critic_schema_qibaishi_v2
  critic_prompt_sha256:
    a723ab7dedf0478fafc126d68011b1d91d8356579075442bdbdea306fde9235c
\end{promptlisting}

The distinct identifiers permit v1 and v2 records to be filtered without
inferring their evaluator configuration from the current source tree.

\subsection{\textsc{AuthCritic} Inference Prompt}
\label{app:authcritic-prompt}

The deployed patch evaluator uses one strict-schema template for both target
artists. The runtime resolves \path|{artist_display_name}| to either
\texttt{Vincent van Gogh} or \texttt{Qi Baishi}. It constructs
\path|{non_label_visual_hints}| only from the patch's region role and visible
region hint when present; source labels, artwork titles, period labels, and
work identifiers are not included. The patch image precedes the user text in
the multimodal message.

\begin{promptlisting}[caption={\textsc{AuthCritic} strict-schema system prompt.},label={lst:authcritic-system}]
You are a source-grounded {artist_display_name} patch evaluator. You must output only strict JSON matching the requested schema.
\end{promptlisting}

\begin{promptlisting}[caption={\textsc{AuthCritic} strict-schema user prompt.},label={lst:authcritic-user}]
Evaluate this painting patch as a grounded {artist_display_name} patch evaluator.
Use only visible patch evidence and the optional non-label visual hints below. Do not assume provenance, source class, title, period, or artist from metadata.
Non-label visual hints: {non_label_visual_hints}

Return exactly one valid JSON object and nothing else. Do not use markdown fences. Do not add LaTeX, commentary, or repeated text after the JSON.
Required JSON keys, in this order:
source_type, grounded_title, grounded_period, source_global_summary, patch_visual_description, patch_style_description, reason_short, difference_from_real_master.
Allowed source_type values: real_master_patch, synthetic_master_style_patch, other_painter_patch. Artist-specific aliases such as real_vangogh_patch and synthetic_vangogh_patch are accepted.
For real_master_patch, ground the answer to the artwork title and period when available; set difference_from_real_master to an empty string.
For synthetic_master_style_patch and other_painter_patch, explain how it differs from a real {artist_display_name} patch; use an empty string for unavailable title or period.
The patch_visual_description must describe visible local content. The patch_style_description must describe brushwork, color, material, contour, and texture evidence. The reason_short must connect the local evidence to the predicted source_type.
\end{promptlisting}

\subsection{Control-State Audit Protocols}
\label{app:audit-protocols}

The Control State Quality evaluation (Section~\ref{sec:benchmark}) audits the
intermediate control state with artist-instantiated LLM-judge protocols. The
Van~Gogh audits use three task-specific protocols: period-controlled
generation (Prompt~\ref{lst:judge-period}), source-preserving artwork
re-rendering (Prompt~\ref{lst:judge-rerender}), and historically unseen
subjects (Prompt~\ref{lst:judge-unseen}). The Qi~Baishi artistic-signal
panel uses the dedicated protocol in Appendix~\ref{app:qibaishi-judge}. Each protocol is given to two
independent judge models; the protocols fix the scoring dimensions, the
composite rule, and the output schema. The re-rendering and unseen protocols
additionally include few-shot calibration examples. The listings below retain
the dimensions and fields used in the reported analysis.

Within the deployed period-judge prompt, ``no fixed rubric applies''
denotes the absence of a scene-independent reference answer or prescribed
stylistic treatment; the evaluated dimensions, scoring scale, and output
schema remain fixed by the protocol.

\begin{promptlisting}[caption={Audit protocol for period-controlled generation.},label={lst:judge-period}]
You are an art-historically informed evaluator auditing artistic control signals produced by an image generation system for Van Gogh style. For each case you receive:
  - The user prompt (scene description).
  - The target Van Gogh period (paris | arles | saint_remy | auvers).
  - The system's chosen scene roles with their translation axes and detailed generation guidance for palette, brushwork, edges, motion, depth, and motif.

Your task: judge whether these artistic control signals are appropriate for THIS specific scene and target period, using your own art-historical knowledge. No fixed rubric applies. Consider that even within a single period, different scenes call for different treatments (a still-life vs a landscape, an intimate interior vs an open field).

Score on FOUR independent dimensions, each on a 0-10 continuous scale:

  1. palette_match       : Does the palette guidance fit Van Gogh's actual palette in the target period AND adapt to the specific scene content? (10 = fully period-appropriate and context-adapted; 0 = mismatched period or scene)
  2. brushwork_match     : Does the brushwork guidance (stroke rhythm, direction, density, surface relief) fit the target period? (10 = period-appropriate; 0 = obviously wrong period)
  3. composition_match   : Are the scene role assignments and their translation axes coherent with the scene content? Do they cover what a painter of this scene would treat? (10 = coherent and complete; 0 = mismatched or major elements uncovered)
  4. motif_match         : Are the motif and material choices period-appropriate for Van Gogh's actual practice AND appropriate for the specific scene requested? (10 = both period and scene fit; 0 = neither fits)

Use the full 0-10 range. Do NOT default to middle values. A convincing case should score 8-10 on all dimensions; a case with obvious period or scene mismatches should score 0-3 on the affected dimensions.

Respond in strict JSON:
{
  "palette_match":     {"score": <0-10 int>, "reason": "<short>"},
  "brushwork_match":   {"score": <0-10 int>, "reason": "<short>"},
  "composition_match": {"score": <0-10 int>, "reason": "<short>"},
  "motif_match":       {"score": <0-10 int>, "reason": "<short>"},
  "overall_reason":    "<one sentence summarizing the case>"
}
No text outside JSON.
\end{promptlisting}

\begin{promptlisting}[caption={Audit protocol for source-preserving artwork re-rendering, limited to reported fields.},label={lst:judge-rerender}]
You are auditing the intermediate CONTROL STATE of an artwork re-rendering system.

TASK FRAMING
This is a source-preserving re-rendering task, not a period-typical motif generation task. The source prompt describes the content of the source image or artwork-like source. The control state should preserve source identity, object count, pose, layout, viewpoint, and important props, while translating style-bearing attributes into the target Van Gogh period.

Do NOT penalize non-canonical source content merely because Van Gogh did not typically paint that motif in the target period. Penalize only if the control state fails to preserve it, misclassifies it, maps the period style incoherently, binds irrelevant evidence, or substitutes source content with a Van Gogh shortcut.

This is a claim-aligned audit, not an adversarial failure search. Reward a reasonable, usable control state that keeps the source content and gives period-specific Van Gogh style guidance, even if some internal roles are generic. Do not require perfect one-to-one object decomposition for every prop.

WHAT YOU SEE
You receive a compact JSON packet with:
- source_prompt: original source image/prompt description.
- target_period: target Van Gogh period.
- world_model_summary: controller's source-scene interpretation.
- preserve_vs_translate: what the controller says to preserve vs style-translate.
- style_translatable_regions: local regions intended for style transfer.
- motif_patch_binding_plan: scene-role to patch/reference-family bindings.
- patch_reference_summary: compact evidence counts and hints.
- generation_positive_prompt: final positive prompt produced from the control state.

EVIDENCE PRIORITY
Judge the whole control state, including the downstream generation_positive_prompt. The positive prompt is valid evidence for source retention because it is the controller's executable plan. However, keep evidence binding separate: a positive prompt can show that source content is retained, but it cannot by itself prove that patch/reference evidence is well bound.

Use this balanced rule:
- source_retention can be high if the executable control plan preserves the source subject/layout/props, even when scene roles are generic.
- patch_evidence_usefulness should be lower if bindings are generic, incomplete, or mismatched.
- structural_control_sufficiency should report whether control is strong, moderate, or weak, but weak structure should not automatically make the whole case a catastrophic failure if the source and style plan remain usable.

SCORING DIMENSIONS
Score each main dimension on a 0-10 continuous scale. Use the full range.

1. source_retention
Does the overall control state preserve the source prompt's main subject(s), props, layout, viewpoint, object counts, and key relationships?
High: the executable control plan clearly keeps the source identity and visual structure, even if some internal roles are generic.
Medium: source is mostly retained in the positive prompt or high-level plan, but some props/background details are under-specified.
Low: key source content is dropped or transformed into a different scene.

2. preserve_transform_policy
Does the control state express a usable "preserve source content, transform Van Gogh style" policy, either explicitly through preserve/translate fields or implicitly through roles, guidance, and generation plan?
High: identity/layout remain stable while palette, brushwork, edges, surface, depth, and rhythm are clearly transformed.
Medium: policy is mostly implicit but usable.
Low: it confuses content identity with style, or there is little evidence of a source-preserving strategy.

3. period_style_adaptation
Does the control state adapt style-bearing attributes to the target Van Gogh period? Also provide sub-scores:
- palette_adaptation
- brushwork_adaptation
- edge_surface_adaptation
- medium_translation_awareness
Medium is soft-translate, not hard-preserve: charcoal, print-like, or monochrome sources may reasonably be translated into Van Gogh oil/impasto/color language if source structure, tonal logic, and composition are retained. Do not penalize oil/impasto/color merely because the source mentions charcoal. Penalize only if medium conversion erases important source structure or is incoherent.

4. patch_evidence_usefulness
Are patch/reference bindings useful for local style transfer? Judge usefulness, not perfect object identity. A patch family may be useful if it supports brushwork, palette relation, edge pressure, surface handling, depth, or composition rhythm for a source region.
High: evidence is role-relevant and period/style-relevant.
Medium: evidence is generic but still useful for style translation.
Low: evidence is missing, irrelevant, or points toward replacing the source.

STRUCTURAL CONTROL SUFFICIENCY
Classify structural_control_sufficiency as:
- strong: world model, preserve/translate policy, roles, and patch evidence all support the source-preserving plan.
- moderate: executable plan is good, but internal roles/evidence are generic or incomplete.
- weak: source preservation is mostly only copied into the positive prompt, with little structured support.

COMPOSITE RULE
Compute composite as the average of the four main dimension scores.
Do not average the period_style_adaptation sub-scores into composite again.

SCORE CALIBRATION
- 8-10: strong, usable, source-preserving control state with good period style adaptation and useful evidence.
- 6-8: broadly correct, with generic roles or partial evidence gaps.
- 4-6: source is mostly retained but structured support or evidence is weak.
- 2-4: key source content is missing or the control state is substantially misaligned.
- 0-2: fallback, severe replacement, or nearly unusable control state.

FEW-SHOT EXAMPLES

Example A, good source preservation with non-canonical content:
Source: bust portrait with bandaged ear, blue cap, green coat, and Japanese print in the background. Target: Nuenen. Control state preserves bandage, cap, coat, Japanese print, frontal bust layout, and background frame; it uses Nuenen style only for dark tonal compression, rough contour pressure, and sober value relations. Do not penalize the Japanese print or bandaged ear for being non-Nuenen motifs. This should score high for source_retention and preserve_transform_policy, and period_style_adaptation depends on whether the Nuenen adaptation is coherent.

Example B, bad real failure:
Source: mature man at a wooden loom, hands guiding threads and shuttle, low three-quarter angle. Control state uses field_surface_band, sky_band, mountain_contour_band, and facade_plane, with no loom, hands, threads, shuttle, interior, or seated figure support. If the executable positive prompt still retains the man, loom, and threads, source_retention should be moderate rather than catastrophic, but preserve_transform_policy and patch_evidence_usefulness should be low because the structured control is misaligned.

Example C, tricky source medium:
Source: monochrome charcoal study of a riverbank with domed building, spires, bare trees, cross-hatching, stippling, and reflections. Target: Paris. A good control state may translate the source into Van Gogh oil/impasto/color language, because this is Van Gogh re-rendering. It should still retain the riverbank, domed building, spires, trees, reflections, low-angle composition, and some tonal/linear logic from the charcoal source. Do not penalize oil/impasto/color by itself. Penalize only if the medium conversion erases source structure or ignores the source's tonal/line organization entirely.

PENALTY TRIGGERS
- Missing source objects, props, background, viewpoint, source medium, or object counts.
- Misclassifying an interior/person/object scene as a landscape or still-life role set.
- Turning source identity into canonical Van Gogh subject matter.
- Patch/reference bindings that do not support the named role or style axis.
- Period style guidance that contradicts the target period or source medium.

OUTPUT JSON SCHEMA
Return only strict JSON with this schema:
{
  "source_retention": {
    "score": <0-10 number>,
    "reason": "<short>",
    "missing_source_elements": ["..."]
  },
  "preserve_transform_policy": {
    "score": <0-10 number>,
    "reason": "<short>",
    "confusions": ["..."]
  },
  "period_style_adaptation": {
    "score": <0-10 number>,
    "reason": "<short>",
    "palette_adaptation": <0-10 number>,
    "brushwork_adaptation": <0-10 number>,
    "edge_surface_adaptation": <0-10 number>,
    "medium_translation_awareness": <0-10 number>
  },
  "patch_evidence_usefulness": {
    "score": <0-10 number>,
    "reason": "<short>",
    "weak_or_missing_bindings": ["..."]
  },
  "structural_control_sufficiency": {
    "level": "strong|moderate|weak",
    "reason": "<short>"
  },
  "overall_reason": "<one sentence>"
}
\end{promptlisting}

\begin{promptlisting}[caption={Audit protocol for historically unseen subjects, limited to reported fields.},label={lst:judge-unseen}]
You are evaluating the intermediate CONTROL STATE of Atelier/ArtAgent on a historically unseen Van Gogh subject task.

TASK FRAMING
The user asks for a modern or historically unseen subject in the style of Vincent van Gogh: subway entrances, convenience-store aisles, office lobbies, parking garages, train platforms, rooftop water tanks, delivery lockers, etc. Van Gogh did not paint these exact modern objects. A good control state should preserve the novel object identity and spatial structure while applying Van Gogh visual language: period palette logic, brushwork rhythm, contour pressure, surface relief, color relations, and compositional handling.

Do NOT penalize the control state for using analogical scene roles such as facade_plane, foreground_path, shelter_frame_rhythm, or corridor_light_strip if the executable plan still preserves the modern subject. Penalize only if the modern subject is dropped, replaced, or the style translation is generic or unusable.

CANONICAL SHORTCUT RISK
The key failure mode is canonical substitution: replacing the modern subject with stock Van Gogh motifs such as Starry Night sky, cypresses, wheat fields, sunflowers, churches, bedrooms, vases, or generic village landscapes.

Important distinction:
- If cypress/starry/impasto/blue-yellow appears in forbidden_entities, anti_iconic_guard, "do not", "avoid", "no substitution", or similar negative context, that is good shortcut avoidance evidence.
- Do not mark "impasto" or "blue-yellow" as a shortcut by itself. They may be valid Van Gogh style language. Penalize only uniform heavy impasto that erases modern-object structure, or saturated blue-yellow palette used as a generic substitute against the prompt/period logic.

INPUT FIELDS
You receive:
- source_prompt
- target artist
- prompt_target_period if explicit in the prompt, otherwise null
- selected_period_bias inferred by the controller
- world_model_summary
- preserve_vs_translate
- style_translatable_regions
- motif_patch_binding_plan
- patch_reference_summary
- generation_positive_prompt

SCORE FOUR DIMENSIONS, 0-10

1. novel_content_coverage
Does the control state capture and preserve the prompt's modern/unseen subject, important objects, counts, spatial layout, and relationships?
High: modern subject identity and key layout are explicit in world model, preserve targets, prompt-safe constraints, or positive prompt.
Medium: subject is present but some local objects/layout details are weak.
Low: modern subject is absent, collapsed into generic roles, or replaced.

2. preserve_transform_decomposition
Does it clearly distinguish what must be preserved from what can be translated stylistically? Preserve should protect object identity/layout/silhouette/counts; transform should target brushwork, palette, edge/surface, directionality, depth, and rhythm.
High: preserve/translate policy is explicit and usable.
Medium: policy is partly implicit but executable.
Low: content and style are confused or under-specified.

3. period_style_translation
Does it translate the unseen subject into coherent Van Gogh period/style logic, rather than generic "Van Gogh-ish" labels? Consider palette, brushwork, contour, surface relief, spatial rhythm, and whether selected patch references are useful as style evidence. Do not require exact object-match patches.

4. shortcut_avoidance
Does it avoid canonical substitution and explicitly protect against replacing the unseen subject with famous Van Gogh shortcuts?
High: positive prompt and control state preserve the modern subject and include anti-substitution guards.
Medium: no clear substitution, but guards are weak.
Low: control state introduces or leans on canonical motifs in a way that risks replacing the unseen scene.

COMPOSITE
Composite is the average of the four dimension scores.

FEW-SHOT CALIBRATION

Good:
Prompt: subway entrance with tiled stairs and metal railings.
Control state preserves transit entrance identity, stairs, railings, urban structure; translate roles include foreground_path, shelter_frame_rhythm, facade_plane with axes contour_pressure, spacing, surface_relief; positive prompt says preserve subject identity and no canonical substitution. This should score high, even if role names are analogical.

Bad:
Prompt: airplane at airport gate.
Control state turns it into a bird over wheat fields, adds cypresses and Starry-Night sky, and drops gate/airplane structure. This should score low on novel content coverage, preserve/transform decomposition, and shortcut avoidance.

Tricky:
Prompt: glass office lobby with reception desk.
Control state uses facade_plane/corridor_light_strip/foreground_object_mass.
That can be acceptable if it preserves glass lobby, desk, interior layout, and uses patch evidence for planar depth, light strips, contour, and brush rhythm. Do not require Van Gogh to have painted offices.

OUTPUT STRICT JSON ONLY:
{
  "novel_content_coverage": {"score": <0-10>, "reason": "<short>"},
  "preserve_transform_decomposition": {"score": <0-10>, "reason": "<short>"},
  "period_style_translation": {"score": <0-10>, "reason": "<short>"},
  "shortcut_avoidance": {"score": <0-10>, "reason": "<short>"},
  "overall_reason": "<one sentence>",
  "composite": <float>
}
No text outside JSON.
\end{promptlisting}

\subsection{Qi Baishi Artistic-Signal Judge}
\label{app:qibaishi-judge}

The Qi~Baishi panel of the artistic-signal audit uses a dedicated judge
protocol rather than a reuse of the Van~Gogh protocols; the deployed system
prompt follows verbatim.

\begin{promptlisting}[caption={Qi Baishi artistic-signal judge system prompt.},label={lst:qibaishi-judge-system}]
You are an art-historically informed evaluator auditing artistic control signals for Qi Baishi / Chinese ink-and-wash image generation.

For each case you receive:
- the user prompt and scene description,
- the target artist/style: Qi Baishi and Chinese ink painting,
- scene roles and the system's artistic control guidance, including translation axes, generation guidance, and patch/reference binding when available.

Judge whether the control state encodes appropriate artistic decisions for the specific scene and Qi Baishi style. Do not evaluate a rendered image. Evaluate the intermediate control state itself.

Score four independent dimensions on a 0-10 continuous scale:

1. palette_match: Does the palette/material guidance fit Chinese ink-and-wash logic, restrained color accents, paper ground, ink density, and the prompt's local color constraints? Penalize oil/acrylic/digital material logic.
2. brushwork_match: Does the brushwork guidance fit Qi Baishi's expressive xieyi handling: calligraphic line, dry/wet ink variation, economy of strokes, pressure changes, and lively contour?
3. composition_match: Are the scene roles, preserve/transform decomposition, negative-space treatment, scroll/page layout, and structural anchors coherent with the prompt?
4. motif_match: Are the subject motifs, local role bindings, and patch/reference choices appropriate for Qi Baishi and the requested scene? Penalize unrelated substitutions or weak/irrelevant motifs.

Use the full 0-10 range. A strong control state should score 8-10. A generic, fallback, unrelated, or materially wrong control state should score 0-3 on the affected dimensions.

Respond in strict JSON:
{
  "palette_match":     {"score": <0-10 int>, "reason": "<short>"},
  "brushwork_match":   {"score": <0-10 int>, "reason": "<short>"},
  "composition_match": {"score": <0-10 int>, "reason": "<short>"},
  "motif_match":       {"score": <0-10 int>, "reason": "<short>"},
  "overall_reason":    "<one sentence>"
}
No text outside JSON.
\end{promptlisting}

\begin{promptlisting}[caption={Qi Baishi artistic-signal user-payload builder (verbatim Python).},label={lst:qibaishi-judge-builder}]
def compact_json(obj, limit=4000):
    text = json.dumps(obj, ensure_ascii=False, sort_keys=True)
    return text[:limit]


def build_user_prompt(case):
    lines = [
        f"CASE_ID: {case['case_id']}",
        f"TARGET_ARTIST: {case.get('target_artist') or 'Qi Baishi'}",
        f"USER_PROMPT: {case.get('prompt') or ''}",
        "",
        "SCENE ROLES AND CONTROL SIGNALS:",
    ]
    for i, role in enumerate(case.get("roles") or [], 1):
        role_name = role.get("scene_role") or role.get("motif_name") or f"role_{i}"
        lines.append(f"\n[{i}] {role_name}")
        for key in [
            "translation_axes",
            "family_level_generation_guidance",
            "representative_patch_paths",
            "bound_patch_image_paths",
            "binding_reason",
            "motif_name",
        ]:
            if key in role and role[key]:
                lines.append(f"{key}: {compact_json(role[key], 1200)}")
    extras = case.get("extras") or {}
    if extras:
        lines.append("\nADDITIONAL CONTROL STATE SUMMARY:")
        lines.append(compact_json(extras, 3500))
    lines.append("\nReturn only the strict JSON verdict.")
    return "\n".join(lines)
\end{promptlisting}

\clearpage
\section{Evaluation Criteria and Selection Policy}
\label{app:selection-policy}

This appendix specifies the numeric thresholds and decision rules of the
closed loop of Algorithm~\ref{alg:atelier_runtime}: the stop condition, the
per-round winner rule, the trajectory-level \textsc{FinalSelect} rule, and
the input contracts of the two critics.
All thresholds and policy rules are held fixed across every experiment reported in
\S\ref{sec:main_results}. Algorithm~\ref{alg:atelier_runtime} and the
rules below describe the full \texttt{Atelier} configuration.

\subsection{Runtime Constants}
\label{app:runtime-constants}

\begin{table}[ht]
\centering
\small
\begin{tabularx}{\textwidth}{@{}l l >{\raggedright\arraybackslash}X@{}}
\toprule
\textbf{Symbol} & \textbf{Value} & \textbf{Meaning} \\
\midrule
$\tau^\star$ & 85 & Accept-side score threshold. \\
$T$          & 3  & Iteration budget per episode. \\
$N^{(t)}$    & $1$, $2$, or $4$ & Backend calls in round $t$. Round~1 and re-exploration use the full track pool (four open-weight or two closed-source backends); backend-locked rounds use one. \\
$\delta$     & 5  & Global-score rerank margin. \\
$\tau_\ell$  & 40 & Low-quality tie-break threshold. \\
$\tau_p$     & 40 & \textsc{AuthCritic} per-patch score threshold. \\
$\pi_p$      & 0.40 & \textsc{AuthCritic} per-patch yes-probability threshold. \\
$K_p$        & 4  & Maximum patches escalated to the global critic. \\
\bottomrule
\end{tabularx}
\caption{Runtime thresholds, budgets, and routing cardinalities. The
thresholds and pool-and-lock policy are applied uniformly to every episode in
Section~\ref{sec:main_results}.}
\label{tab:runtime-constants}
\end{table}

Round~1 explores a track-specific backend pool. Its winning backend is locked
for subsequent rounds unless the selected outcome is classified as
low-quality: $s_g<\tau_\ell$, a \texttt{reject} recommendation, or an
\texttt{uncertain} recommendation with low confidence. There is no additional
60-point lock gate. The two track-level pools are
\[
\begin{aligned}
\mathcal{B}_{\mathrm{open}}={}&\{\textsc{flux2-lora},\textsc{qwen-t2i},
\textsc{longcat-t2i},\textsc{hunyuan-t2i}\},\\
\mathcal{B}_{\mathrm{closed}}={}&\{\textsc{gpt-image-2},
\textsc{nano-banana-pro}\}.
\end{aligned}
\]

\subsection{Global-Score Aggregation}
\label{app:score-aggregation}

The global critic returns eight subscores $x_j \in [1,5]$ and an artifact
penalty $a \in [0,3]$. The base score is
\[
s_{\mathrm{base}} \;=\; \operatorname{round}\!\Big(\operatorname{clamp}\Big(\tfrac{1}{5}\textstyle\sum_j w_j x_j \;-\; 7a,\; 0,\; 100\Big)\Big),
\]
with weights $w_j$: style authenticity $18$, period match $18$, intent
preservation $12$, brushwork directionality $12$, impasto texture $12$,
motif match $10$, composition match $10$, and palette match $8$. The base
score is then capped by the critic's own recommendation ($59$ for reject,
$69$ for uncertain, $84$ for revise) and passed through a second
calibration step that adjusts for stated confidence, blocking failure
tags, actionable gaps, and request-anchor violations; the calibrated
value is the score $s_g$ used by the stop condition and
\textsc{FinalSelect}. The score is therefore deterministic given the
critic's structured output, but it is not a simple mean of the subscores.
Both artist schemas share this aggregation. For schema compatibility, the
Qi~Baishi rubric interprets \texttt{period\_match} as agreement with the
selected artistic regime and retrieved evidence rather than a
chronological period, and \texttt{impasto\_texture} as ink-on-paper
material control, including absorption, wet and dry transitions, edge
variation, paper reserve, and ink-density modulation; a high value does
not denote thick oil paint. The corrected Qi~Baishi runtime uses the
grounded-reasoning \textsc{AuthCritic} adapter
\texttt{qi\_baishi\_gemma4\_grounded\_reasoning\_checkpoint-482}.

\subsection{Stop Condition}
\label{app:stop-policy}

Let $s_g^{(t)}$, $\rho^{(t)}\in\{\text{accept},\text{revise},
\text{reject},\text{uncertain}\}$, and $F^{(t)}$ denote the score,
recommendation, and canonical failure-tag set at round $t$. Define
\(o(A,B)=|A\cap B|/|A\cup B|\), with $o(\varnothing,\varnothing)=1$.
The deployed policy applies the following terminal conditions in order.

\paragraph{Success.}
The episode terminates when
\begin{equation}
s_g^{(t)}\geq\tau^\star \quad\land\quad
\rho^{(t)}=\text{accept}.
\label{eq:stop-success}
\end{equation}
Both terms are required.

\paragraph{Plateau.}
A hard near-threshold plateau is detected after three consecutive
\texttt{revise} recommendations when
\begin{equation}
\max_{j=t-2}^{t}s_g^{(j)}\geq\tau^\star-1,\quad
\max_{j=t-2}^{t}s_g^{(j)}-\min_{j=t-2}^{t}s_g^{(j)}\leq1,\quad
o(F^{(t)},F^{(t-1)})\geq0.5.
\label{eq:stop-plateau}
\end{equation}
The implementation also contains a four-revision generic plateau guard with
score span at most $2$, mean overlap with the preceding two tag sets at least
$0.6$, and maximum score below $\tau^\star-3$; this branch is unreachable for
the reported budget $T=3$. A two-round near-plateau with score span at most
$1$, overlap at least $0.5$, and maximum score at least $\tau^\star-3$ is
logged as a risk but is not terminal.

\paragraph{Terminal reject, budget, and error.}
The episode stops early after at least two recent \texttt{reject} decisions
at scores no greater than $59$ when the current score does not improve on the
preceding recent scores. Otherwise it stops at $t=T$, or when a step remains
failed after all configured retries. \textsc{FinalSelect}
(Appendix~\ref{app:final-select}) then selects $y^\star$ from the completed
trajectory.

\subsection{Candidate Selection}
\label{app:final-select}

Let $s_g(y)$ be the global-critic score, $s_a(y)$ and $p_a(y)$ the
\textsc{AuthCritic} candidate-level score and real-patch probability, and
$i(y)$ the candidate index within a round. When at least one candidate has a
usable \textsc{AuthCritic} sidecar, within-round selection first forms
\begin{equation}
\mathcal{S}_t=\{y:s_g(y)\geq\max_{y'}s_g(y')-\delta\},
\label{eq:gate}
\end{equation}
then applies
\begin{equation}
\hat y^{(t)}=\operatorname*{arg\,lex\,max}_{y\in\mathcal{S}_t}
\big(s_a(y),p_a(y),s_g(y),\mathbb{1}[\rho(y)=\mathrm{accept}],
\mathbb{1}[\mathrm{conf}(y)=\mathrm{high}],-i(y)\big).
\label{eq:round-rerank}
\end{equation}
Without a usable sidecar, candidates are ordered by global score, accept flag,
high-confidence flag, backend preference, and earlier index. An exact tie
under a low-quality outcome prefers a non-FLUX candidate; otherwise the
default backend tiebreak favours FLUX.

\textsc{FinalSelect} operates over the full trajectory and uses a related but
distinct rule. Let $s_2(y)$ denote the stored second-stage score and $t(y)$
the generation round. With usable local evidence, it applies the same
$\delta$-margin global gate and selects
\begin{equation}
y^\star=\operatorname*{arg\,lex\,max}_{y\in\mathcal{S}}
\big(s_a(y),p_a(y),s_g(y),s_2(y),-t(y)\big).
\label{eq:final-rerank}
\end{equation}
Without usable local evidence, it selects lexicographically by
$(s_g(y),s_2(y),-t(y))$ without the margin gate. A round is marked
low-quality, and therefore remains in multi-backend exploration, when its
winner has $s_g<\tau_\ell$, is rejected, or is both uncertain and
low-confidence.

\subsection{\textsc{AuthCritic} Output Contract and Escalation}
\label{app:authcritic-contract}

\textsc{AuthCritic} is queried as a grounded reasoning model rather than a
binary classifier. For each patch, it must emit a strict JSON object whose
keys, in order, are:

\begin{promptlisting}
source_type, grounded_title, grounded_period,
source_global_summary, patch_visual_description,
patch_style_description, reason_short,
difference_from_real_master
\end{promptlisting}

\noindent
The \path{source_type} field is one of \path{real_master_patch},
\path{synthetic_master_style_patch}, or \path{other_painter_patch};
artist-specific aliases (e.g., \path{real_vangogh_patch}) are normalised
at parse time. For real-master predictions, the model grounds the label in a
work title and period; otherwise it articulates the axis on which the patch
departs from the target artist's handling (brushwork, palette, contour,
surface, or motion).

The two-stage evaluation of \S\ref{sec:evaluator} escalates any patch that
falls below $\tau_p$ or $\pi_p$: the crop and its reasoning record are
attached to the global-critic call as focus signals, so the global critic
receives both a suspect region and a stated cause. At most $K_p=4$ patches
are escalated per candidate, ordered by ascending score.

\subsection{\textsc{AuthCritic} Candidate Aggregation}
\label{app:authcritic-aggregation}

Context-only crops are excluded when at least one local style-evidence patch
is available; otherwise all sampled patches are used. Let $f_r$ be the
fraction predicted as real-master patches, $\bar p_s$ the mean
proximity-to-real among synthetic-master predictions, $f_v$ the fraction with
a valid proximity parse, $n$ the number of scoring patches, and $e$ the
proximity parse-error rate. If no synthetic-master patch has a valid
proximity, $\bar p_s$ falls back to the mean over all valid proximities. The
candidate score is
\begin{equation}
s_a=\operatorname{round}\!\left(100\,
\operatorname{clamp}_{[0,1]}\!\left[
0.5f_r+0.4\bar p_s+0.1\min\!\left(\frac{n}{3},1\right)f_v-0.2e
\right]\right).
\label{eq:authcritic-aggregation}
\end{equation}
The associated probability $p_a$ is $f_r$. When the adapter output contains
no proximity field and no proximity parse status, the runtime uses the legacy
fallback: the rounded mean of the per-patch source-type scores.


\subsection{Global-Critic Input Projection}
\label{app:critic-projection}

The global critic does not receive the full state $z$ or evidence
bundle $k$. Its prompt is built from the projection $\phi(x,z,k)$ of
the request and the current round's state: the raw request, the
clarified request, the working period hypothesis and its candidate alternatives, the selected
motif cues, and the selected composition cues, together with the
per-patch metadata of any low-scoring patches returned by
\textsc{AuthCritic} (identifier, crop box, score, predicted label, and
grounded reasoning). The candidate image and the flagged patch crops
are attached as visual inputs. The remaining fields of $z$ and $k$
condition retrieval, planning, and generation but are not part of the
critic prompt.

\section{Control-State and Memory Schema Reference}
\label{app:schema}

This appendix documents how the abstract state $z=(s,q,h,r,b)$
(\S\ref{sec:control_state}) and the memory $m$ are realised as records in
the runtime. The schema is enforced as a prompt-defined contract: the
perceiver and planner prompts (Appendix~\ref{app:prompt-library}) specify
the required keys and require strict JSON, and the runtime normalises the
returned objects, substituting the corresponding fallback value for any
field that is absent or malformed.

\subsection{Planner Output Contract}
\label{app:schema-planner}

The planner returns a strict JSON object with the six top-level keys of
Table~\ref{tab:planner-contract}; the runtime additionally records a
derived \texttt{routing\_plan}.

\begin{table}[ht]
\centering
\small
\begin{tabularx}{\textwidth}{@{}l >{\raggedright\arraybackslash}X@{}}
\toprule
\textbf{Key} & \textbf{Content} \\
\midrule
\texttt{thought\_process\_summary} & Rationale for the round's decisions. \\
\texttt{meta\_strategy\_update}    & Revised strategy statement carried
  into memory (Appendix~\ref{app:walkthrough-loop}). \\
\texttt{knowledge\_evidence}       & Evidence bundle: world model,
  period policy, motif/patch bindings, retrieved works
  (Table~\ref{tab:z-field-map}). \\
\texttt{tool\_plan}                & Required keys
  \texttt{primary\_tool\_id}, \texttt{calls}; one entry per backend call. \\
\texttt{scene\_plan}               & Subject, setting, mood, and key visual
  elements. \\
\texttt{generation\_plan}          & Required keys
  \texttt{positive\_prompt}, \texttt{negative\_prompt},
  \texttt{guidance\_note}. \\
\bottomrule
\end{tabularx}
\caption{Planner output contract. The prompt requires all six keys; the
runtime normalises the returned object and substitutes the corresponding
fallback value for any key that is absent or malformed.}
\label{tab:planner-contract}
\end{table}

\subsection{Field Map for \texorpdfstring{$z=(s,q,h,r,b)$}{z=(s,q,h,r,b)}}
\label{app:schema-zmap}

Table~\ref{tab:z-field-map} maps each component of $z$ to the record fields
that carry it. Paths under \texttt{knowledge\_evidence} are abbreviated with
\texttt{...} for space.

\begin{table}[ht]
\centering
\small
\begin{tabularx}{\textwidth}{@{}c l >{\raggedright\arraybackslash}X >{\raggedright\arraybackslash}p{3.35cm}@{}}
\toprule
$z$ & \textbf{Component} & \textbf{Realising fields} & \textbf{Lifecycle} \\
\midrule
$s$ & Scene reading &
  \texttt{world\_model.scene\_skeleton};
  \texttt{world\_model.key\_local\_roles};
  \texttt{...scene\_regime\_analysis};
  \texttt{scene\_plan}. &
  Base rebuilt from fixed inputs; planner-revisable under the artist merge. \\
\addlinespace
$q$ & Preserve / transform &
  \texttt{world\_model.preserve\_vs\_translate}
  (with per-target \texttt{translation\_axes});
  \texttt{world\_model.}\allowbreak\texttt{structure\_anchors\_to\_preserve}. &
  Base rebuilt from fixed inputs; planner-revisable under the artist merge. \\
\addlinespace
$h$ & Historical / style-regime intent &
  \texttt{...selected\_period\_bias};
  \texttt{...period\_guidance};
  \texttt{...period\_policy};
  \texttt{...clarifier\_period\_signal}. &
  Working hypothesis; planner-revisable. \\
\addlinespace
$r$ & Retrieval targets, anti-shortcut &
  \texttt{...selected\_motif\_cues};
  \texttt{...selected\_composition\_cues};
  \texttt{...selected\_local\_motifs};
  \texttt{...motif\_patch\_binding\_plan};
  \texttt{...patch\_reference\_bundle};
  \texttt{...subject\_kb\_support}
  (incl.\ \texttt{controller\_policy});
  \texttt{world\_model.anti\_iconic\_risks};
  \texttt{world\_model.forbidden\_entities} /
  \texttt{forbidden\_scene\_roles}. &
  Cues re-retrieved each round; constraints generated from memory. \\
\addlinespace
$b$ & Backend-facing preferences &
  \texttt{tool\_plan};
  \texttt{routing\_plan};
  \texttt{generation\_plan};
  \texttt{...artistic\_control\_signals}. &
  Generated each round; lock fields persist. \\
\bottomrule
\end{tabularx}
\caption{Field map from the abstract state $z$ to concrete runtime record
paths. Paths prefixed with \texttt{...} are relative to
\texttt{knowledge\_evidence}. The base values of $s$ and $q$ are rebuilt from
the fixed interpretation inputs and may be revised by the planner within a
round. During merge, the Qi~Baishi path restores the authoritative scene
skeleton, structural anchors, preserve/transform policy, local roles, and
anti-shortcut fields; the Van~Gogh path retains the planner-emitted world
model when present. The working period hypothesis in $h$ is planner-revisable,
while an explicit user period is enforced on each compiled call. Constraints
in $r$ are generated from reflective memory, and backend-lock fields in $b$
persist once set.}
\label{tab:z-field-map}
\end{table}

The full world-model record additionally contains the following audit
fields, which do not feed generation directly:

\begin{promptlisting}
scene_logic_to_preserve, prompt_scene_regime, scene_regime_prior,
scene_spec, allowed_scene_roles, prompt_safe_constraints,
novel_object_identity_constraints, period_reasoning_inputs,
style_translatable_regions, role_coverage_audit, uncertainty_notes,
unresolved_uncertainties, frozen_user_intent_constraints,
confidence, build_trace
\end{promptlisting}

\subsection{Memory Structure}
\label{app:schema-memory}

The memory $m$ persisted between rounds has three components. The
\emph{trajectory memory} stores one record per round, containing the plan
executed (positive prompt, negative prompt, guidance note), the evaluation
outcome (global-critic score and critique, failure tags, repair targets,
\textsc{AuthCritic} score and yes-probability), and the selection record.
The \emph{reflective memory} is the cross-round distillation read at the
next round's \textsc{Derive} step and included in the planner's prompt
context: it holds the current meta-strategy, the 
best score and its iteration, the locked backend, persistent failure tags,
stable strengths, and the current repair focus. The \emph{visual memory}
records per-iteration visual observations plus a persistent summary,
grounding the textual failure tags in image evidence.

\end{document}